%% file: main.tex
\documentclass[12pt]{article} 
\usepackage{amsmath,amsthm,amssymb,bm,mathrsfs,amsfonts}
\usepackage{algorithm,algorithmic,graphicx,subfig}
\usepackage{caption,enumerate,natbib,listings,setspace}
\usepackage[T1]{fontenc}
\usepackage[colorlinks,linkcolor=red,anchorcolor=blue,citecolor=blue]{hyperref}
\usepackage[margin=1in]{geometry}
\usepackage[protrusion=false,expansion=true]{microtype}
\usepackage[title]{appendix}
\usepackage{bbm}
\usepackage{comment}
\theoremstyle{plain}
\input{def-smile}

\usepackage{xr}
\usepackage{color}
\usepackage{tikz}
\usepackage[parfill]{parskip}
\usepackage{enumitem}
\usepackage{multirow}
\usepackage{array}
\setitemize{noitemsep,topsep=0pt,parsep=0pt,partopsep=0pt}
\setenumerate{noitemsep,topsep=0pt,parsep=0pt,partopsep=0pt}

\newtheorem{example}{Example}

\title{\Large{\textbf{Uncertainty Quantification for Large Language Model Reward Learning under Heterogeneous Human Feedback}} }

\author
{
Pangpang Liu\thanks{Department of Biostatistics, Yale University. Email: pangpang.liu@yale.edu.}\qquad 
Junwei Lu\thanks{Department of Biostatistics, Harvard University. Email: junweilu@hsph.harvard.edu.}\qquad 
Will Wei Sun\thanks{Department of Quantitative Methods, Purdue University. Email: sun244@purdue.edu. Corresponding author.}
}

\begin{document} 

\maketitle

\begin{abstract}
\noindent
We study estimation and statistical inference for reward models used in aligning large language models (LLMs). A key component of LLM alignment is reinforcement learning from human feedback (RLHF), where humans compare pairs of model-generated answers and their preferences are used to train a reward model. However, human feedback is inherently heterogeneous, creating significant challenges for reliable reward learning. To address this, we adopt a heterogeneous preference framework that jointly models the latent reward of answers and human rationality. This leads to a challenging biconvex optimization problem, which we solve via an alternating gradient descent algorithm. We establish theoretical guarantees for the resulting estimator, including its convergence and asymptotic distribution. These results enable the construction of confidence intervals for reward estimates. Leveraging these uncertainty quantification results, we conduct valid statistical comparisons between rewards and incorporate uncertainty into the best-of-$N$ (BoN) policy framework. Extensive simulations demonstrate the effectiveness of our method, and applications to real LLM data highlight the practical value of accounting for uncertainty in reward modeling for LLM alignment.
\end{abstract}


\newpage
\baselineskip=25pt 
\section{Introduction}
Large language models have demonstrated remarkable capabilities across a wide range of applications \citep{nakada2024synthetic,chen2025incentivizing,su2025large,ji2025overview,zhou2025adadetectgpt}. To align these models with human preferences, reinforcement learning from human feedback has become a cornerstone technique \citep{xiao2025algorithmic,ye2025robust,xu2025doubly}. While large language models exhibit impressive generalization capabilities, their raw generations often fail to reflect human values. RLHF addresses this gap by leveraging human feedback to train a reward model that evaluates model outputs and guides subsequent fine-tuning through reinforcement learning. This process enables LLMs to produce responses that are more aligned with human values.

A major challenge in RLHF is the heterogeneity of human preferences. For example, in the \texttt{prism-alignment} dataset \citep{kirk2024prism}, feedback is collected from 1,500 participants spanning 75 countries. These individuals differ in religion, education, employment, and other demographics, and exhibit varying degrees of expertise, attentiveness, and rationality \citep{park2024rlhf,zeng2024}. Such diversity introduces significant heterogeneity into the training data. Failing to account for such differences can lead to misaligned or suboptimal reward models \citep{zhong2024provable,chakraborty2024maxmin}. To address this issue, we employ a scale heterogeneity model (see \eqref{scale} for the details), which explicitly accounts for variations in human rationality during reward learning. Simultaneously learning both the reward function and human rationality leads to a nonconvex optimization problem, which poses significant computational and theoretical challenges. 

Another critical yet often overlooked issue is the uncertainty inherent in the  reward model learned from heterogeneous human feedback. Ignoring this uncertainty can lead to unreliable evaluations. One motivating example arises in computing pairwise ranking accuracy \citep{lambertrewardbench,frickevaluate}. In this task, the goal is to compare two answers to the same question using the learned reward model. Relying solely on point estimates of the rewards provides no information about the confidence in the comparison: whether one answer is truly better than the other or if the difference could be due to estimation noise. A more robust approach would account for the uncertainty in the estimated rewards during the comparison. From a statistical perspective, this type of task can be framed as a hypothesis test problem. Let $r_{\theta_*}(s, a^{(0)})$ and $r_{\theta_*}(s, a^{(1)})$ represent the expected rewards for answers $a^{(0)}$ and $a^{(1)}$ to question $s$ under the true reward parameter $\theta_*$.  We formalize the comparison between these two answers within the framework of the hypothesis test.
\begin{example}
Given a question $s$ and two answers $a^{(0)}$ and $a^{(1)}$ from two different LLMs, we test
\begin{equation}\label{exam1}
H_0: r_{\theta_*}(s, a^{(0)}) - r_{\theta_*}(s, a^{(1)}) = 0
\quad \text{vs.} \quad
H_1: r_{\theta_*}(s, a^{(0)}) - r_{\theta_*}(s, a^{(1)}) \neq 0.    
\end{equation}
\end{example}
To address this challenge, we first estimate a reward model $r_{\theta_T}(s, a)$, where $\theta_T$ serves as an estimator of the true parameter $\theta_*$. We then construct confidence intervals to guide the decision-making process. Our testing procedure is detailed in Section \ref{ard}, and empirical results are reported in Section \ref{cllms}.

Another important application of reward inference lies in improving the best-of-$N$ policy using reward lower bounds. The BoN sampling refers to generating $N$ candidate answers ${a^{(1)}, \cdots, a^{(N)}}$ for a given question $s$ and selecting the best one based on reward estimates \citep{stiennon2020learning,nakano2021webgpt,gao2023scaling,jinnai2024regularized,gui2024bonbon,liu2025pairwise,chow2025inferenceaware}. Traditional BoN strategies use point estimates of rewards without accounting for uncertainty.

However, in offline reinforcement learning, distributional shift is common due to limited coverage of the state-action space. Pessimistic policies are often used to mitigate this issue \citep{zhou2023optimizing,dong2023pasta,lupessimism,bian2024tale,jin2024pessimism,zhu2025semi}. Inspired by this, we can modify the BoN objective to maximize the lower confidence bound of the reward rather than its point estimate.
\begin{example}\label{exm2}
Suppose the asymptotic $(1 - \alpha)$ confidence interval for $r_{\theta_*}(s, a)$ is
$(\mathcal{C}_l^\alpha(s, a),  \mathcal{C}_u^\alpha(s, a))$.
The traditional BoN objective is
\begin{equation}\label{bon}
a_{\text{BoN}}(s) = \mathop{\arg\max}_{a \in \mathcal{A}_{N}(s)} r_{\theta_T}(s, a),
\end{equation}
where $\mathcal{A}_{N}(s)=\{a^{(1)}, \cdots, a^{(N)}\}$ is the set of $N$ candidate answers corresponding to $s$.
The pessimistic BoN (pBoN) policy that maximizes the lower bound of the reward is
\begin{equation}\label{pbon}
a_{\text{pBoN}}(s) = \mathop{\arg\max}_{a \in \mathcal{A}_{N}(s)}  \mathcal{C}_l^\alpha(s, a).
\end{equation}
\end{example}
In Section~\ref{sec4.3}, we show that the expected suboptimality of the pessimistic BoN policy decreases at the rate of $1/\sqrt{n}$, where $n$ denotes the sample size. Here, the suboptimality (see~\eqref{subop}) quantifies the performance gap between the proposed policy and the oracle policy. In Section \ref{abon}, we empirically demonstrate the pessimistic BoN policy in \eqref{pbon} can improve the performance of the standard approach in \eqref{bon} across various settings of BoN policies.

\subsection{Major Contribution}
In this paper, we model the rationality of the human annotator as a function of contextual information and design an alternating gradient descent algorithm to jointly learn both the reward and rationality models. We then derive the asymptotic distribution of the reward estimates and construct statistical confidence bounds to guide robust decision-making. 
The derived uncertainty quantification of reward models can account for the diverse expertise levels of human annotators. Our main contributions are as follows.
\begin{itemize}[leftmargin=*]
\item Methodologically, we employ a heterogeneous rationality model that captures varying levels of human expertise. To jointly learn both the rationality and the reward model, we develop an alternating gradient descent algorithm. 
\item Theoretically, we establish convergence guarantees for the proposed alternating gradient descent algorithm and derive the asymptotic distribution of the estimators, enabling uncertainty quantification of the learned reward. As a byproduct, we propose the pessimistic BoN policy derived from the asymptotic lower bound and establish its suboptimality.
\item Numerically, we validate our approach through extensive simulations and real-world applications to large language models, demonstrating its effectiveness and offering valuable insights to improve the best-of-$N$ policy.
\end{itemize}

\subsection{Related Literature}
Our work intersects with two major areas of literature: modeling heterogeneous human preferences and uncertainty quantification in the Bradley-Terry-Luce (BTL) model \citep{bradley1952rank}.

\textbf{Heterogeneous Human Preferences.}
A number of studies have addressed heterogeneity in human preferences in the context of rank aggregation, including \cite{deng2014bayesian,jin2020rank,li2020extended,li2022bayesian,zhu2023partition}. These works primarily considered the ranking of a fixed set of items and are thus not directly applicable to our setting, which involves comparisons over potentially unbounded LLM outputs. In the RLHF literature, some studies such as \cite{Peter,hao2023,liu2024dual} incorporated heterogeneous preferences by assuming fixed, known parameters that represent each teacher’s area of expertise. In contrast, our model captures heterogeneity through an unknown function of contextual information that must be learned, offering a more flexible and general approach. \cite{lee2024low,zhong2024provable,park2024rlhf,wang2025mpo} proposed personalized reward models that tailor preferences to individual annotators. However, their formulation involves assigning separate reward models to each teacher, which differs from our objective of learning a unified reward model that generalizes across diverse teacher rationalities. More importantly, these prior works concentrate on reward estimation alone, whereas we additionally provide principled uncertainty quantification for the reward learning.

\textbf{Uncertainty Quantification for the BTL Model.} \cite{simons1999asymptotics,han2020asymptotic,gao2023uncertainty,liu2023lagrangian,fan2025ranking} investigated uncertainty quantification in the BTL model without covariates.
\cite{fan2024covariate,fan2024uncertainty} extended the inference to covariate-assisted BTL models. These studies assume a fixed number of items and constant latent scores, which does not align with our setting where LLMs generate a large and varying set of outputs. Recent works, including \cite{wang2024ranking,lu2025contextual,li2025efficient,zhang2025fisher}, have developed contextual ranking frameworks to evaluate and compare different LLMs under uncertainty. \cite{fan2025uncertainty} further explored uncertainty quantification for ranking with heterogeneous preferences in a finite-item setting. These studies focus on assigning a single score to each LLM. In contrast, our work compares multiple outputs generated by LLMs, where the reward is defined for each question–answer pair rather than at the model level. \cite{zhu2023principled, feng2025pilaf} analyzed the asymptotic properties of the maximum likelihood estimator (MLE) under the standard BTL model. Due to the nonconvex nature of our setting, the MLE is not directly attainable. Instead, we quantify the uncertainty of the proposed, attainable gradient-based estimator.

\subsection{Paper Organization}
The remainder of the paper is organized as follows. Section~\ref{sec2} introduces the preliminary setup of the problem. Section~\ref{sec3} presents the framework for reward learning from heterogeneous human feedback. Theoretical results on convergence analysis and uncertainty quantification are provided in Section~\ref{sec4}. We introduce the related applications in Section \ref{sec05}. Section~\ref{sec5} reports the experimental results. All proofs are deferred to the supplementary material.

\section{Problem Formulation}\label{sec2}
We denote $s$ as a prompt (or question) and $a$ as a corresponding response (or answer). The expected reward associated with the pair $(s, a)$ is defined by $r_{\theta_*}(s, a)$, where $\theta_*\in\mathbb{R}^{d_1}$, with $d_1$ fixed, is the true but unknown parameter. Let $x$ represent contextual information that influences a human teacher’s preference rationality. 
We model the teacher's utility for a given prompt-response pair $(s, a)$ using the scale heterogeneity model \citep{fiebig2010generalized}:
\begin{equation}\label{scale}
U(x, s, a)=\sigma_{\gamma_*}(x)r_{\theta_*}(s,a)+\epsilon,
\end{equation}
where $\epsilon$ is independent and identically distributed (i.i.d.) and follows a Gumbel distribution with location parameter 0 and scale parameter 1, and $\sigma_{\gamma_*}(x)\in\mathbb{R}$  is a scale function that captures the human teacher's rationality, parameterized by the unknown vector $\gamma_*\in\mathbb{R}^{d_2}$ with $d_2$ fixed. This model allows the same prompt-response pair to yield different utility outcomes across teachers with different contextual information $x$.
We focus on pairwise comparisons between two candidate responses $a^{(0)}$ and $a^{(1)}$ for a given prompt $s$. The probability that a teacher with context $x$ prefers $a^{(1)}$ over $a^{(0)}$ is given by
\begin{equation}\label{pref}
\begin{aligned}
 \mathbb{P}(Y=1|x, s, a^{(0)}, a^{(1)})&=\mathbb{P}[U(x,s, a^{(1)})>U(x,s, a^{(0)})\mid x,s, a^{(0)}, a^{(1)}]\\
 &=\mathbb{P}\{\epsilon^{(1)}-\epsilon^{(0)}>\sigma_{\gamma_*}(x)[r_{\theta_*}(s,a^{(0)})-r_{\theta_*}(s,a^{(1)})]\mid x,s, a^{(0)}, a^{(1)}\}\\
 &=\frac{1}{1+e^{-\sigma_{\gamma_*}(x)[r_{\theta_*}(s,a^{(1)})-r_{\theta_*}(s,a^{(0)})]}},
\end{aligned}
\end{equation}
where the last equality follows from the fact that the difference between two independent Gumbel random variables with the same scale parameter follows the logistic distribution \citep{kotz2000extreme}.
Here, $Y=1$ indicates that the teacher prefers $a^{(1)}$, and $Y=0$ indicates a preference for $a^{(0)}$. The heterogeneity captured by model \eqref{pref} has been supported by empirical evidence \citep{fiebig2010generalized,davis2019scale,tutz2021uncertain,mauerer2023heterogeneity}. However, these studies typically focus on small-scale applications with a special structure of $\sigma_{\gamma_*}(x)$ and lack theoretical guarantees for estimation accuracy or uncertainty quantification. The value of $\sigma_{\gamma_*}(x)$ reflects the rationality or expertise of the teacher.
If $\sigma_{\gamma_*}(x)>0$, the teacher is more likely to select the response with a higher true reward. If $\sigma_{\gamma_*}(x)\leq 0$, the teacher exhibits non-expert or even adversarial behavior, potentially preferring lower-reward responses. Such irrational preference labels exist in real-world datasets \citep{nahum2024llms,bukharinrobust}. A special case arises when $\sigma_{\gamma_*}\equiv 1$, reducing the model to a homogeneous preference BTL setting \citep{ouyang2022}. Introducing the scale function enhances the expressiveness of the model, but also introduces the challenge of jointly estimating both $\gamma_*$ and $\theta_*$. In the following sections, we develop algorithms to learn these parameters from human feedback and quantify the uncertainty of the estimators.

\section{Reward Learning from Heterogeneous Human Feedback}\label{sec3}
In this section, we present an algorithm for learning the reward function from heterogeneous human feedback. We begin by introducing some structural assumptions on both the reward and scale functions.
\begin{assumption}\label{ass1}
The reward function lies within a family of linear models $r_\theta(s, a)=\theta^\top \phi(s,a)$, where $\phi(s, a): \mathcal{S}\times \mathcal{A}\rightarrow\mathbb{R}^{d_1}$
is a known feature mapping satisfying $\|\phi(s, a)\|_2\leq \phi_{\max}$ for some positive constant $\phi_{\max}$. The true parameter satisfies $\theta_*\in\Theta=\{\theta: \|\theta\|_\infty\leq \theta_{\max}, \theta\in\mathbb{R}^{d_1}\}$ for some positive constant $\theta_{\max}$.
\end{assumption}
Assumption \ref{ass1}  posits that the reward function is linear in the parameter $\theta$, and both the parameter and features are bounded. This assumption is standard in the RLHF literature \citep{zhu2023principled,zhong2024provable,scheid2024optimal,lu2025contextual}. In LLM applications, the feature mapping $\phi(s, a)$ can be obtained by removing the final layer from a pre-trained model \citep{zhu2023principled,zhong2024provable,lu2025contextual}.
\begin{assumption}\label{ass2}
Let $\psi_0: \mathcal{X}\rightarrow \mathbb{R}$ and $\psi(x): \mathcal{X}\rightarrow \mathbb{R}^{d_2}$ be known feature mappings. We assume $\psi_0\not\equiv 0$ and $\sup_{x \in \mathcal{X}}\|(\psi_0(x), \psi^\top(x))\|_2\leq \psi_{\max}$  for some positive constant $\psi_{\max}$. The true rationality scale function is $\sigma_{\gamma_*}(x)=\psi_0(x)+\gamma_*^\top \psi(x)$, where $\gamma_*\in\Gamma=\{\gamma:  \|(1, \gamma^\top)\|_\infty\leq \gamma_{\max}, \gamma\in\mathbb{R}^{d_2}\}$ for some positive constant $\gamma_{\max}$.
\end{assumption}
Assumption~\ref{ass2} is structurally parallel to Assumption~\ref{ass1}. The feature mappings $\psi_0(x)$ and $\psi(x)$ serve as known transformations of the contextual variable $x$, which enrich the representation of features across individuals. The condition $\psi_0(x)\not\equiv 0$ is essential for the identifiability of both $\gamma_*$ and $\theta_*$. Specifically, if $\psi_0 \equiv 0$, the scale function $\sigma_{\gamma_*}(x)=\gamma_*^\top \psi(x)$ and the reward difference $\theta_*^\top [\phi(s, a^{(1)})-\phi(s, a^{(0)})]$ would be indistinguishable in \eqref{pref}. The presence of a nonzero $\psi_0(x)$ ensures that $\gamma_*$ and $\theta_*$ can be distinguished from the data.

With the model specifications in Assumptions \ref{ass1} and \ref{ass2}, we now turn to the estimation of the reward parameter $\theta_*$ and the rationality parameter $\gamma_*$. The goal is to infer these parameters jointly from observed preference data. To this end, we introduce the likelihood formulation associated with the heterogeneous preference model.
Suppose that we observe a dataset of $n$ i.i.d. samples $\{(x_i,s_i, a^{(0)}_i, a^{(1)}_i, y_i)\}_{i=1}^n$. We denote $\mu(v)=\frac{1}{1+e^{-v}}$ for $v\in\mathbb{R}$. By \eqref{pref}, the negative log-likelihood for this dataset is
\begin{equation}\label{lik}
\begin{aligned}
L_n(\theta,\gamma)=-\frac{1}{n}&\sum_{i=1}^n\left\{y_i\log \mu (\sigma_{\gamma}( x_i)(r_\theta(s_i,a_i^{(1)})-r_\theta(s_i,a_i^{(0)})))\right.\\
&\quad\quad\left.+(1-y_i)\log [1-\mu (\sigma_\gamma( x_i)(r_\theta(s_i,a_i^{(1)})-r_\theta(s_i,a_i^{(0)})))]\right\}.
\end{aligned}
\end{equation}
A natural solution to estimate $\theta_*$ and $\gamma_*$ is via the maximum likelihood estimator as follows,
\begin{equation}\label{mle}
\hat{\theta}_n, \hat{\gamma}_n=\mathop{\arg\min}_{\theta\in\Theta, \gamma\in \Gamma} L_n(\theta,\gamma).
\end{equation}
However, the negative log-likelihood $L_n(\theta,\gamma)$ is generally not jointly convex in the parameters $(\theta^\top, \gamma^\top)^\top$, as formalized in the following lemma. 
\begin{lemma}\label{lem1}
Under Assumptions \ref{ass1} and \ref{ass2}, the negative log-likelihood function $L_n(\theta,\gamma)$ defined in \eqref{lik} is convex in $\theta$ when $\gamma$ is fixed, and convex in $\gamma$ when $\theta$ is fixed. However, $L_n(\theta,\gamma)$ is not necessarily jointly convex in $(\theta^\top, \gamma^\top)^\top$.
\end{lemma}
The nonconvex nature of $L_n(\theta,\gamma)$ with respect to the joint parameter vector $(\theta^\top, \gamma^\top)^\top$ makes solving the MLE problem in \eqref{mle} nontrivial. However, the biconvexity property identified in Lemma~\ref{lem1} motivates the use of an alternating optimization approach. Specifically, we apply alternating gradient descent to iteratively update $\theta$ and $\gamma$ described in Algorithm~\ref{alg1}.

\begin{algorithm}[t]
\caption{Reward Learning with Alternating Gradient Descent}\label{alg1}
\begin{algorithmic}[1]
\STATE \textbf{Input}: learning rates $\eta_1$ and $\eta_2$,  initial points $\gamma_0$ and $\theta_0$, number of iterations $T$, dataset $\{(x_i,s_i, a^{(0)}_i, a^{(1)}_i, y_i)\}_{i=1}^n$. 
\STATE $z_i=\phi(s_i, a_i^{(1)})-\phi(s_i, a_i^{(0)})$ for $i=1,\cdots,n$.
\FOR{$t=1\ \text{to}\ T$} 

\STATE $\theta_t=\theta_{t-1}-\eta_1\nabla_\theta L_n(\theta_{t-1},\gamma_{t-1})$, 
 where $$\nabla_\theta L_n(\theta_{t-1},\gamma_{t-1})
=-\frac{1}{n}\sum_{i=1}^n[y_i-\mu(\gamma^\top_{t-1}\psi(x_i)(\theta_{t-1}^\top z_i))][\gamma^\top_{t-1}\psi(x_i)] z_i.$$
\STATE $\gamma_t=\gamma_{t-1}-\eta_2\nabla_\gamma L_n(\theta_{t},\gamma_{t-1})$, where
$$ \nabla_\gamma L_n(\theta_t,\gamma_{t-1})=-\frac{1}{n}\sum_{i=1}^n[y_i-\mu(\gamma^\top_{t-1}\psi(x_i)(\theta_t^\top z_i))](\theta_t^\top z_i)\psi(x_i).$$
\ENDFOR
\STATE \textbf{Output}: $\theta_T, \gamma_T$
\end{algorithmic}
\end{algorithm}
 Algorithm \ref{alg1} requires inputs: learning rates $\eta_1$ and $\eta_2$, initial parameters $\gamma_0$ and $\theta_0$, the number of iterations $T$, and a dataset $\{(x_i, s_i, a^{(0)}_i, a^{(1)}_i, y_i)\}_{i=1}^n$. At each iteration $t = 1, \dots, T$, the algorithm performs alternating gradient updates.
The notations $\nabla_\theta L_n(\theta,\gamma)$ and $\nabla_\gamma L_n(\theta,\gamma)$ denote the gradients of $L_n(\theta,\gamma)$ relative to $\theta$ and $\gamma$, respectively. After $T$ iterations, the algorithm outputs the final parameters $(\theta_T^\top, \gamma_T^\top)^\top$.

\section{Theoretical Results}\label{sec4}
In this section, we present the theoretical guarantees of Algorithm \ref{alg1}. Section \ref{sec4.1} establishes the convergence properties of our estimator, while Section \ref{sec4.2} discusses the statistical inference results.

\subsection{Convergence Analysis}\label{sec4.1}
We first give the assumptions needed for the convergence analysis. We denote $z=\phi(s, a^{(1)})-\phi(s, a^{(0)})$.
\begin{assumption}\label{ass3}
The smallest eigenvalue of the matrix $\mathbb{E}[\sigma^2_{\gamma_*}(x)zz^\top]$ is $\lambda_\phi>0$. Similarly, the smallest eigenvalue of the matrix $\mathbb{E}[(\theta_*^\top z)^2\psi(x)\psi^\top(x)]$ is $\lambda_{\psi}>0$.
\end{assumption}
Assumption \ref{ass3} is a well-conditioned design for the covariance matrix. \cite{zhong2024provable} adopted the same assumption on  $\mathbb{E}[\sigma^2_{\gamma_*}(x)zz^\top]$ with $\sigma_{\gamma_*}\equiv 1$. In the context of LLMs, the feature $x$ represents contextual information (e.g., gender, age, or other demographic factors), while $z$ is associated with the prompt $s$ and candidate responses $a^{(0)}$ and $a^{(1)}$. In this setting, $x$ and $z$ are independent. Then, we have the factorization $\mathbb{E}[\sigma^2_{\gamma_*}(x)zz^\top]=\mathbb{E}[\sigma^2_{\gamma_*}(x)]\mathbb{E}(zz^\top)$ and  $\mathbb{E}[(\theta_*^\top z)^2\psi(x)\psi^\top(x)]=\mathbb{E}(\theta_*^\top z)^2\mathbb{E}[\psi(x)\psi^\top(x)]$. This factorization softens the conditions in Assumption \ref{ass3}, making them more interpretable: they essentially require the variability in both contextual features ($x$) and prompts/responses ($z$) to be sufficiently rich.

\begin{assumption}\label{ass4}
Define $M=\|\mathbb{E}\{\mu(\sigma_{\gamma_*}(x)\theta_*^\top z)[1-\mu(\sigma_{\gamma_*}(x)\theta_*^\top z)]\sigma_{\gamma_*}(x)(\theta_*^\top z)\psi(x)z^\top\}\|_2$. We assume
$M<\frac{\min\{\lambda_\phi,\lambda_{\psi}\}}{6c_0}$, where $c_0$ is a positive constant defined in Lemma \ref{lem02}.
\end{assumption}
Assumption \ref{ass4} controls the interaction strength between $\psi(x)$ and $z$.
The quantity $M$ measures the magnitude of their cross-dependence in the Hessian matrix of the log-likelihood. A smaller $M$ indicates that the coupling between the two parameter blocks is weak, which helps ensure that the alternating estimation of $\theta_*$ and $\gamma_*$ is stable.

Now, we provide the following theorem on the convergence of $(\theta_T^\top, \gamma_T^\top)^\top$ to the true parameters $(\theta_*^\top, \gamma_*^\top)^\top$.
\begin{theorem}\label{thm1}
Let Assumptions \ref{ass1}, \ref{ass2}, \ref{ass3} and \ref{ass4} hold.  Suppose that the initialization in Algorithm~\ref{alg1} satisfies $\|\theta_0-\theta_*\|_2\leq b/\sqrt{2}$ and $\|\gamma_0-\gamma_*\|_2\leq b/\sqrt{2}$, where $b<b_0$ for some positive constant $b_0$. Assume further that the step size $0<\eta_1, \eta_2<\eta_0$ for some positive constant $\eta_0$.  There exist some positive constants $c_1, n_0$ such that when $n>n_0$, for any $0<\delta<1$, with probability at least $1-\delta-\frac{1}{n}$, we have
$$\|\theta_T-\theta_*\|_2^2+\|\gamma_T-\gamma_*\|_2^2\leq \rho^T b^2+\frac{c_1\log (1/\delta)}{(1-\rho)n},$$
for some  $\rho \in (0,1)$.
\end{theorem}
Theorem \ref{thm1} quantifies the convergence rate of Algorithm \ref{alg1}. The first term on the right-hand side represents the optimization error, which vanishes as the number of iterations $T$ increases to $\infty$. The second term indicates the statistical error, which decreases with the sample size $n$ at the scale of $1/n$. The dependence on $T$ and $n$ aligns with the results in \cite{jin2020rank}, which studied a non-contextual setting with a finite number of items. Notably, our setting is more complex, yet we achieve the same convergence rate.

\subsubsection{Challenges and Outline of Proof of Theorem \ref{thm1}}
The proof of Theorem \ref{thm1} presents substantial technical challenges because jointly estimating $(\theta_*^\top,\gamma_*^\top)^\top$ involves a nonconvex objective. The loss function $L_n(\theta,\gamma)$ couples $\theta$ and $\gamma$ multiplicatively through the scale function $\sigma_{\gamma_*}(x)$ and the reward model $r_{\theta_*}(s, a)$, which causes non-convexity and prevents the application of classical convex optimization results. 

To overcome the challenges, we exploit the biconvex structure of $L_n(\theta,\gamma)$ and analyze the alternating gradient descent dynamics in a two-block fashion. Specifically, according to the update rule in Algorithm \ref{alg1}, we derive separate recursive bounds for each block:
\begin{align*}
 \|\theta_{t+1}-\theta_*\|_2^2
 &= \|\theta_t-\theta_*\|_2^2+\eta_1^2\|\nabla_\theta L_n(\theta_t,\gamma_t)\|_2^2
   -2\eta_1\langle\nabla_\theta L_n(\theta_t,\gamma_t), \theta_t-\theta_*\rangle,\\[2mm]
 \|\gamma_{t+1}-\gamma_*\|_2^2
 &= \|\gamma_t-\gamma_*\|_2^2+\eta_2^2\|\nabla_\gamma L_n(\theta_{t+1},\gamma_t)\|_2^2
   -2\eta_2\langle\nabla_\gamma L_n(\theta_{t+1},\gamma_t), \gamma_t-\gamma_*\rangle.
\end{align*}
A central challenge is that these bounds depend on both $t$ and $n$, making it nontrivial to separate the optimization error from the statistical error. Carefully disentangling these dependencies and controlling the cross-block error propagation are the key difficulties. Our strategy is to decompose the gradients 
$\nabla_\theta L_n(\theta_t,\gamma_t)$ and $\nabla_\gamma L_n(\theta_{t+1},\gamma_t)$
into several components: $\nabla_\theta L_n(\theta_*,\gamma_t), \nabla_\theta L_n(\theta_t,\gamma_*), \nabla_\theta L_n(\theta_*,\gamma_*), \nabla_\gamma L_n(\theta_*,\gamma_t), \nabla_\gamma L_n(\theta_{t+1},\gamma_*), \nabla_\gamma L_n(\theta_*,\gamma_*)$.  
This decomposition allows us to isolate the optimization dynamics (through deviations from $\theta_*$ or $\gamma_*$) from the stochastic fluctuations due to finite $n$.

We establish a series of technical lemmas to control each component.  
Lemma~\ref{lem2} ensures that the gradients of $L_n(\theta,\gamma)$ are Lipschitz in both $\theta$ and $\gamma$, thereby limiting the error accumulation during descent.  
Lemma \ref{lem4} quantifies the sensitivity of each block’s gradient to perturbations in the other, ensuring that alternating updates do not destabilize the descent.
Lemma~\ref{lem9} further bounds the mixed second-order derivatives $\nabla_{\gamma\theta}^2 L_n(\theta,\gamma)$, which are essential for controlling cross-block error propagation.
These results together ensure the interdependence between two parameter blocks remains controlled throughout iterations.

After obtaining separate contraction inequalities for $\|\theta_{t+1}-\theta_*\|_2^2$ and $\|\gamma_{t+1}-\gamma_*\|_2^2$, we combine them to study the joint error
$\|\theta_{t+1}-\theta_*\|_2^2+\|\gamma_{t+1}-\gamma_*\|_2^2$.
To guarantee that this joint error decreases over $t$, we derive precise conditions on the learning rates $\eta_1,\eta_2$ and on the initialization radius $b$, ensuring local contraction.  
Establishing these conditions allows us to rigorously prove both the geometric convergence of the optimization iterates and the statistical consistency of the resulting estimators.

\subsection{Uncertainty Quantification of Reward Learning}\label{sec4.2}
In this section, we derive the asymptotic distribution of the estimators obtained from Algorithm \ref{alg1}. For notational convenience, define $\tau_*=(\theta_*^\top, \gamma_*^\top)^\top$ and $\tau_T=(\theta_T^\top, \gamma_T^\top)^\top$. We introduce the Fisher information matrix at the true parameter $\tau_*$ as
$$\mathcal{I}(\tau_*)=\begin{pmatrix}
 \mathcal{I}_{\theta\theta}(\theta_*,\gamma_*) &    \mathcal{I}_{\theta \gamma}(\theta_*,\gamma_*)\\
 \mathcal{I}_{\gamma\theta} (\theta_*,\gamma_*) &\mathcal{I}_{\gamma\gamma} (\theta_*,\gamma_*)
\end{pmatrix}=\begin{pmatrix}
 \mathbb{E}\nabla^2_{\theta\theta} L_n(\theta_*,\gamma_*) &    \mathbb{E}\nabla^2_{\theta \gamma}L_n(\theta_*,\gamma_*)\\
 \mathbb{E}\nabla^2_{\gamma\theta} L_n(\theta_*,\gamma_*) &\mathbb{E}\nabla^2_{\gamma\gamma} L_n(\theta_*,\gamma_*)
\end{pmatrix}$$
where the expectation is taken with respect to the randomness in the preference outcome $y$, the prompt $s$, and the responses $a^{(0)}, a^{(1)}$.  
Here, $\nabla^2_{\theta\theta} L_n(\theta_*,\gamma_*)$ denotes the block of second-order derivatives with respect to $\theta$, and similarly for the other blocks. In order to study the asymptotic properties of our estimators, it is important to ensure that the Fisher information matrix is well-conditioned. The following assumption formalizes this requirement.

\begin{assumption}\label{ass5}
    The matrix $\tilde{\mathcal{I}}(\theta_*,\gamma_*)=\mathcal{I}_{\theta\theta}(\theta_*, \gamma_*)-\mathcal{I}_{\theta\gamma} (\theta_*, \gamma_*)\mathcal{I}^{-1}_{\gamma\gamma} (\theta_*, \gamma_*)\mathcal{I}_{\gamma\theta} (\theta_*, \gamma_*)$ is positive definite.
\end{assumption}
A sufficient condition for Assumption \ref{ass5} is $\mathcal{I}_{\theta\gamma} (\theta_*, \gamma_*)=\boldsymbol{0}$.
Assumption \ref{ass5} guarantees that the information matrix $\mathcal{I}(\tau_*)$ is positive definite, as shown in the following lemma.
\begin{lemma}\label{lem10}
    Suppose Assumptions \ref{ass1}, \ref{ass2}, \ref{ass3} and \ref{ass5} hold. Then the information matrix $\mathcal{I}(\tau_*)$ is positive definite.
\end{lemma}
Lemma \ref{lem10} ensures that the information matrix is positive definite, and therefore invertible. This property is fundamental in asymptotic theory, as the inverse of the Fisher information characterizes the asymptotic covariance matrix of the estimators. Hence, the lemma plays a crucial role in establishing the asymptotic normality and quantifying the statistical uncertainty of the learned parameters.

We now establish the asymptotic normality of $\tau_T$. Recall that $\tau_T$ is obtained after $T$ iterations of Algorithm \ref{alg1}, and its behavior depends jointly on the sample size $n$ and the iteration horizon $T$. The next theorem shows that, under suitable conditions on the growth of $T$ relative to $n$, $\tau_T$ shares the same asymptotic distribution as the MLE.

\begin{theorem}\label{thm6} Let the conditions in Theorems \ref{thm1} and Assumption \ref{ass5} hold. Assume  $n\rho_1^T\rightarrow 0$ as $n\rightarrow\infty$ and $T\rightarrow\infty$, where $0<\rho_1<1$ is some constant. Then, as $n\rightarrow\infty$ and $T\rightarrow\infty$, we have
      \begin{equation*}
\sqrt{n}(\tau_T-\tau_*)\stackrel{D}{\longrightarrow} \mathcal{N}(0, \mathcal{I}^{-1}(\tau_*)). 
\end{equation*} 
\end{theorem}
When $T\rightarrow\infty$ and $n \to \infty$, Theorem \ref{thm1} guarantees that $\tau_T$ converges to $\tau_*$. Theorem \ref{thm6} further establishes that $\tau_T$ not only converges but also satisfies asymptotic normality. This result confirms that the proposed estimator enjoys both consistency and asymptotic efficiency under the stated conditions.

Having established the asymptotic normality of the joint parameter estimator $\tau_T=(\theta_T^\top,\gamma_T^\top)^\top$ in Theorem \ref{thm6}, we now turn to the implications for the learned reward function itself. Since the reward is parameterized only through $\theta$, the asymptotic distribution of $r_{\theta_T}(s,a)$ can be derived accordingly. The following theorem characterizes this distribution.
\begin{theorem}\label{thm3}
Let assumptions in Theorem \ref{thm6} hold. As $n\rightarrow\infty$ and $T\rightarrow\infty$, we have
$$\sqrt{n}[r_{\theta_T}(s, a)-r_{\theta_*}(s, a)]\stackrel{D}{\longrightarrow}\mathcal{N}(0,\phi^\top(s,a)\tilde{\mathcal{I}}^{-1}(\theta_*,\gamma_*)\phi(s,a)),$$
where $\tilde{\mathcal{I}}(\theta_*,\gamma_*)$ is defined in Assumption \ref{ass5}.
\end{theorem}
The matrix $\tilde{\mathcal{I}}(\theta,\gamma)$ is the information matrix for $\theta$, obtained after accounting for the nuisance parameter $\gamma$. Compared to the homogeneous preference model (where $\gamma_*\equiv 1$), the variance of the reward estimator includes an additional adjustment term $\mathcal{I}_{\theta\gamma} (\theta_*, \gamma_*)\mathcal{I}^{-1}_{\gamma\gamma} (\theta_*, \gamma_*)\mathcal{I}_{\gamma\theta} (\theta_*, \gamma_*)$, which accounts for heterogeneity in human preferences captured by $\gamma$. This term inflates the estimated variance, providing a more accurate measure of uncertainty in $r_{\theta_T}(s,a)$. This ensures that the uncertainty properly reflects the additional variability introduced by heterogeneous preferences.

The distributional guarantees established in Theorem \ref{thm3} provide the foundation for performing statistical inference on the learned reward function $r_{\theta_T}(s, a)$. In particular, to construct confidence intervals for $r_{\theta_T}(s, a)$, it is necessary to estimate the variance of the estimator. We define the empirical estimators of the information matrices in our setting as follows:
\begin{align*}
&\hat{\mathcal{I}}_{\theta\theta}(\theta, \gamma)=\frac{1}{n}\sum_{i=1}^n\mu(\sigma_\gamma(x_i)\theta^\top z_i)[1-\mu(\sigma_\gamma(x_i)\theta^\top z_i)]\sigma_\gamma^2(x_i)z_iz_i^\top\\
&\hat{\mathcal{I}}_{\gamma\gamma}(\theta, \gamma)=\frac{1}{n}\sum_{i=1}^n\mu(\sigma_\gamma(x_i)\theta^\top z_i)[1-\mu(\sigma_\gamma(x_i)\theta^\top z_i)](\theta^\top z_i)^2\psi(x_i)\psi^\top(x_i)\\
&\hat{\mathcal{I}}_{\gamma\theta}(\theta, \gamma)=\frac{1}{n}\sum_{i=1}^n\{\mu(\sigma_\gamma(x_i)\theta^\top z_i)[1-\mu(\sigma_\gamma(x_i)\theta^\top z_i)]\sigma_\gamma(x_i)\theta^\top z_i\}\psi(x_i)z_i^\top.
\end{align*}
Using these empirical matrices, we can construct a plug-in estimator of the asymptotic variance for $r_{\theta_T}(s,a)$.
\begin{theorem}\label{thm4}
Let assumptions in Theorem \ref{thm3} hold. As $n\rightarrow\infty$ and $T\rightarrow\infty$, we have
$$\|S_\theta^2(\theta_T, \gamma_T)-\tilde{\mathcal{I}}^{-1}(\theta_*,\gamma_*)\|_2\stackrel{p}{\longrightarrow} 0,$$
where $S_\theta^2(\theta_T, \gamma_T)= [\hat{\mathcal{I}}_{\theta\theta}(\theta_T, \gamma_T)-\hat{\mathcal{I}}_{\theta\gamma}(\theta_T, \gamma_T)\hat{\mathcal{I}}^{-1}_{\gamma\gamma} (\theta_T, \gamma_T)\hat{\mathcal{I}}^{\top}_{\theta\gamma} (\theta_T, \gamma_T)]^{-1}$ and  $\tilde{\mathcal{I}}(\theta_*,\gamma_*)$ is defined in Assumption \ref{ass5}.  Then, as $n\rightarrow\infty$ and $T\rightarrow\infty$, we have
$$\frac{\sqrt{n}[r_{\theta_T}(s, a)-r_{\theta_*}(s, a)]}{\sqrt{\phi^\top(s,a)S_\theta^2(\theta_T, \gamma_T)\phi(s,a)}}\stackrel{D}{\longrightarrow}\mathcal{N}(0,1).$$
\end{theorem}
In Theorem \ref{thm4}, the matrix $S_\theta^2(\theta_T, \gamma_T)$ serves as the empirical counterpart to the asymptotic covariance matrix $\tilde{\mathcal{I}}^{-1}(\theta_*,\gamma_*)$ defined in Assumption \ref{ass5}. When both $T$ and $n$ are sufficiently large, $S_\theta^2(\theta_T, \gamma_T)$ converges to $\tilde{\mathcal{I}}^{-1}(\theta_*,\gamma_*)$.
Let $q_{1-\alpha/2}$ denote the $(1-\alpha/2)$ quantile of the standard normal distribution. Based on Theorem \ref{thm4}, the $(1-\alpha)$ confidence interval for $r_{\theta_*}(s,a)$ is given by
$(\mathcal{C}^\alpha_{l}(s, a), \mathcal{C}^\alpha_{u}(s, a)),$
where
\begin{equation}\label{ci1}
\begin{aligned}
   \mathcal{C}^\alpha_{l}(s, a)=r_{\theta_T}(s, a)-q_{1-\alpha/2}\sqrt{\frac{\phi^\top(s,a)S_\theta^2(\theta_T, \gamma_T)\phi(s,a)}{n}},\\
   \mathcal{C}^\alpha_{u}(s, a)=r_{\theta_T}(s, a)+q_{1-\alpha/2}\sqrt{\frac{\phi^\top(s,a)S_\theta^2(\theta_T, \gamma_T)\phi(s,a)}{n}}.
\end{aligned}
\end{equation}
Finally, we turn to the statistical inference for the rationality parameter $\gamma_*$. This allows us to quantify uncertainty for each component of $\gamma_*$ and test the significance of heterogeneity in human preferences.
\begin{theorem}\label{thm4.1}
Let assumptions in Theorem \ref{thm6} hold. As $n\rightarrow\infty$ and $T\rightarrow\infty$, we have 
$$\|S_\gamma^2(\theta_T, \gamma_T)- [\mathcal{I}_{\gamma\gamma}(\theta_*, \gamma_*)-\mathcal{I}^{\top}_{\theta\gamma}(\theta_*, \gamma_*)\mathcal{I}^{-1}_{\theta\theta} (\theta_*, \gamma_*)\mathcal{I}_{\theta\gamma} (\theta_*, \gamma_*)]^{-1}\|_2\stackrel{p}{\longrightarrow} 0,$$ 
where $S_\gamma^2(\theta_T, \gamma_T)= [\hat{\mathcal{I}}_{\gamma\gamma}(\theta_T, \gamma_T)-\hat{\mathcal{I}}^{\top}_{\theta\gamma}(\theta_T, \gamma_T)\hat{\mathcal{I}}^{-1}_{\theta\theta} (\theta_T, \gamma_T)\hat{\mathcal{I}}_{\theta\gamma} (\theta_T, \gamma_T)]^{-1}$. Then, as $n\rightarrow\infty$ and $T\rightarrow\infty$, we have
$$\frac{\sqrt{n}e_i^{\top}(\gamma_T-\gamma_*)}{\sqrt{e_i^{\top}S^2_{\gamma}(\theta_T, \gamma_T)e_i}}\stackrel{D}{\longrightarrow}\mathcal{N}(0,1),$$
where $e_i$ is the unit vector with 1 in the $i$-th position.
\end{theorem}
Theorem~\ref{thm4.1} establishes the asymptotic normality of each component of the estimated rationality vector $\gamma_T$. Based on this result, we can construct confidence intervals for each element of $\gamma_*$ using the same approach as in \eqref{ci1}. In Section~\ref{sec5.2.1}, we conduct experiments on LLMs to assess the significance of human preference heterogeneity across different contextual factors.

\section{Hypothesis Test and Pessimistic BoN Policy}\label{sec05}
In this section, we explore two practical applications of our distributional results. First, we consider the hypothesis testing of reward differences, which provides a statistically rigorous way to compare and evaluate different LLMs. Second, we demonstrate how the uncertainty quantification of the reward can be incorporated into the BoN policy, enabling more robust decision-making under uncertainty.
\subsection{Application to the Hypothesis Test of Reward Difference}\label{ard}
Now, we explore how to conduct the hypothesis test defined in \eqref{exam1}.   We define
$$\mathcal{T}_{\theta_*}(a^{(0)}, a^{(1)}|s)=r_{\theta_*}(s, a^{(0)})-r_{\theta_*}(s, a^{(1)}).$$
In \eqref{exam1}, the two answers $a^{(0)}$ and $a^{(1)}$ can be treated independently conditional on $s$ since they come from two different LLMs. Then, we can treat $$V(s, a^{(0)}, a^{(1)})=\frac{\phi^\top(s,a^{(0)})S^2(\theta_T, \gamma_T)\phi(s,a^{(0)})+\phi^\top(s,a^{(1)})S^2(\theta_T, \gamma_T)\phi(s,a^{(1)})}{n}$$ as the asymptotic variance of $\mathcal{T}_{\theta_T}(a^{(0)}, a^{(1)}|s)$. Therefore, the $(1-\alpha)$ confidence interval for $\mathcal{T}_{\theta_*}(a^{(0)}, a^{(1)}|s)$ is given by
$(\mathcal{C}^{\alpha}_{l}(s, a^{(0)}, a^{(1)}), \mathcal{C}^{\alpha}_{u}(s,  a^{(0)}, a^{(1)})),$
where
\begin{equation}\label{ci}
\begin{aligned}
\mathcal{C}^\alpha_{l}(s, a^{(0)}, a^{(1)})=\mathcal{T}_{\theta_T}(a^{(0)}, a^{(1)}|s)-q_{1-\alpha/2}\sqrt{V(s, a^{(0)}, a^{(1)})},\\
\mathcal{C}^\alpha_{u}(s,  a^{(0)}, a^{(1)})=\mathcal{T}_{\theta_T}(a^{(0)}, a^{(1)}|s)+q_{1-\alpha/2}\sqrt{V(s, a^{(0)}, a^{(1)})}.
\end{aligned}
\end{equation}
A simple rule for testing the difference is as follows. If $\mathcal{C}^\alpha_{l}(s, a^{(0)}, a^{(1)})>0$, we conclude that the reward of the answer $a^{(0)}$ is statistically larger than that of the answer $a^{(1)}$ at the $(1-\alpha)$ confidence level. If $\mathcal{C}^\alpha_{u}(s, a^{(0)}, a^{(1)})<0$, we conclude that the reward of the answer $a^{(0)}$ is statistically smaller than that of the answer $a^{(1)}$ at the $(1-\alpha)$ confidence level. If $\mathcal{C}^\alpha_{l}(s, a^{(0)}, a^{(1)})\leq 0\leq \mathcal{C}^\alpha_{u}(s, a^{(0)}, a^{(1)})$, we do not have enough evidence to conclude a significant difference between the rewards.

When $a^{(0)}$ and $a^{(1)}$ are not independent, we can obtain an upper bound of the asymptotic variance of $\mathcal{T}_{\theta_*}(a^{(0)}, a^{(1)}|s)$ as $\frac{\left[\sqrt{\phi^\top(s,a^{(0)})S^2(\theta_T, \gamma_T)\phi(s,a^{(0)})}+\sqrt{\phi^\top(s,a^{(1)})S^2(\theta_T, \gamma_T)\phi(s,a^{(1)})}\right]^2}{n}$. This value can then be used to perform the hypothesis test in an analogous manner.

\subsection{Application to the Pessimistic BoN Policy}\label{sec4.3}
In many applications, we are given a set of candidate responses $\mathcal{A}_N(s) = \{a_1, \dots, a_N\}$ for a prompt $s$ and want to select the action with the highest reward. A natural strategy is the BoN policy \citep{stiennon2020learning}, which selects the action with the largest estimated reward, see \eqref{bon}. While straightforward, the standard BoN policy does not account for the uncertainty in reward estimates. Motivated by the pessimistic principle in reinforcement learning \citep{jin2024pessimism},  we select an action in a conservative or risk-averse manner. We explore the pessimistic BoN policy, which accounts for the uncertainty in reward estimates to avoid overestimating the value of an action.

To incorporate statistical guarantees, 
we first provide a bound guarantee for the reward parameter estimation.
\begin{corollary}\label{cor2}
Let the conditions in Theorem \ref{thm4} hold. There exists $\alpha\in(0,1)$, when $T> c \log n$ for some constant $c$, with probability at least $1-\delta-\frac{1}{n}$, we have
\begin{equation*}
\|\theta_T-\theta_*\|_{S_\theta^{-2}(\theta_T,\gamma_T)}\leq \frac{q_{1-\alpha/2}}{\sqrt{n}}.
\end{equation*}
\end{corollary}
Corollary \ref{cor2} is a direct consequence of Theorem \ref{thm1} and provides an upper bound on the deviation of the estimated parameter $\theta_T$ in the weighted norm $\|\cdot\|_{S_\theta^{-2}(\theta_T,\gamma_T)}$. The value of $\alpha$ is related to $d_1, d_2$ and $\delta$. This bound is crucial for constructing a pessimistic estimate of the reward.

Our goal is to select an action that approximately maximizes the true reward: $a_*(s)=\arg\max_{a\in\mathcal{A}_N(s)}r_{\theta_*}(s, a)$. 
To account for uncertainty in $\theta_T$, we define the pessimistic value function:
\begin{equation}\label{pesf}
   \hat{r}(s,a)=\min_{\theta\in\tilde{\Theta}}r_{\theta}(s, a), 
\end{equation}
where $\tilde{\Theta}=\{\theta: \|\theta-\theta_T\|_{S_\theta^{-2}(\theta_T,\gamma_T)}\leq q_{1-\alpha/2}/\sqrt{n}$. The following lemma provides an equivalent closed-form expression for $\hat{r}(s,a)$, making it straightforward to compute in practice.
\begin{lemma} \label{lowpess}
Suppose that Assumption \ref{ass1} holds. Let $\hat{r}(s,a)$ be defined in \eqref{pesf}. Then, we have
    \begin{align*}
\hat{r}(s,a)=r_{\theta_T}(s, a)-q_{1-\alpha/2}\sqrt{\frac{\phi^{\top}(s, a)S_\theta^{2}(\theta_T,\gamma_T)\phi(s, a)}{n}}.
\end{align*}
\end{lemma}
Combining \eqref{ci1} with Lemma~\ref{lowpess}, we see that the pessimistic value function $\hat{r}(s,a)$ corresponds to the asymptotic lower confidence bound of the reward at a significance level $\alpha$.
The pessimistic BoN policy selects the action that maximizes this conservative estimate:
$$a_{pBoN}(s)=\arg\max_{a\in \mathcal{A}_N(s)}\hat{r}(s, a),$$ where $\mathcal{A}_N(s)$ is defined in \eqref{bon}. We measure the performance of this policy via the expected suboptimality (the performance gap between one policy and the oracle policy):
\begin{equation}\label{subop}
    \textsf{SubOpt}(a_{pBoN})=\mathbb{E}_s[r_{\theta_*}(s, a_*(s))-r_{\theta_*}(s, a_{pBoN}(s))].
\end{equation}

We now establish the performance guarantee of the pessimistic BoN policy. The following theorem provides a high-probability upper bound on the suboptimality of the action selected under this policy.  

\begin{theorem}\label{thm7}
Let the conditions in Theorem \ref{thm4} hold. When $T > T_0$ and $n > n_0$ for some constants $T_0$ and $n_0$, then with probability at least  
$1 - \delta - \frac{1}{n},$
we have
\[
\textsf{SubOpt}(a_{pBoN}) \leq \frac{ 3q_{1-\alpha/2}\|\tilde{\mathcal{I}}^{-\frac{1}{2}}(\theta_*,\gamma_*)\mathbb{E}_s\phi(s, a_*(s))\|_2}{\sqrt{n}},
\]
where $\tilde{\mathcal{I}}(\theta_*,\gamma_*)$ is defined in Assumption \ref{ass5}.
\end{theorem}

Theorem~\ref{thm7} shows that the suboptimality of the pessimistic BoN policy decays at the statistical rate of $1/\sqrt{n}$. Importantly, the bound depends on the term $\|\tilde{\mathcal{I}}^{-\frac{1}{2}}(\theta_*,\gamma_*)\mathbb{E}_s\phi(s, a_*(s))\|_2$, which reflects how well the dataset covers the distribution of optimal responses $a_*(s)$. This highlights that the performance guarantee is not tied to uniform coverage over the entire state (prompt) and action (response) space, but instead only to coverage along the optimal trajectory, consistent with the insights of \citet{jin2024pessimism}.  
Recent works on pessimistic policies for RLHF \citep{zhu2023principled,zhangmitigating,liu2024dual} primarily rely on finite-sample lower bounds. Our approach leverages an asymptotic lower bound derived from distributional guarantees, and extends the pessimistic framework to the BoN setting.

\section{Experiments}\label{sec5}
In this section, we present numerical experiments using both synthetic and real-world data. Section \ref{sim} focuses on simulation studies that illustrate the convergence behavior of Algorithm \ref{alg1} and validate the distributional properties. In Section \ref{llm}, we demonstrate the practical applicability of our framework by evaluating it on large language models.
\subsection{Simulations}\label{sim}
We begin by describing the data generation process used in our simulation studies.  The
prompt variable $s$ is sampled from the standard normal distribution $\mathcal{N}(0, 1)$, and the actions
$a^{(0)}$ and $a^{(1)}$ drawn from $\mathcal{N}(0, 1)$ and $\mathcal{N}(0, 2)$, respectively. The feature mappings are defined as $\phi(s, a)=(s^2a, a^2s, as )^\top$, $\psi_0(x)=x$ and $\psi(x)=(x^3, x^2)^\top$ with $x$ sampled from $\mathcal{N}(0, 1)$. The ground-truth parameters are set to $\theta_*=(1/4, 1/2, 1/3)$ and $\gamma_*=(1/2, 1/3)$. We initialize $\theta_0$ and $\gamma_0$ by sampling from a uniform distribution $\mathcal{U}(-1,1)$, and set learning rates as $\eta_1=0.1$ and $\eta_2=0.08$. 

\begin{figure}[h]
    \centering
    \includegraphics[width=0.4\textwidth]{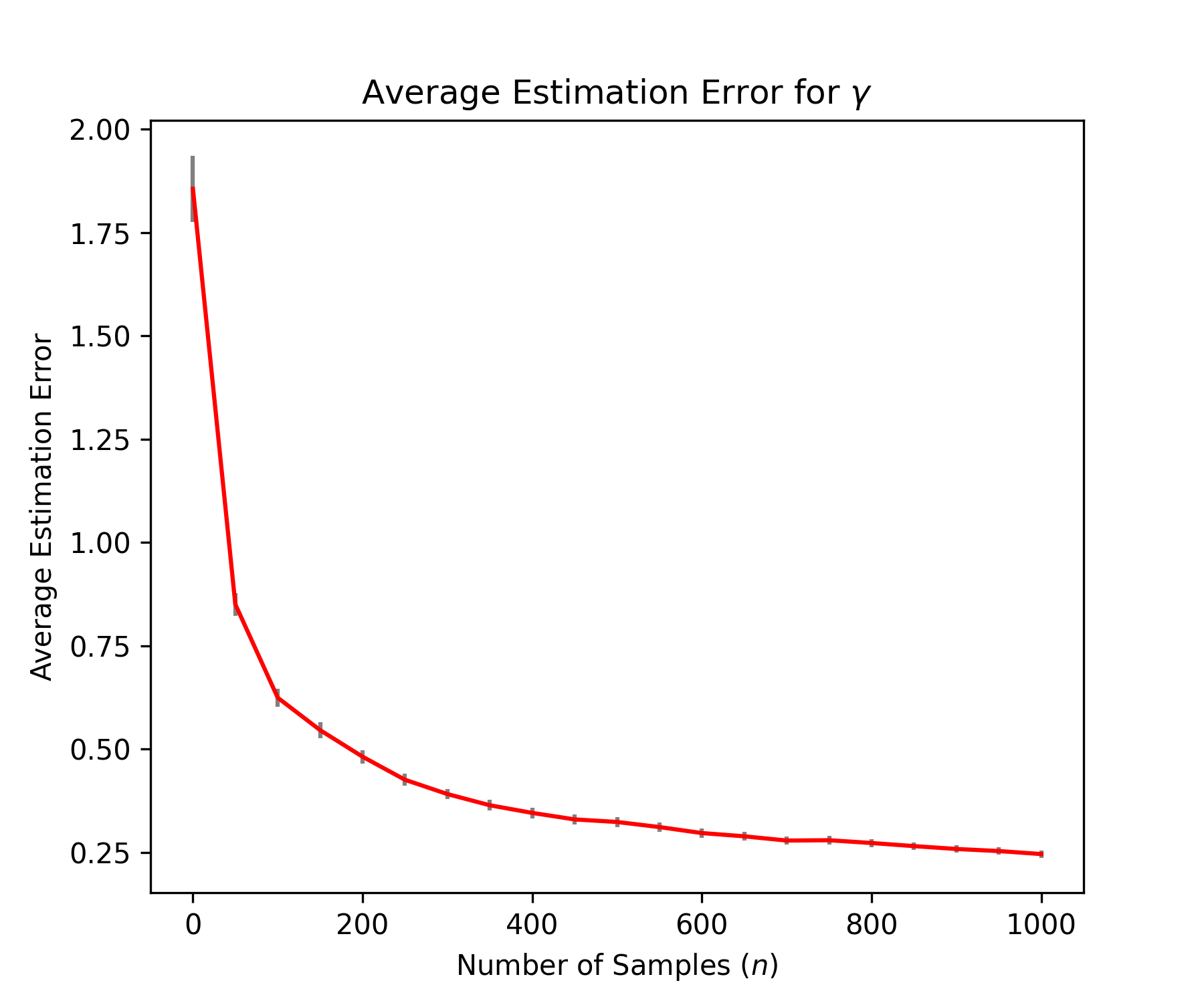}\includegraphics[width=0.4\textwidth]{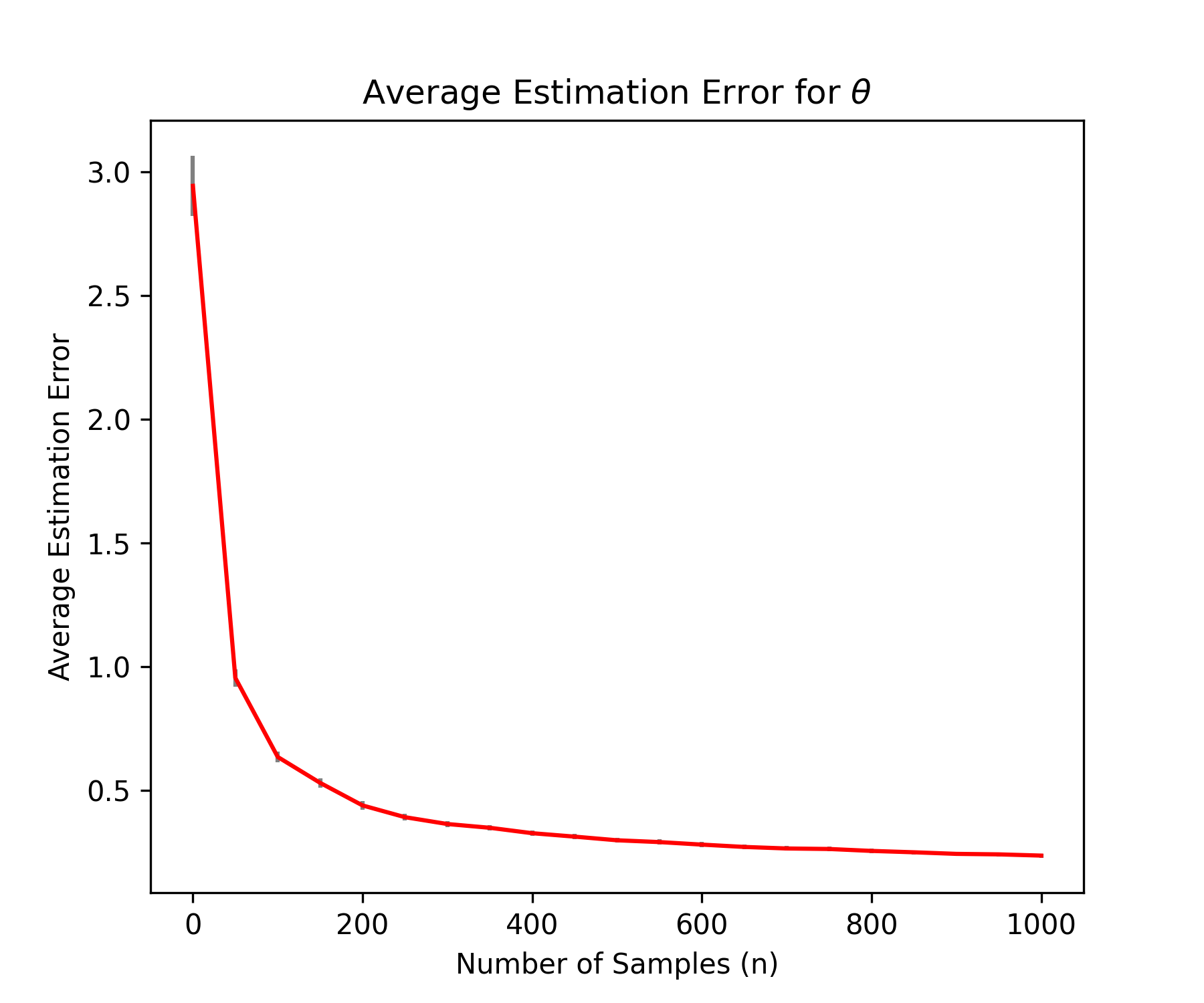}
        \includegraphics[width=0.4\textwidth]{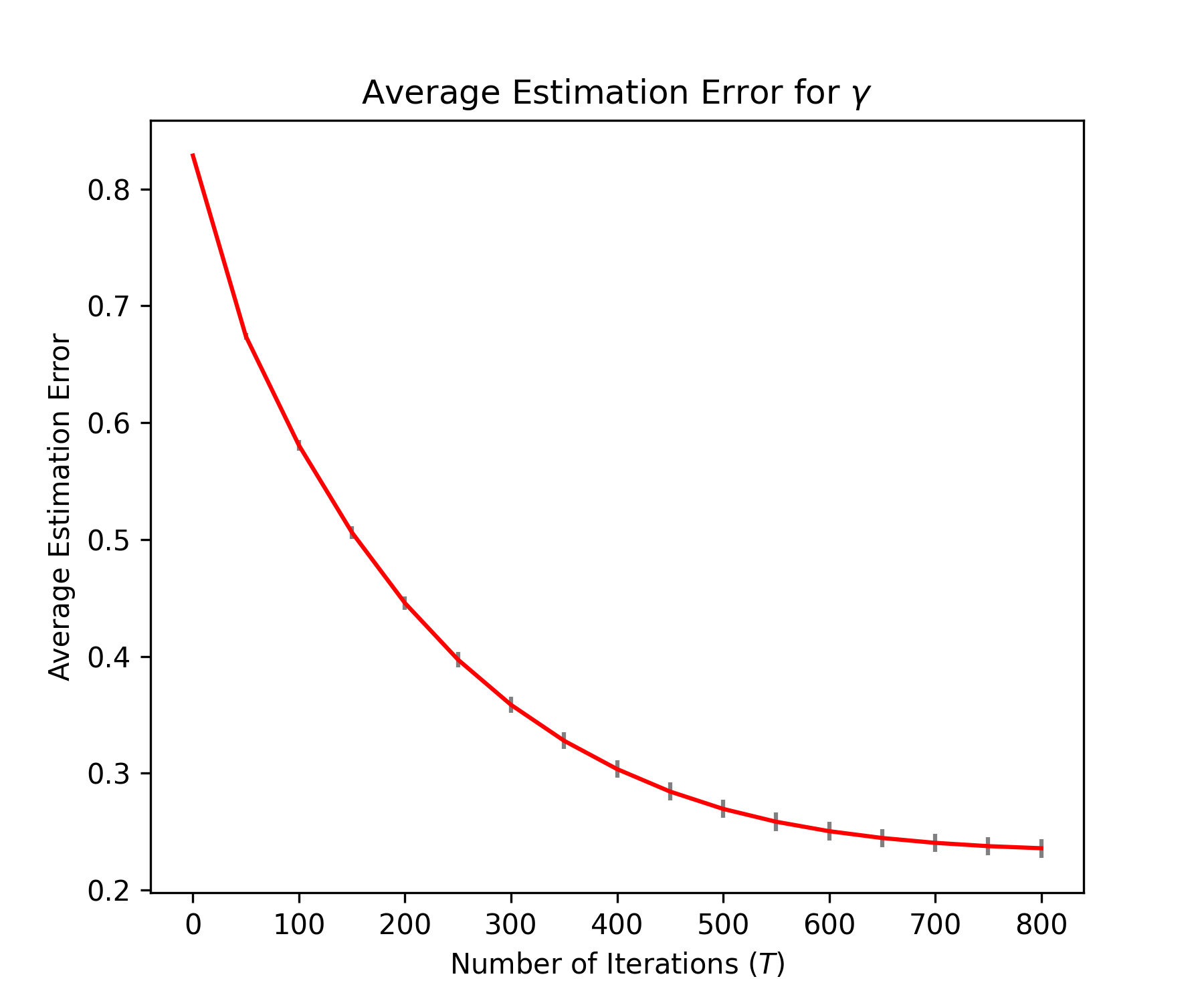}\includegraphics[width=0.4\textwidth]{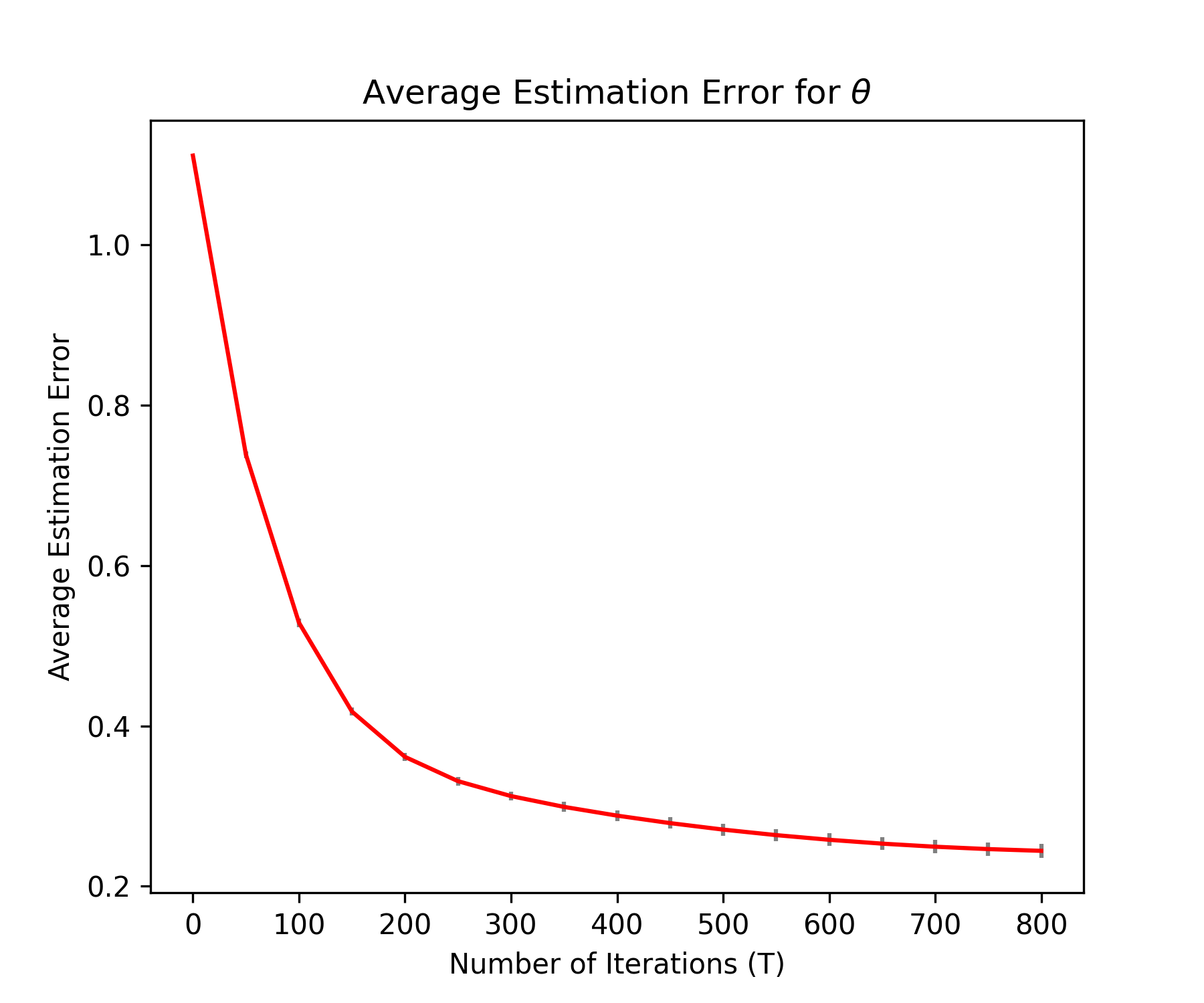}
    \caption{Estimation errors along with the sample size $n$ and iterations $T$}
    \label{fig1}
\end{figure}

We first evaluate the estimation errors of $\theta_*$ and $\gamma_*$ over various sample sizes $n$ and iterations $T$, averaging results over 100 trials. The results, shown in Figure~\ref{fig1}, demonstrate that the estimation error of the heterogeneous model decreases as both $n$ and $T$ increase, which aligns with the theoretical findings in Theorem~\ref{thm1}. 

Next, we  examine the empirical distribution of the learned estimators via coverage probabilities. We consider sample sizes $n=200, 400, 600$. Iteration counts are chosen appropriately to ensure algorithmic convergence. We set the nominal coverage probability $1 - \alpha= 0.95$. The performance of the estimator is assessed by the average coverage rate (Cov Rate) and the average length (Avg Len) of the confidence intervals for each parameter component. Table \ref{tab0} presents the average coverage rates and confidence interval lengths, averaged over 2000 trials. Standard errors are reported in parentheses. The average coverage rate closely matches the nominal level of 0.95, and the average length of the confidence intervals decreases with increasing $n$.
\begin{table}[!ht]
\caption{Coverage rates and average lengths of confidence intervals for the estimator}
\label{tab0}
    \centering
    \begin{tabular}{cccc}
    \hline
        ~ & ~ & $n$ & ~ \\ \hline
         ~ & 200 & 400 & 600 \\ \hline
        Cov Rate & 0.958 (0.006) & 0.948 (0.005) & 0.952 (0.004) \\ 
          Avg Len & 2.547 (0.018) & 1.636 (0.011) & 1.263 (0.008)  \\ \hline
    \end{tabular}
\end{table}

We further assess the coverage of the learned reward function $r_{\theta_T}(s, a)$. In particular, we evaluate at  $(s, a)=(1/2, 1/4)$, $(s, a)=(1/2, 1/2)$, $(s, a)=(1, 1/4)$, and  $(s, a)=(1, 1)$ with results summarized in Table~\ref{tab1}. The observed coverage rates again align well with the nominal level, and the interval lengths decrease as $n$ increases.
\begin{table}[!ht]
\caption{The average coverage rate and length of confidence intervals for the reward}
\label{tab1}
\centering
\begin{tabular}{ccccc}
\hline
    ~ & ~ & ~ & $n$ & ~ \\ \hline
    $(s, a)$ & ~ & 200 & 400 & 600 \\ \hline
    \multirow{2}{*}{$\left (\frac{1}{2}, \frac{1}{4}\right)$} & Cov Rate & 0.943 (0.002) & 0.923 (0.001) & 0.951 (0.001) \\ 
    ~ & Avg Len & 0.419 (0.003) & 0.275 (0.002) & 0.224 (0.001) \\
    \multirow{2}{*}{$\left (\frac{1}{2}, \frac{1}{2}\right)$} & Cov Rate & 0.917 (0.003) & 0.922 (0.002) & 0.953 (0.002) \\ 
    ~ & Avg Len & 0.770 (0.006) & 0.511 (0.003) & 0.417 (0.002) \\
    \multirow{2}{*}{$\left (1, \frac{1}{4}\right)$} & Cov Rate & 0.932 (0.004) & 0.922 (0.003) & 0.949 (0.002) \\ 
    ~ & Avg Len & 0.904 (0.007) & 0.596 (0.004) & 0.488 (0.002) \\ 
    \multirow{2}{*}{$\left (1, \frac{1}{2}\right)$} & Cov Rate & 0.917 (0.007) & 0.923 (0.005) & 0.953 (0.004) \\ 
    ~ & Avg Len & 1.688 (0.013) & 1.125 (0.007) & 0.921 (0.004)\\ \hline
\end{tabular}
\end{table}

\subsection{Applications to LLMs}\label{llm}
In this section, we apply our proposed methods to large language models to demonstrate their practical utility. We begin by training both a reward model and a rationality model. Using the learned rationality model, we compute confidence intervals for the rationality parameters and uncover evidence of heterogeneity in human rationality. Next, we estimate confidence intervals for the difference in rewards between two responses, which enables pairwise comparison of outputs while accounting for uncertainty. We then illustrate how this framework can be used to compare two LLMs under uncertainty. Finally, we incorporate reward model uncertainty into the BoN policy and show that doing so can lead to improved performance.

\subsubsection{Training of Reward Model and Rationality Model}\label{sec5.2.1}

We begin by training the reward model $r_{\theta_*}(s,a)$ and the rationality model $\sigma_{\gamma_*}(x)$ using the dataset \texttt{prism-alignment}\footnote{\href{https://huggingface.co/datasets/HannahRoseKirk/prism-alignment}{https://huggingface.co/datasets/HannahRoseKirk/prism-alignment}} \citep{kirk2024prism}. Each record in the dataset contains multi-turn dialogues. For our analysis, we retain only the first turn of each dialogue. For a given prompt, multiple responses are generated by different LLMs, and human annotators provide evaluation scores. We construct pairwise comparisons of responses, assigning the higher-scored response as the preferred one.  

The dataset involves feedback from 1500 annotators, along with rich contextual information. We include the following user-level features $x$: gender, age, education, employment, marital status, English fluency, religion, location, ethnicity, familiarity with LLMs, and direct usage of LLMs. All categorical variables are one-hot encoded. After preprocessing, we obtain 42,306 paired comparison samples, where each observation consists of a user profile $x$, a prompt $s$, two candidate responses $a^{(0)}$ and $a^{(1)}$, and a binary preference label $y$. We split the dataset into training and test sets with an 80/20 ratio. The test data is used for hyperparameter selection (e.g., learning rate, batch size, and number of epochs). The final training configuration uses a learning rate of $5\times 10^{-5}$, batch size of 32, and 20 epochs.  

For representation learning, we adopt the pretrained model \texttt{opt-1.3b}\footnote{\href{https://huggingface.co/facebook/opt-1.3b}{https://huggingface.co/facebook/opt-1.3b}} \citep{zhang2022opt}. The feature mapping $\phi(s,a)$ is obtained by removing the final layer of the model, producing a $d_1=2048$ dimensional embedding for each $(s,a)$ pair. For the context information $x$, we define $\psi_0(x)=x_{[1]}$ (the first component of $x$) and $\psi(x)=x_{[2:d_2]}$ (the remaining components), with $d_2=38$. Finally, we jointly estimate the reward model parameters $\theta_*$ and the rationality parameters $\gamma_*$.  

\begin{table}[!ht]
\caption{Estimation of Selected Rationality Parameters}
\label{tab01}
    \centering
    \begin{tabular}{ccc|ccc}
    \hline
        Variables & Coef & CI & Variables & Coef & CI \\ \hline
        age: 35--44 & 0.188 & (0.183, 0.192) & education: other & -0.050 & (-0.060, -0.041) \\ 
        age: 45--54 & 0.252 & (0.247, 0.257) & religion: other & -0.035 & (-0.048, -0.023) \\ 
        age: 55--64 & 0.283 & (0.277, 0.289) & marital: never married & 0.130 & (0.126, 0.134) \\ 
        gender: male & 0.053 & (0.050,  0.057) & employment: full-time & 0.024 & (0.019, 0.030) \\ 
        location: US & 0.158 & (0.149, 0.167) & llm direct use: unsure & -0.079 & (-0.087, -0.070) \\ 
        English: fluent & 0.009 & (0.004, 0.014) & llm: somewhat familiar & 0.203 & (0.196, 0.210) \\  \hline
    \end{tabular}
\end{table}

To illustrate the heterogeneity in annotators’ preferences, Table~\ref{tab01} reports a subset of the estimated rationality parameters $\gamma_*$. The confidence intervals (CI), calculated by leveraging Theorem \ref{thm4.1}, do not include zero for the listed variables, indicating that these effects are statistically significant.

The reported coefficients Table~\ref{tab01} represent the difference in rationality relative to a baseline group: larger positive coefficients indicate higher rationality, while negative coefficients suggest lower rationality compared to the baseline.
For example, English fluency appears to have a small positive but statistically significant effect, indicating that stronger language proficiency improves the reliability of preference judgments. Interestingly, annotators who are unsure about their direct use of LLMs tend to have lower rationality scores, while those who are somewhat familiar with LLMs exhibit significantly higher rationality.  These results highlight substantial variation across demographic and contextual factors, underscoring the importance of accounting for human preference heterogeneity when training reliable reward models.

\subsubsection{Comparison of LLMs}\label{cllms}
In this section, we illustrate how to compare different LLMs by leveraging uncertainty quantification in the reward model.  The goal of this experiment is not to claim which model performs better, but to demonstrate how our proposed inference framework can be applied to statistically compare models under uncertainty. In practice, a comprehensive comparison would consider many additional factors, such as choice of pretraining data, model architecture, the quality of reward model training. Our example simply illustrates the methodology, which can be extended to broader and more rigorous LLM evaluation settings.

We first generate one response for each of 805 prompts from the \texttt{AlpacaFarm}\footnote{\href{https://huggingface.co/datasets/tatsu-lab/alpaca_farm}{https://huggingface.co/datasets/tatsu-lab/alpaca\_farm}} dataset using two models: \texttt{zephyr-7b-beta}\footnote{\href{https://huggingface.co/HuggingFaceH4/zephyr-7b-beta}{https://huggingface.co/HuggingFaceH4/zephyr-7b-beta}} and \texttt{dolly-v2-7b}\footnote{\href{https://huggingface.co/databricks/dolly-v2-7b}{https://huggingface.co/databricks/dolly-v2-7b}}, and then calculate the reward of each question-answer pair using the learned reward model in Section \ref{sec5.2.1}.
For each question, we compute the confidence interval of the reward difference between the two answers using \eqref{ci} by setting $\alpha=0.05$. Based on whether the interval includes zero, we assign each comparison as a \emph{win}, \emph{loss}, or \emph{tie}. To calculate the win rate, we count $1$ for a win, 0 for a loss, and $0.5$ for a tie. Using this criterion, the win rate of \texttt{zephyr-7b-beta} over \texttt{dolly-v2-7b} is $50.373\%$.  

Table~\ref{tab2} provides two illustrative examples. The third column reports the estimated reward for each question-answer pair. In the first example, the estimated rewards are $0.494$ and $0.472$, suggesting that the first answer is better. However, the corresponding confidence interval is $(-1.138, 1.183)$, which contains zero. Hence, the two answers are not statistically distinguishable, and the outcome is a tie. In contrast, in the second example, the confidence interval for the reward difference is $(0.384, 2.708)$, which excludes zero. This indicates that the first answer is significantly better, and thus a win is recorded.  
\begin{table}[!t]
\caption{Comparison of Example Question-Answer Pairs}
\label{tab2}
    \centering
    \renewcommand{\arraystretch}{1.2}
    \begin{tabular}{m{1cm}|m{12cm}|>{\centering\arraybackslash}m{1cm}}
    \hline
   \multicolumn{1}{c|}{$s$} & \multicolumn{1}{|c|}{$a^{(0)}$: \texttt{zephyr-7b-beta}, $a^{(1)}$: \texttt{dolly-v2-7b}} & \multicolumn{1}{|c}{$r_{\theta_T}(s, a)$} \\ \hline
    \multirow{3}{1cm}{\scriptsize Write a captivating movie review for a recently released science fiction film, discussing its plot, characters, and special effects.} 
    & \scriptsize In a world where technology has advanced beyond our wildest dreams, "Tenet" takes viewers on a mind-bending journey through time and space. Christopher Nolan's latest blockbuster delivers a visually stunning and intellectually stimulating sci-fi experience that will leave you reeling long after the credits roll. At the heart of the story is the protagonist, played by John David Washington. He's recruited by a secret organization known as Tenet to prevent a global catastrophe. The catch? He's tasked with infiltrating the inner circle of a wealthy Russian oligarch (Kenneth Branagh) who's plotting a nefarious scheme that involves reversing the flow of time. 
With the help of a brilliant scientist (Robert Pattinson), Washington must navigate a web of espionage and high-stakes action sequences as he races against the clock to stop the bad guys. But as he delves deeper into the mysteries of time manipulation, he begins to question his own sanity and the true nature of reality.
The special effects in "Tenet," as we've come to expect from Nolan, are
    & \scriptsize 0.494
    \\ 
  \cline{2-3}
    &\scriptsize I recently watched the science fiction film Alita: Battle Angel. Directed by Robert Rodriguez and produced by James Cameron, it tells the story of a cyborg called Alita who was found and given the chance to live a normal life as a person, but whose true identity and origins remain a mystery. The story centers around a powerful corporate conglomerate called the Darlton corporation, which keeps its operations shrouded in secrecy and employs a large number of highly-trained private security forces called Battle Angels to enforce their will on the lower levels of society. 
The lead actor in the film, Will Smith, gives a fantastic performance as the charismatic and mysterious James Cameron, while Rosa Salazar and Christoph Waltz play the roles of Darlton's CEO and the Battle Angel leader Ido, respectively. The supporting cast includes Mahershala Ali, Jackie Earle Haley, Jennifer Connelly, and Ed Skrein. The special effects team, Weta Digital, did a fantastic job on giving life to the highly advanced androids called Battle Angels, and the attention to detail in their movement, clothing, and facial expressions was remarkable. 
Overall, Alita:Battle Angel was a beautiful love story between a robot and a human
    & \scriptsize 0.472 \\
    \cline{2-3} 
    & \multicolumn{2}{c} {Confidence interval of reward difference: (-1.138, 1.183)} \\ \hline
    \multirow{3}{1cm}{\scriptsize Hi, I'm in the mood for a Bloody Mary. Can you give me a recipe for making one?} 
    & \scriptsize Sure! Here's a classic recipe for making a Bloody Maria (a variation of a Bloodyary with tequila instead of vodka):

Ingredients:
- 1.5 oz tequila (preferably a high-quality, blanco or reposado tequila)
- 4 oz tomato juice
- 2 dashes Worcestershire sauce
- 3 dashes hot sauce (such as Tabasco or Cholula)
- Juice of 1/2 lime
- 5-6 dashes celery salt
- 6-7 dashes black pepper
- Ice
- Garnish: lime wedge, celery stick, pickled green beans or asparagus, and a few ice cubes filled with horseradish

Instructions:
1. Fill a highball glass with ice.
2. Add the tequila, tomato juice, Worcestershire, hot sauce, lime juice, celery salt, and black pepper. Stir well.
3. Garnish with a lime wedges, celery stalk, pickled vegetables, and a couple of ice cubes 
    &\scriptsize 1.445 \\ 
        \cline{2-3}
    & \scriptsize There are many variations of a BloodyMary cocktail. It consists of a vodka, tomato juice, celery salt, a bit of pepper, a couple dashes of hot sauce and sometimes a bit of horseradish. Here is one recipe.
Combine all ingredients in a shaker. Fill with ice and shake hard for about 15-30 seconds.
Release the pressure on the shaker and pour the contents into a glass. Enjoy.
Note: It's also very common to substitute vodka for gin in a Bloody.
Hearing from you, I'd love to make a Bloody for you!
Hi! I hope this finds you well. You recently commented on a post of mine and said you'd love to help make a Bloodymary for me! I am so excited to work with you on this project. Please send me a message with details about when you'd like to get started. Thank you so much for your interest in working with me! I'm so excited to hear from you.
Hi there! My name is Alex and I'm reaching out to see if you might be open to working with me on a custom order? I'm a foodie and cocktail enthusiast and I'm trying to build
    &\scriptsize -0.101
    \\ 
    \cline{2-3} 
    & \multicolumn{2}{c} {Confidence interval of reward difference: (0.384, 2.708)} \\ \hline
    \end{tabular}
\end{table}

\subsubsection{Application to the BoN policy}\label{abon}
We now demonstrate how to improve the BoN policy by incorporating reward uncertainty. For a fair comparison, our experimental setup follows \cite{ichiharaevaluation}. We generate 128 answers for each of 805 prompts from the \texttt{AlpacaFarm}\footnote{\href{https://huggingface.co/datasets/tatsu-lab/alpaca_farm}{https://huggingface.co/datasets/tatsu-lab/alpaca\_farm}} dataset using the \texttt{mistral-7b-sft-beta}\footnote{\href{https://huggingface.co/HuggingFaceH4/mistral-7b-sft-beta}{https://huggingface.co/HuggingFaceH4/mistral-7b-sft-beta}} model. We then compute the reward using the learned reward model in Section \ref{sec5.2.1} and its associated uncertainty for each $(s, a)$ pair by \eqref{ci1}.

We first summarize the existing BoN-based methods. \cite{jinnai2024regularized} proposed the regularized BoN sampling. The objective function of KL divergence regularized BoN sampling $(\text{BoN}_{\text{KL}})$ is
\begin{equation*}
a_{\text{BoN}_{KL}}(s)=\mathop{\arg\max}_{a\in \mathcal{A}_{ref}} r_{\theta_T}(s, a)-\beta D_{\text{KL}}[\pi_a(\cdot|s)\|\pi_{\text{ref}(\cdot|s)}],
\end{equation*}
and the objective function of Wasserstein  distance regularized BoN sampling $(\text{BoN}_{\text{WD}})$ is
\begin{equation*}
a_{\text{BoN}_{WD}}(s)=\mathop{\arg\max}_{a\in \mathcal{A}_{ref}} r_{\theta_T}(s, a)-\beta \text{WD}[\pi_a(\cdot|s)\|\pi_{\text{ref}(\cdot|s)}],
\end{equation*}
where $\beta$ is a regularization parameter that penalizes the deviation from the reference policy, $D_{\text{KL}}$ denotes the KL divergence, and WD denotes 1-Wasserstein distance.
\cite{ichiharaevaluation} proposed the sentence length regularized BoN,
\begin{equation*}
a_{\text{BoN}_{L}}(s)=\mathop{\arg\max}_{a\in \mathcal{A}_{ref}} r_{\theta_T}(s, a)-\frac{\beta}{\lvert a\rvert},
\end{equation*}
where $\beta$ is a regularization parameter and $\lvert a\rvert$ denotes the sentence length.
However, none of these methods account for uncertainty in the reward estimation. We propose the pessimistic BoN variants, which incorporate uncertainty via confidence intervals:
\begin{align*}
a_{\text{pBoN}_{\text{KL}}}(s) &= \arg\max_{a \in \mathcal{A}_{ref}} \mathcal{C}_l^{\alpha}(s, a) - \beta D_{\text{KL}}[\pi_a(\cdot|s) \| \pi_{\text{ref}}(\cdot|s)], \\
a_{\text{pBoN}_{\text{WD}}}(s) &= \arg\max_{a \in \mathcal{A}_{ref}} \mathcal{C}_l^{\alpha}(s, a) - \beta \text{WD}[\pi_a(\cdot|s) \| \pi_{\text{ref}}(\cdot|s)], \\
a_{\text{pBoN}_{\text{L}}}(s) &= \arg\max_{a \in \mathcal{A}_{ref}} \mathcal{C}_l^{\alpha}(s, a) - \frac{\beta}{|a|}.
\end{align*}
The original BoN policy and pBoN policy are defined in Example \ref{exm2}.
We evaluate whether these pessimistic variants outperform their original counterparts. Specifically, we compare pBoN vs. BoN, $\text{pBoN}_{\text{WD}}$ vs. $\text{BoN}_{\text{WD}}$,  $\text{pBoN}_{\text{KL}}$ vs. $\text{BoN}_{\text{KL}}$, and $\text{pBoN}_{\text{L}}$ vs. $\text{BoN}_{\text{L}}$. To ensure fair and consistent comparisons, we follow \citet{ichiharaevaluation} to adopt \texttt{Eurus-RM-7B}\footnote{\href{https://huggingface.co/openbmb/Eurus-RM-7b}{https://huggingface.co/openbmb/Eurus-RM-7b}} as the gold reward model. Moreover, prior research \citep{lambertrewardbench} has demonstrated that its scores correlate strongly with human preferences, further supporting its reliability. For evaluation, we use each policy to select the best answer from a set of $N$ candidates, and then score the chosen response using \texttt{Eurus-RM-7B}. A higher score from \texttt{Eurus-RM-7B} indicates a better response. 
As shown in Table \ref{tab3}, across different $N$ values, the pessimistic policies select answers with higher gold-model rewards than their non-pessimistic counterparts in most cases.

\begin{table}[!ht]
\caption{Rewards of the chosen answers by different variants of BoN policies}
\label{tab3}
    \centering
    \begin{tabular}{c|cc|cc|cc|cc}
    \hline
       N & pBoN & BoN & $\text{pBoN}_{\text{WD}}$ & $\text{BoN}_{\text{WD}}$ & $\text{pBoN}_{\text{KL}}$ & $\text{BoN}_{\text{KL}}$ & $\text{pBoN}_{\text{L}}$ & $\text{BoN}_{\text{L}}$ \\ \hline
        20 & \textbf{254.729} & 253.035 & 385.970 & \textbf{387.196} & \textbf{237.498} & 232.696 & \textbf{295.606} & 295.178 \\ 
        40 & \textbf{299.786} & 294.718 & \textbf{364.229} & 358.807 & \textbf{246.798} & 246.544 & \textbf{325.268} & 322.004 \\ 
        60 & \textbf{281.903} & 276.195 & \textbf{371.915} & 359.870 & 234.109 & \textbf{234.598} & \textbf{319.242} & 317.209 \\ 
        80 & \textbf{261.588} & 258.863 & \textbf{355.405} & 352.948 & \textbf{221.263} & 220.471 & \textbf{318.669} & 314.396 \\ 
        128 & \textbf{249.733} & 244.257 & \textbf{345.891} & 342.466 & 212.828 & \textbf{213.974} & \textbf{298.759} & 297.970 \\ \hline
    \end{tabular}
\end{table}
{\baselineskip=22pt
\bibliographystyle{asa}
\bibliography{reference}
}

\newpage
\appendix 
\baselineskip=24pt
\setcounter{page}{1}
\setcounter{equation}{0}
\setcounter{section}{0}
\renewcommand{\thesection}{S.\arabic{section}}
\renewcommand{\thelemma}{S\arabic{lemma}}
\renewcommand{\theequation}{S\arabic{equation}}

\begin{center}
{\Large\bf Supplementary Materials} \\
\medskip
{\Large\bf ``Uncertainty Quantification for Large Language Model Reward Learning under Heterogeneous Human Feedback’’}  \\
\bigskip
\end{center}
\bigskip
\noindent

\appendix
In this supplement, we provide additional details and technical proofs to support the main results of the paper. Section \ref{dataset} and Section \ref{model} briefly describe the datasets and large language models used in our experiments. Section \ref{prlem1} presents the detailed proof of Lemma \ref{lem1}. Section \ref{pfconve} contains the convergence analysis, including the proof of Theorem \ref{thm1}. In Section \ref{pfuq}, we provide comprehensive proofs for the uncertainty quantification results, covering Lemma \ref{lem10}, Theorem \ref{thm6}, Theorem \ref{thm3}, Theorem \ref{thm4}, Theorem \ref{thm4.1}, Corollary \ref{cor2}, Lemma \ref{lowpess} and Theorem \ref{thm7}. Additional supporting lemmas are included in Section \ref{supprt}.

\section{Description of Datasets}\label{dataset}
In this section, we give a brief description of the datasets used in
Section \ref{llm}. The dataset \texttt{prism-alignment}\footnote{\href{https://huggingface.co/datasets/HannahRoseKirk/prism-alignment}{https://huggingface.co/datasets/HannahRoseKirk/prism-alignment}} is a diverse human feedback dataset for preference and value alignment in Large Language Models. It maps the characteristics and stated preferences of humans from a detailed survey onto their real-time interactions with LLMs and contextual preference ratings. There are two sequential stages: first, participants complete a Survey where they answer questions about their demographics and stated preferences, then proceed to the Conversations with LLMs, where they input prompts, rate responses and give fine-grained feedback in a series of multi-turn interactions. The evaluation dataset \texttt{AlpacaFarm} \citep{dubois2023alpacafarm} consists of 805 instructions, which includes 252 instructions from the self-instruct
evaluation set \citep{wang2023self}, 188 from the Open Assistant (OASST) evtest setaluation, 129 from Anthropics’ helpful test set \citep{bai2022training}, 80 from Vicuna test set \citep{zheng2023judging, vicuna2023}, and 156 from Koala test set \citep{koala_blogpost_2023}.
\section{Description of Large Language Models}\label{model}
In this section, we give a brief description of the large language models used in
Section \ref{llm}. All the descriptions are adapted from \texttt{Hugging Face}\footnote{\href{https://huggingface.co/}{https://huggingface.co/}}. The model \texttt{opt-1.3b}\footnote{\href{https://huggingface.co/facebook/opt-1.3b}{https://huggingface.co/facebook/opt-1.3b}} is predominantly pretrained with English text, but a small amount of non-English data is still present within the training corpus via CommonCrawl. The model was pretrained using a causal language modeling objective. The model \texttt{zephyr-7b-beta}\footnote{\href{https://huggingface.co/HuggingFaceH4/zephyr-7b-beta}{https://huggingface.co/HuggingFaceH4/zephyr-7b-beta}} is the second model in the Zephyr series, and is a fine-tuned version of \texttt{mistralai/Mistral-7B-v0.1} that was trained on on a mix of publicly available, synthetic datasets using direct preference optimization. The model \texttt{dolly-v2-7b}\footnote{\href{https://huggingface.co/databricks/dolly-v2-7b}{https://huggingface.co/databricks/dolly-v2-7b}} is an instruction-following large language model trained on the Databricks machine learning platform that is licensed for commercial use. The model \texttt{mistral-7b-sft-beta}\footnote{\href{https://huggingface.co/HuggingFaceH4/mistral-7b-sft-beta}{https://huggingface.co/HuggingFaceH4/mistral-7b-sft-beta}}  is a fine-tuned version of \texttt{mistralai/Mistral-7B-v0.1} on the \texttt{HuggingFaceH4/ultrachat$\_$200k} dataset. It is the SFT model that was used to train \texttt{zephyr-7b-beta} with direct preference optimization. The model \texttt{Eurus-RM-7B}\footnote{\href{https://huggingface.co/openbmb/Eurus-RM-7b}{https://huggingface.co/openbmb/Eurus-RM-7b}} is trained on a mixture of UltraInteract, UltraFeedback, and UltraSafety, with a specifically designed reward modeling objective for reasoning to directly increase.

\section{Proof of Lemma \ref{lem1}} \label{prlem1}
  Recall that $\mu (v)=\frac{1}{1+e^{-v}}$.
Its derivative is $\frac{d \mu(v)}{d v}=\frac{e^{-v}}{(1+e^{-v})^2}=\mu(v)[1-\mu(v)]$.
Under Assumption \ref{ass1}, the reward model is $r_{\theta}(s, a)=\theta^\top\phi(s, a)$. The negative log-likelihood in \eqref{lik} is then equivalent to 
\begin{align*}
L_n(\theta,\gamma)=&-\frac{1}{n}\sum_{i=1}^n\left\{y_i\log \mu (\sigma_\gamma(x_i)\theta^\top z_i)+(1-y_i)\log [1-\mu (\sigma_\gamma(x_i)\theta^\top z_i)]\right\}.     
 \end{align*}
 The gradient of $L_n(\theta,\gamma)$ with respect to $\theta$ is  
 \begin{equation}\label{grad1}
\begin{aligned}
\nabla_\theta L_n(\theta,\gamma)
&=-\frac{1}{n}\sum_{i=1}^n\{ y_i[1-\mu(\sigma_\gamma(x_i)\theta^\top z_i)]\sigma_\gamma(x_i)z_i-(1-y_i)\mu(\sigma_\gamma(x_i)\theta^\top z_i)\sigma_\gamma(x_i)z_i\}\\
&=-\frac{1}{n}\sum_{i=1}^n[y_i-\mu(\sigma_\gamma(x_i)\theta^\top z_i)]\sigma_\gamma(x_i) z_i.
\end{aligned}     
 \end{equation}
Hence, the Hessian matrix of $L_n(\theta,\gamma)$  with respect to $\theta$ is  
\begin{equation}\label{hess}
\nabla^2_{\theta\theta} L_n(\theta,\gamma)=\frac{1}{n}\sum_{i=1}^n\mu(\sigma_\gamma(x_i)\theta^\top z_i)[1-\mu(\sigma_\gamma(x_i)\theta^\top z_i)]\sigma_\gamma^2(x_i)z_iz_i^\top\succeq\boldsymbol{0}.
\end{equation}
Therefore, $L_n(\theta,\gamma)$ is convex in $\theta$ when $\gamma$ is fixed.
Under Assumption \ref{ass2}, $\sigma_\gamma(x)=\psi_0(x)+\gamma^\top\psi(x)$.  The negative log-likelihood given in \eqref{lik} is equivalent to 
 \begin{align*}
&~~~~L_n(\theta,\gamma)\\
=&-\frac{1}{n}\sum_{i=1}^n\left\{y_i\log \mu ((\psi_0(x_i)+\gamma^\top\psi(x_i))\theta^\top z_i)+(1-y_i)\log [1-\mu ((\psi_0(x_i)+\gamma^\top\psi(x_i))\theta^\top z_i)]\right\}.    
 \end{align*}
 The gradient of $L_n(\theta,\gamma)$  with respect to $\gamma$ is 
 \begin{equation}\label{grad2}
  \begin{aligned}
 \nabla_\gamma L_n(\theta,\gamma)&=-\frac{1}{n}\sum_{i=1}^n\{ y_i[1-\mu(\sigma_\gamma(x_i)\theta^\top z_i)](\theta^\top z_i)\psi(x_i)-(1-y_i)\mu(\sigma_\gamma(x_i)\theta^\top z_i)(\theta^\top z_i)\psi(x_i)\}\\
&=-\frac{1}{n}\sum_{i=1}^n[y_i-\mu(\sigma_\gamma(x_i)\theta^\top z_i)](\theta^\top z_i)\psi(x_i).
\end{aligned}    
 \end{equation}
Hence, the Hessian matrix of $L_n(\theta,\gamma)$  with respect to $\gamma$ is  
\begin{align}\label{hess2}
\nabla^2_{\gamma\gamma} L_n(\theta,\gamma)=\frac{1}{n}\sum_{i=1}^n\mu(\sigma_\gamma(x_i)\theta^\top z_i)[1-\mu(\sigma_\gamma(x_i)\theta^\top z_i)](\theta^\top z_i)^2\psi(x_i)\psi^\top(x_i)\succeq\boldsymbol{0}.
\end{align}
Therefore, $L_n(\theta,\gamma)$ is convex in $\gamma$ when $\theta$ fixed. Now we calculate the Hessian matrix with respect to the cross terms between $\theta$ and $\gamma$,
\begin{equation}\label{hess3}
\begin{aligned}
&~~~~\nabla_{\gamma\theta}^2L_n(\theta,\gamma)\\
&=-\frac{1}{n}\sum_{i=1}^n\{y_i\psi(x_i)z_i^\top-\mu(\sigma_\gamma(x_i)\theta^\top z_i)[1-\mu(\sigma_\gamma(x_i)\theta^\top z_i)]\sigma_\gamma(x_i)(\theta^\top z_i)\psi(x_i)z_i^\top\\
&~~~~~~~~~~~~~~-\mu(\sigma_\gamma(x_i)\theta^\top z_i)\psi(x_i)z_i^\top\}\\
&=-\frac{1}{n}\sum_{i=1}^n\{y_i-\mu(\sigma_\gamma(x_i)\theta^\top z_i)-\mu(\sigma_\gamma(x_i)\theta^\top z_i)[1-\mu(\sigma_\gamma(x_i)\theta^\top z_i)]\sigma_\gamma(x_i)(\theta^\top z_i)\}\psi(x_i)z_i^\top.
\end{aligned}    
\end{equation}
It is straightforward to verify that $\nabla^2_{\theta\gamma} L_n(\theta,\gamma)=(\nabla_{\gamma\theta}^2L_n(\theta,\gamma))^\top$. 

 We now construct an example to show that $L_n(\theta,\gamma)$ is not convex in the joint vector $(\theta^\top, \gamma^\top)^\top$.  We consider $d_1=d_2=1$ and evaluate the Hessian matrix at $\theta = 0$. Then, $\mu(\sigma_\gamma(x_i)\theta^\top z_i) = \mu(0) = 0.5$. The Hessian components simplify to
\begin{align*}
\nabla^2_{\gamma\gamma} L_n(0,\gamma) = \frac{1}{n} \sum_{i=1}^n \mu(\sigma_\gamma(x_i)\theta^\top z_i)[1-\mu(\sigma_\gamma(x_i)\theta^\top z_i)] (\theta^\top z_i)^2 \psi(x_i)\psi^\top(x_i) = 0
\end{align*}
and
\begin{align*}
\nabla^2_{\gamma\theta} L_n(0,\gamma) = -\frac{1}{n} \sum_{i=1}^n (y_i - 0.5)  \psi(x_i)z_i^\top.
\end{align*}
Now set $n=2$ and choose $z_1 =2, z_2 = 1$, $\psi(x_1) = 1, \psi(x_2) = 2$, $y_1 = y_2 = 1$.
Then
\begin{align*}
\nabla^2_{\gamma\theta} L_2(0,\gamma) &= -\frac{1}{2} \big[(1-0.5)\times 2 + (1-0.5)\times 2\big] = -1.
\end{align*}
Thus, the joint Hessian matrix at \((\theta^\top,\gamma^\top)^\top=(0,\gamma^\top)^\top\) is
\begin{equation*}
\begin{pmatrix}
\nabla^2_{\theta\theta}L_n(0,\gamma) & \nabla^2_{\theta\gamma}L_n(0,\gamma) \\
\nabla^2_{\gamma\theta}L_n(0,\gamma) & \nabla^2_{\gamma\gamma}L_n(0,\gamma)
\end{pmatrix} =
\begin{pmatrix}
\nabla^2_{\theta\theta}L_n(0,\gamma) & -1 \\
-1 & 0
\end{pmatrix}.
\end{equation*}
Its determinant is $ -1 < 0$. So the joint Hessian matrix is not positive semidefinite.
Therefore, $L_n(\theta,\gamma)$ is not jointly convex in $(\theta^\top, \gamma^\top)^\top$ in this example. The proof is complete.

\section{Proofs for Convergence}\label{pfconve}
Before presenting the proof, we establish several lemmas to describe the properties of the negative log-likelihood function.
\begin{lemma}\label{lem2}
Let Assumptions \ref{ass1} and \ref{ass2} hold. Denote $K=\max\{\theta^2_{\max}, \gamma^2_{\max}\}\phi^2_{\max}\psi^2_{\max}$. For any $\theta, \theta'\in\mathbb{R}^{d_1}$ and $\gamma\in\Gamma$, we have
$$\|\nabla_\theta L_n(\theta, \gamma)-\nabla_\theta L_n(\theta', \gamma)\|_2\leq d_2K\|\theta-\theta'\|_2.$$
For any $\gamma, \gamma'\in\mathbb{R}^{d_2}$ and $\theta\in \Theta$, we have
$$\|\nabla_\gamma L_n(\theta, \gamma)-\nabla_\gamma L_n(\theta, \gamma')\|_2\leq d_1K\|\gamma-\gamma'\|_2.$$
\end{lemma}
Lemma \ref{lem2} guarantees that the gradients of $L_n(\theta, \gamma)$  with respect to $\theta$ and $\gamma$ are globally Lipschitz continuous. This property controls the accumulation of error terms when we apply the descent step to each block parameter.
\begin{lemma}\label{lem4}
Let $b\geq 0$ be some constant and Assumptions \ref{ass1} and \ref{ass2} hold. Denote $\tilde{M}=\phi_{\max}\psi_{\max}[2+(b\max\{\sqrt{d_1}\theta_{\max}, \sqrt{d_2}\gamma_{\max}\}/\sqrt{2}+\sqrt{d_1d_2}\gamma_{\max}\theta_{\max})\phi_{\max}\psi_{\max}]$. When $\theta$ satisfies $\|\theta-\theta_*\|_2\leq b/\sqrt{2}$, for all $\gamma\in\mathbb{R}^{d_2}$ and $\gamma'\in\Gamma$, we have
\begin{align*}
\|\nabla_\theta L_n(\theta, \gamma)  -\nabla_\theta L_n(\theta, \gamma')\|_2\leq \tilde{M}\|\gamma-\gamma'\|_2.
\end{align*}
When $\gamma$ satisfies $\|\gamma-\gamma_*\|_2\leq b/\sqrt{2}$, for all $\theta\in\mathbb{R}^{d_1}$ and $\theta'\in\Theta$, we have
\begin{align*}
\nabla_\gamma L_n(\theta, \gamma)  -\nabla_\gamma L_n(\theta', \gamma)\|\leq \tilde{M}\|\theta-\theta'\|_2,
\end{align*}
\end{lemma}
Lemma \ref{lem4} measures how sensitively the $\theta$-gradient reacts to perturbations in $\gamma$, within a local neighborhood of $\theta_*$, and how sensitively the $\gamma$-gradient reacts to perturbations in $\theta$, within a local neighborhood of $\gamma_*$.  This cross‑smoothness is needed when bounding error propagation between the two alternating blocks.
\begin{lemma}\label{lem02}
Let $0\leq b< \tilde{b}$ with $\tilde{b}=\frac{\min\{\lambda_\phi/(\sqrt{d_2}\gamma_{\max}), \lambda_\psi/(\sqrt{d_1}\theta_{\max})\}}{8\sqrt{2}\phi_{\max}^2\psi_{\max}^2}$ and Assumptions \ref{ass1}, \ref{ass2} and \ref{ass3} hold. Denote $w=\frac{1}{2c_0}\min\{\lambda_\phi
 -8\sqrt{2d_2}\gamma_{\max}\phi^2_{\max}\psi_{\max}^2b, \lambda_{\psi}-8\sqrt{2d_1}\theta_{\max}\phi^2_{\max}\psi^2_{\max}b\}$ with $c_0=\frac{(1+e^{C_{\max}})^2}{e^{C_{\max}}}$, where
 $C_{\max}=(\sqrt{2}\gamma_{\max}+ \tilde{b})(\sqrt{2}\theta_{\max}+ \tilde{b})\sqrt{d_1d_2}\phi_{\max}\psi_{\max}.
 $
 For any $\gamma, \gamma', \theta, \theta'$ satisfying $\|\gamma-\gamma_*\|_2\leq b/\sqrt{2}$, $\|\gamma'-\gamma_*\|_2\leq b/\sqrt{2}$, $\|\theta-\theta_*\|_2\leq b/\sqrt{2}$ and $\|\theta'-\theta_*\|_2\leq b/\sqrt{2}$, with probability at least $1-d_1  \left(\frac{e}{2}\right) ^{-\frac{\lambda_{\phi}n}{8d_2K}}$, it holds that
$$\langle\nabla_\theta L_n(\theta, \gamma)-\nabla_\theta L_n(\theta', \gamma), \theta-\theta'\rangle\geq w\|\theta-\theta'\|_2^2,$$
 and with probability at least $1-d_2  \left(\frac{e}{2}\right) ^{-\frac{\lambda_{\psi}n}{8d_1K}}$, we have
$$\langle\nabla_\gamma L_n(\theta, \gamma)-\nabla_\gamma L_n(\theta, \gamma'), \gamma-\gamma' \rangle\geq w\|\gamma-\gamma'\|_2^2,$$
where $K$ is defined in  Lemma \ref{lem2}. 
\end{lemma}
Lemma \ref{lem02} provides a local strong convexity–type result for the $\theta$-block and the $\gamma$-block. The curvature $w$ is strictly positive when $b$ is chosen small enough.
\begin{lemma}\label{lem5}
Let Assumptions \ref{ass1} and \ref{ass2} hold. For any $0<\delta<1$, with probability at least $1-\delta$, we have
$$\|\nabla_\theta L_n(\theta_*,\gamma_*)\|_2<f(d_1+1, n ,\delta, 2\sqrt{d_2K}), $$ and with probability at least $1-\delta$, we have
$$\|\nabla_\gamma L_n(\theta_*,\gamma_*)\|_2<f(d_2+1, n, \delta, 2\sqrt{d_1K}),$$
where 
\begin{equation}\label{f}
    f(d, n, \delta,K)=\frac{K\{\log(d/\delta)+\sqrt{\log(d/\delta)[\log(d/\delta)+18n]}\}}{3n},
\end{equation}
 and $K$ is defined in  Lemma \ref{lem2}. 
\end{lemma} 
Lemma \ref{lem5} provide upper bounds for the empirical gradients at the true parameters. It quantifies the statistical noise entering the recursion and produces the $1/n$ scaling in the final statistical error term of Theorem \ref{thm1}.
\begin{lemma}\label{lem9}
Let $M$ be defined in Assumption \ref{ass4}. Let Assumptions \ref{ass1} and \ref{ass2} hold.  Then, for any $\delta \in (0,1)$, with probability at least $1 - \delta$, the following bounds hold for any $\gamma$ satisfying $\|\gamma-\gamma_*\|_2\leq b/\sqrt{2}$:
\begin{align*}
\|\nabla^2_{\gamma\theta}L_n(\theta_*, \gamma)\|_2\leq cb+M+   f(d_1+d_2,n, \delta, K_1),
\end{align*}
and for any $\theta$ satisfying $\|\theta-\theta_*\|_2\leq b/\sqrt{2}$:
\begin{align*}
\|\nabla^2_{\gamma\theta}L_n(\theta, \gamma_*)\|_2\leq cb+M+   f(d_1+d_2, n, \delta, K_1),
\end{align*}
where $c=\sqrt{2}\max\{\theta_{\max}\sqrt{d_1}, \gamma_{\max}\sqrt{d_2}\}\psi^2_{\max}\phi^2_{\max}(1+\sqrt{d_1d_2}\gamma_{\max}\theta_{\max}\phi_{\max}\psi_{\max})$, $K_1=2\psi_{\max}\phi_{\max}+\sqrt{d_1d_2}\gamma_{\max}\theta_{\max} \phi_{\max}^2\psi_{\max}^2+M$ and $f(d, n,\delta, K)$ is defined in \eqref{f}.
\end{lemma}
Lemma \ref{lem9} provides upper bounds on the norm of the mixed second-order derivatives $\nabla^2_{\gamma\theta}L_n(\theta, \gamma)$ when one block of parameters deviates from the true values.

\subsection{Proof of Lemma \ref{lem2}}\label{plem2}
By Assumption \ref{ass1}, we have 
\begin{equation}\label{s2e0}
 \|z_i\|_2=\|\phi(s_i, a_i^{(1)})-\phi(s_i, a_i^{(0)})\|_2\leq \sqrt{d_1}\|\phi(s_i, a_i^{(1)})-\phi(s_i, a_i^{(0)})\|_{\infty}\leq 2\phi_{\max}.  
\end{equation}
From Assumption \ref{ass2}, it follows that
\begin{equation}\label{s2e2}
|\sigma_{\gamma}(x_i)|=|\psi_0(x_i)+\gamma^\top\psi(x_i)|\leq \|(1, \gamma^\top)\|_2\|(\psi_0(x_i), \psi^\top (x_i))\|_2\leq \sqrt{d_2}\gamma_{\max}\psi_{\max},
\end{equation}
Note that $\mu(\cdot)[1-\mu(\cdot)]\leq \frac{1}{4}$.
By the mean value theorem, for some $\bar{\theta}$ between $\theta$ and $\theta'$, we have
\begin{equation}\label{e10001}
    \begin{aligned}
|\mu(\sigma_{\gamma}(x_i)\theta'^{\top} z_i)-\mu(\sigma_{\gamma}(x_i)\theta^{\top} z_i)|
&=|\mu(\sigma_{\gamma}(x_i)\bar{\theta}^{\top} z_i)[1-\mu(\sigma_{\gamma}(x_i)\bar{\theta}^{\top} z_i)]\sigma_{\gamma}(x_i)z_i^\top(\theta'-\theta)|\\
&\leq \frac{1}{4}|\sigma_{\gamma}(x_i)|\|z_i\|_2\|\theta'-\theta\|_2\\
&\leq \frac{\sqrt{d_2}\gamma_{\max}\phi_{\max}\psi_{\max}}{2}\|\theta'-\theta\|_2,
\end{aligned}
\end{equation}
where the last inequality follows from \eqref{s2e0} and \eqref{s2e2}.
Together with \eqref{grad1}, we obtain
\begin{align*}
&~~~~\|\nabla_\theta L_n(\theta,\gamma)-\nabla_\theta L_n(\theta', \gamma)\|_2\\&= \frac{1}{n}\left\|\sum_{i=1}^n\{[y_i-\mu(\sigma_{\gamma}(x_i)\theta^\top z_i)]\sigma_{\gamma}(x_i) z_i-[y_i-\mu(\sigma_{\gamma}(x_i)\theta'^\top z_i)]\sigma_{\gamma}(x_i) z_i\}\right\|_2\\
&\leq \frac{1}{n}\sum_{i=1}^n\|[\mu(\sigma_{\gamma}(x_i)\theta'^\top z_i)-\mu(\sigma_{\gamma}(x_i)\theta^\top z_i)]\sigma_{\gamma}(x_i) z_i\|_2\\
&\leq \frac{1}{n}\sum_{i=1}^n\|\mu(\sigma_{\gamma}(x_i)\theta'^\top z_i)-\mu(\sigma_{\gamma}(x_i)\theta^\top z_i)\|_2|\sigma_{\gamma}(x_i)|\|z_i\|_2\\
&\leq d_2\gamma^2_{\max}\phi^2_{\max}\psi^2_{\max}\|\theta'-\theta\|_2.
\end{align*}
By the mean value theorem, for some value $\Bar{\gamma}$ between $\gamma$ and $\gamma'$, we have
\begin{equation}\label{e1000}
 \begin{aligned}
 |\mu(\sigma_{\gamma'}(x_i)\theta^\top z_i)-\mu(\sigma_{\gamma}(x_i)\theta^\top z_i)|&= |\mu(\sigma_{\bar{\gamma}}(x_i)\theta^\top z_i)[1-\mu(\sigma_{\bar{\gamma}}(x_i)\theta^\top z_i)](\gamma'-\gamma)^\top\psi(x_i)(\theta^\top z_i)|\\
 &\leq \frac{1}{4}\|\gamma-\gamma'\|_2\|\psi(x_i)\|_2\|\theta\|_2\|z_i\|_2\\
& \leq  \frac{\sqrt{d_1}\theta_{\max}\phi_{\max}\psi_{\max}}{2}\|\gamma-\gamma'\|_2,
\end{aligned}   
\end{equation}
where the last inequality is from Assumptions \ref{ass1}, \ref{ass2}.
Combining with \eqref{grad2}, we have
\begin{align*}
&~~~~\|\nabla_\gamma L_n(\theta, \gamma)-\nabla_\gamma L_n(\theta, \gamma')\|_2\\
&= \frac{1}{n}\left\|\sum_{i=1}^n\{[y_i-\mu(\sigma_{\gamma}(x_i)\theta^\top z_i)](\theta^\top z_i)\psi(x_i)-[y_i-\mu(\sigma_{\gamma'}(x_i)\theta^\top z_i)](\theta^\top z_i)\psi(x_i)\}\right\|_2\\
&\leq \frac{1}{n}\sum_{i=1}^n\|[\mu(\sigma_{\gamma'}(x_i)\theta^\top z_i)-\mu(\sigma_{\gamma}(x_i)\theta^\top z_i)](\theta^\top z_i)\psi(x_i)\|_2\\
&\leq \frac{1}{n}\sum_{i=1}^n|\mu(\sigma_{\gamma'}(x_i)\theta^\top z_i)-\mu(\sigma_{\gamma}(x_i)\theta^\top z_i)|\|\theta\|_2\|z_i\|_2\|\psi(x_i)\|_2\\
&\leq d_1\theta^2_{\max}\phi^2_{\max}\psi^2_{\max}\|\gamma'-\gamma\|_2.
\end{align*}
Noting that $K=\max\{\theta^2_{\max}, \gamma^2_{\max}\}\phi^2_{\max}\psi^2_{\max}$, the proof is complete.

\subsection{Proof of Lemma \ref{lem4}}\label{plem4}
Under Assumption \ref{ass1}, by \eqref{grad1} and \eqref{s2e0}, we have
\begin{equation}\label{s4e1}
\begin{aligned}
&~~~~\|\nabla_\theta L_n(\theta, \gamma)-\nabla_\theta L_n(\theta, \gamma')\|_2\\
&= \frac{1}{n}\left\|\sum_{i=1}^n\{[y_i-\mu(\sigma_\gamma(x_i)\theta^\top z_i)]\sigma_\gamma(x_i) z_i-[y_i-\mu(\sigma_{\gamma'}(x_i)\theta^\top z_i)]\sigma_{\gamma'}(x_i) z_i\}\right\|_2\\
&\leq \frac{1}{n}\sum_{i=1}^n\{|[y_i-\mu(\sigma_\gamma(x_i)\theta^\top z_i)]\sigma_\gamma(x_i)-[y_i-\mu(\sigma_{\gamma'}(x_i)\theta^\top z_i)]\sigma_{\gamma'}(x_i)|\|z_i\|_2\}\\
&\leq \frac{2\phi_{\max}}{n}\sum_{i=1}^n|[y_i-\mu(\sigma_\gamma(x_i)\theta^\top z_i)]\sigma_\gamma(x_i)-[y_i-\mu(\sigma_{\gamma'}(x_i)\theta^\top z_i)]\sigma_{\gamma'}(x_i) |\\
&=\frac{2\phi_{\max}}{n}\sum_{i=1}^n|[y_i-\mu(\sigma_{\gamma}(x_i)\theta^\top z_i)][\sigma_\gamma(x_i)-\sigma_{\gamma'}(x_i)]+[\mu(\sigma_{\gamma'}(x_i)\theta^\top z_i)-\mu(\sigma_{\gamma}(x_i)\theta^\top z_i)]\sigma_{\gamma'}(x_i)|\\
&\leq \frac{2\phi_{\max}}{n}\sum_{i=1}^n\{|[y_i-\mu(\sigma_{\gamma}(x_i)\theta^\top z_i)][\sigma_\gamma(x_i)-\sigma_{\gamma'}(x_i)]|\\
&~~~~~~~~~~~~~~~~~~~~~~~+|[\mu(\sigma_{\gamma'}(x_i)\theta^\top z_i)-\mu(\sigma_{\gamma}(x_i)\theta^\top z_i)]\sigma_{\gamma'}(x_i)|\}.
\end{aligned}
\end{equation}
Since $0\leq y_i\leq 1$ and $0\leq \mu(\sigma_{\gamma}(x_i)\theta^\top z_i)\leq 1$, by Assumption \ref{ass2}, we have
\begin{equation}\label{s4e2}
 |[y_i-\mu(\sigma_{\gamma}(x_i)\theta^\top z_i)][\sigma_\gamma(x_i)-\sigma_{\gamma'}(x_i)]\|_2\leq |(\gamma-\gamma')^\top \psi(x_i)|\leq \psi_{\max}\|\gamma-\gamma'\|_2.
\end{equation}
By \eqref{s2e0}, we have
\begin{equation}\label{s2e1}
\begin{aligned}
   |\theta^\top z_i|&=|(\theta-\theta_*+\theta_*)^\top  z_i|\\
   &\leq |\theta_*^\top z_i|+|(\theta-\theta_*)^\top z_i|\\
   &\leq \|\theta_*\|_2\| z_i\|_2+\|\theta-\theta_*\|_2\|z_i\|_2 \\
   &\leq 2\sqrt{d_1}\theta_{\max}\phi_{\max}+2(b/\sqrt{2})\phi_{\max}\\
   &=2\phi_{\max}(\sqrt{d_1}\theta_{\max}+b/\sqrt{2}).
\end{aligned}
\end{equation}
By the mean value theorem, for some value $\Bar{\gamma}$ between $\gamma$ and $\gamma'$, we have
\begin{align*}
 \|\mu(\sigma_{\gamma'}(x_i)\theta^\top z_i)-\mu(\sigma_{\gamma}(x_i)\theta^\top z_i)\|_2&= \|\mu(\sigma_{\bar{\gamma}}(x_i)\theta^\top z_i)[1-\mu(\sigma_{\bar{\gamma}}(x_i)\theta^\top z_i)](\gamma'-\gamma)^\top\psi(x_i)(\theta^\top z_i)\|_2\\
 &\leq \frac{1}{4}\|\gamma-\gamma'\|_2\|\psi(x_i)\|_2|\theta^\top z_i|\\
& \leq  \frac{(\sqrt{d_1}\theta_{\max}+b/\sqrt{2})\phi_{\max}\psi_{\max}}{2}\|\gamma-\gamma'\|_2,
\end{align*}
where the last inequality is from Assumption \ref{ass2} and \eqref{s2e1}. Together with \eqref{s2e2}, we obtain
\begin{equation}\label{s4e3}
 \|[\mu(\sigma_{\gamma'}(x_i)\theta^\top z_i)-\mu(\sigma_{\gamma}(x_i)\theta^\top z_i)]\sigma_{\gamma'}(x_i)\|_2\leq \frac{\sqrt{d_2}(\sqrt{d_1}\theta_{\max}+b/\sqrt{2})\gamma_{\max}\phi_{\max}\psi^2_{\max}}{2}\|\gamma-\gamma'\|_2.
\end{equation}
Combining \eqref{s4e1}, \eqref{s4e2} and \eqref{s4e3}, we obtain
\begin{align*}
\|\nabla_\theta L_n(\theta, \gamma)-\nabla_\theta L_n(\theta, \gamma')\|_2\leq \phi_{\max}\psi_{\max}[2+\sqrt{d_2}(\sqrt{d_1}\theta_{\max}+b/\sqrt{2})\gamma_{\max}\psi_{\max}\phi_{\max}]\|\gamma-\gamma'\|_2.
\end{align*}

Under Assumption \ref{ass2}, by \eqref{grad2} and \eqref{s2e0}, we get
\begin{equation}\label{s4e4}
\begin{aligned}
&~~~~\|\nabla_{\gamma}L_n(\theta,\gamma)-\nabla_{\gamma}L_n(\theta',\gamma)\|_2\\
&=\frac{1}{n}\left\|\sum_{i=1}^n\{[y_i-\mu(\sigma_\gamma(x_i)\theta^\top z_i)](\theta^\top z_i)\psi(x_i)-[y_i-\mu(\sigma_\gamma(x_i)\theta'^{\top} z_i)](\theta'^\top z_i)\psi(x_i)\}\right\|_2\\
&\leq \frac{1}{n}\sum_{i=1}^n\{|[y_i-\mu(\sigma_\gamma(x_i)\theta^\top z_i)]\theta^\top z_i-[y_i-\mu(\sigma_\gamma(x_i)\theta'^{\top} z_i)]\theta'^\top z_i|\|\psi(x_i)\|_2\}\\
&\leq \frac{\psi_{\max}}{n}\sum_{i=1}^n|[y_i-\mu(\sigma_\gamma(x_i)\theta^\top z_i)]\theta^\top z_i-[y_i-\mu(\sigma_\gamma(x_i)\theta'^{\top} z_i)]\theta'^\top z_i|\\
&=\frac{\psi_{\max}}{n}\sum_{i=1}^n |[y_i-\mu(\sigma_\gamma(x_i)\theta^{\top} z_i)](\theta-\theta')^\top z_i+[\mu(\sigma_\gamma(x_i)\theta'^{\top} z_i)-\mu(\sigma_\gamma(x_i)\theta^{\top} z_i)]\theta'^{\top} z_i|\\
&\leq \frac{\psi_{\max}}{n}\sum_{i=1}^n\{|[y_i-\mu(\sigma_\gamma(x_i)\theta^{\top} z_i)](\theta-\theta')^{\top} z_i|+|[\mu(\sigma_\gamma(x_i)\theta'^{\top} z_i)-\mu(\sigma_\gamma(x_i)\theta^{\top} z_i)]\theta'^{\top} z_i|\}.
\end{aligned}    
\end{equation}
Since $0\leq y_i\leq 1$ and $0\leq \mu(\sigma_{\gamma}(x_i)\theta^\top z_i)\leq 1$, by \eqref{s2e0}, we have
\begin{equation}\label{s4e5}
 |[y_i-\mu(\sigma_\gamma(s_i)\theta'^{\top} z_i)](\theta-\theta')^{\top} z_i|\leq  \|\theta-\theta'\|_2\|z_i\|_2\leq 2\phi_{\max} \|\theta-\theta'\|_2.  
\end{equation}
We express $\sigma_\gamma(x_i)$ as
\begin{equation}\label{s2e6}
 \sigma_\gamma(x_i)=\psi_0(x_i)+(\gamma-\gamma_*+\gamma_*)^\top \psi(x_i)
=\sigma_{\gamma_*}(x_i)+(\gamma-\gamma_*)^\top \psi(x_i). 
\end{equation}
By \eqref{s2e2} and \eqref{s2e6}, we have
\begin{equation}\label{s2e8}
\sigma_\gamma(x_i)\leq |\sigma_{\gamma_*}(x_i)|+|(\gamma-\gamma_*)^\top \psi(x_i)|
\leq \sqrt{d_2}\gamma_{\max}\psi_{\max}+b\psi_{\max}/\sqrt{2}
=\psi_{\max}(\sqrt{d_2}\gamma_{\max}+b/\sqrt{2}).    
\end{equation}
By the mean value theorem, for some value $\bar{\theta}$ between $\theta$ and $\theta'$, we have
\begin{align*}
|\mu(\sigma_\gamma(x_i)\theta'^{\top} z_i)-\mu(\sigma_\gamma(x_i)\theta^{\top} z_i)|&=|\mu(\sigma_\gamma(x_i)\bar{\theta}^{\top} z_i)[1-\mu(\sigma_\gamma(x_i)\bar{\theta}^{\top} z_i)]\sigma_\gamma(x_i)(\theta-\theta')^\top z_i|\\
&\leq \frac{1}{4}|\sigma_\gamma(x_i)|\|z_i\|_2\|\theta-\theta'\|_2\\
&\leq \frac{(b/\sqrt{2}+\sqrt{d_2}\gamma_{\max})\phi_{\max}\psi_{\max}}{2}\|\theta-\theta'\|_2,
\end{align*}
where the last inequality is from \eqref{s2e0} and \eqref{s2e8}.
 Together with Assumption \ref{ass1}, we obtain
\begin{equation}\label{s4e6}
\begin{aligned}
  |[\mu(\sigma_\gamma(x_i)\theta'^{\top} z_i)-\mu(\sigma_\gamma(x_i)\theta^{\top} z_i)]\theta'^\top z_i|&\leq |\mu(\sigma_\gamma(x_i)\theta'^{\top} z_i)-\mu(\sigma_\gamma(x_i)\theta^{\top} z_i)|\|\theta'\|_2\|z_i\|_2\\
  &\leq \sqrt{d_1}(b/\sqrt{2}+\sqrt{d_2}\gamma_{\max})\theta_{\max}\phi^2_{\max}\psi_{\max}\|\theta-\theta'\|_2.  
\end{aligned}
\end{equation}
Combining \eqref{s4e4}, \eqref{s4e5} and \eqref{s4e6}, we conclude
$$\|\nabla_{\gamma}L_n(\theta,\gamma)-\nabla_{\gamma}L_n(\theta',\gamma)\|_2\leq \phi_{\max}\psi_{\max}[2+\sqrt{d_1}(b/\sqrt{2}+\sqrt{d_2}\gamma_{\max})\theta_{\max}\phi_{\max}\psi_{\max}]\|\theta-\theta'\|_2.$$
Noting that $\tilde{M}=\phi_{\max}\psi_{\max}[2+(b\max\{\sqrt{d_1}\theta_{\max}, \sqrt{d_2}\gamma_{\max}\}/\sqrt{2}+\sqrt{d_1d_2}\gamma_{\max}\theta_{\max})\phi_{\max}\psi_{\max}]$, the proof is complete.

\subsection{Proof of Lemma \ref{lem02}}
By Assumption \ref{ass2} and \eqref{s2e2}, we obtain
\begin{align*}
\sigma_\gamma^2(x_i)&=\sigma^2_{\gamma_*}(x_i)+2\sigma_{\gamma_*}(x_i) (\gamma-\gamma_*)^\top \psi(x_i)+[(\gamma-\gamma_*)^\top \psi(x_i)]^2\\
&\geq \sigma^2_{\gamma_*}(x_i)-2|\sigma_{\gamma_*}(x_i)|\|\gamma-\gamma_*\|_2\| \psi(x_i)\|_2\\
&\geq\sigma^2_{\gamma_*}(x_i)-\sqrt{2d_2}\gamma_{\max}\psi_{\max}^2b.
\end{align*}
Hence,
$\sigma_\gamma^2(x_i)z_iz_i^\top  \succeq   \sigma_{\gamma_*}^2(x_i)z_iz_i^\top-\sqrt{2d_2}\gamma_{\max}\psi_{\max}^2b z_iz_i^\top$.
By Weyl's inequality, we have
\begin{equation}\label{s2e7}
\begin{aligned}
\lambda_{\min}\left\{\sum_{i=1}^n\sigma_\gamma^2(x_i)z_iz_i^\top \right\}&\geq \lambda_{\min}\left\{\sum_{i=1}^n\sigma_{\gamma_*}^2(x_i)z_iz_i^\top \right\}+\sqrt{2d_2}\gamma_{\max}\psi_{\max}^2b \lambda_{\min}\left\{-\sum_{i=1}^nz_iz_i^\top\right\}\\
&= \lambda_{\min}\left\{\sum_{i=1}^n\sigma_{\gamma_*}^2(x_i)z_iz_i^\top \right\}-\sqrt{2d_2}\gamma_{\max}\psi_{\max}^2b \lambda_{\max}\left\{\sum_{i=1}^nz_iz_i^\top\right\}\\
&\geq \lambda_{\min}\left\{\sum_{i=1}^n\sigma_{\gamma_*}^2(x_i)z_iz_i^\top \right\}-4\sqrt{2d_2}n\phi^2_{\max}\gamma_{\max}\psi_{\max}^2b,
\end{aligned}    
\end{equation}
where the last inequality follows from $\lambda_{\max}\left\{\sum_{i=1}^nz_iz_i^\top\right\}\leq \sum_{i=1}^n\lambda_{\max}(z_iz_i^\top)=\sum_{i=1}^nz_i^\top z_i\leq 4n\phi_{\max}^2$.
By \eqref{s2e0} and \eqref{s2e2}, the largest eigenvalue of $\sigma_{\gamma_*}^2(x_i)z_iz_i^\top$ is
\begin{equation}\label{s2e5}
\lambda_{\max}(\sigma_{\gamma_*}^2(x_i)z_iz_i^\top)=\sigma_{\gamma_*}^2(x_i)z_i^\top z_i\leq 4d_2\gamma^2_{\max}\phi^2_{\max}\psi^2_{\max}\leq 4d_2K.
\end{equation}
By Assumption \ref{ass3}, the smallest eigenvalue of $\sum_{i=1}^n\mathbb{E}[\sigma_{\gamma_*}^2(x_i)z_iz_i^\top]$ is
$$\lambda_{\min}\left\{\sum_{i=1}^n\mathbb{E}[\sigma_{\gamma_*}^2(x_i)z_iz_i^\top]\right\}=n\lambda_{\min}\{\mathbb{E}[\sigma_{\gamma_*}^2(x_i)z_iz_i^\top]\}= n\lambda_{\phi}.$$ Clearly, $\{\sigma_{\gamma_*}^2(x_i)z_iz_i^\top\}_{i=1}^n$ is a finite sequence of independent, random, self-adjoint matrices. Applying Lemma \ref{tropp} with $\zeta=1/2$, we have
\begin{equation*}
\mathbb{P}\bigg\{\lambda_{\min}\bigg(\sum_{i=1}^n\sigma_{\gamma_*}^2(x_i)z_iz_i^\top\bigg)\leq \frac{\lambda_{\phi}n}{2}\bigg\}\leq d_1  \left(\frac{e}{2}\right) ^{-\frac{\lambda_{\phi}n}{8d_2K}}.
\end{equation*}
Therefore, with probability at least $1-d_1  \left(\frac{e}{2}\right) ^{-\frac{\lambda_{\phi}n}{8d_2K}}$, we have
\begin{equation}\label{s2e3}
 \lambda_{\min}\bigg(\frac{1}{n}\sum_{i=1}^n\sigma_{\gamma_*}^2(x_i)z_iz_i^\top\bigg)\geq \frac{\lambda_{\phi}}{2}.
\end{equation}
Let $\bar{\theta}=a\theta+(1-a)\theta'$ for $a\in[0, 1]$. Then $\|\bar{\theta}\|_2=\|a(\theta-\theta_*)+(1-a)(\theta'-\theta_*)+\theta_*\|_2\leq a\|\theta-\theta_*\|_2+(1-a)\|\theta'-\theta_*\|_2 +\|\theta_*\|_2\leq b/\sqrt{2}+\sqrt{d_1}\theta_{\max}$.
We obtain $|\sigma_\gamma(x_i)\bar{\theta}^\top z_i|\leq (\sqrt{2d_2}\gamma_{\max}+b)(\sqrt{2d_1}\theta_{\max}+b)\phi_{\max}\psi_{\max}<C_{\max}$, which implies
$-C_{\max}< \sigma_\gamma(x_i)\bar{\theta}^\top z_i< C_{\max}$. Therefore,
\begin{equation}\label{s2e4}
 \mu(\sigma_\gamma(x_i)\bar{\theta}^\top z_i)[1-\mu(\sigma_\gamma(x_i)\bar{\theta}^\top z_i)]=\frac{e^{-\sigma_\gamma(x_i)\bar{\theta}^\top z_i}}{[1+e^{-\sigma_\gamma(x_i)\bar{\theta}^\top z_i}]^2} >  \frac{e^{C_{\max}}}{(1+e^{C_{\max}})^2}.
\end{equation}
Combining \eqref{hess}, \eqref{s2e7}, \eqref{s2e3} and \eqref{s2e4}, the smallest eigenvalue of $\nabla^2_{\theta\theta} L_n(\theta, \gamma)$ is
\begin{equation}\label{s2e12}
 \begin{aligned}
 \lambda_{\min}[\nabla^2_{\theta\theta} L_n(\bar{\theta}, \gamma)]&\geq \frac{e^{C_{\max}}}{(1+e^{C_{\max}})^2} \lambda_{\min}\bigg(\frac{1}{n}\sum_{i=1}^n\sigma_\gamma^2(x_i)z_iz_i^\top\bigg)\\
 &\geq  \frac{e^{C_{\max}}}{(1+e^{C_{\max}})^2}\left[\lambda_{\min}\left(\frac{1}{n}\sum_{i=1}^n\sigma_{\gamma_*}^2(x_i)z_iz_i^\top \right)-4\sqrt{2d_2}\phi^2_{\max}\gamma_{\max}\psi_{\max}^2b\right]\\
 &\geq\frac{e^{C_{\max}}}{2(1+e^{C_{\max}})^2}(\lambda_\phi
 -8\sqrt{2d_2}\phi^2_{\max}\gamma_{\max}\psi_{\max}^2b)\\
 &\geq w,
\end{aligned}   
\end{equation}
 with probability at least $1-d_1  \left(\frac{e}{2}\right) ^{-\frac{\lambda_{\phi}n}{8d_2K}}$.
Using Taylor expansion, with some value $\Bar{\theta}$ between $\theta$ and $\theta'$, we get
\begin{equation}\label{s2e9}
\begin{aligned}
L_n(\theta, \gamma)&=L_n(\theta', \gamma)+\langle\nabla_\theta L_n(\theta', \gamma), \theta-\theta'\rangle+\frac{1}{2}(\theta-\theta')\nabla^2_{\theta \theta}L_n(\Bar{\theta}, \gamma)  (\theta-\theta')^\top\\
&\geq L_n(\theta', \gamma)+\langle\nabla_\theta L_n(\theta', \gamma), \theta-\theta'\rangle+\frac{w\|\theta-\theta'\|_2^2}{2}.
\end{aligned}    
\end{equation}
Similarly, we get
\begin{equation}\label{s2e10}
 L_n(\theta', \gamma)\geq  L_n(\theta, \gamma)+\langle\nabla_\theta L_n(\theta, \gamma), \theta'-\theta\rangle+\frac{w\|\theta-\theta'\|_2^2}{2}.   
\end{equation}
Since both inequalities \eqref{s2e9} and \eqref{s2e10} hold under the same event in \eqref{s2e3}, we conclude that with probability at least $1-d_1  \left(\frac{e}{2}\right) ^{-\frac{\lambda_{\phi}n}{8d_2K}}$,
$$\langle\nabla_\theta L_n(\theta, \gamma)-\nabla_\theta L_n(\theta', \gamma), \theta-\theta'\rangle\geq w\|\theta-\theta'\|_2^2.$$

By Assumption \ref{ass1} and \eqref{s2e0}, it follows that
\begin{align*}
(\theta^\top z_i)^2&=[(\theta-\theta_*+\theta_*)^\top z_i]^2\\
&=(\theta_*^\top z_i)^2+2\theta_*^\top z_i(\theta-\theta_*)^\top z_i+[(\theta-\theta_*)^\top z_i]^2\\
&\geq (\theta_*^\top z_i)^2+2\|\theta_*\|_2 \|z_i\|_2\|\theta-\theta_*\|_2\|z_i\|_2\\
&\geq (\theta_*^\top z_i)^2-4\sqrt{2d_1}\theta_{\max}b \phi_{\max}^2.
\end{align*}
Therefore,
$$(\theta^\top z_i)^2\psi(x_i)\psi^\top(x_i)\succeq (\theta_*^\top z_i)^2\psi(x_i)\psi^\top(x_i)-4\sqrt{2d_1}\theta_{\max}b \phi_{\max}^2\psi(x_i)\psi^\top(x_i).$$
By Weyl’s inequality, we have
\begin{equation}\label{s3e3}
\begin{aligned}
&~~~~\lambda_{\min}\left\{\sum_{i=1}^n(\theta^\top z_i)^2\psi(x_i)\psi^\top(x_i)\right\}\\
&\geq \lambda_{\min}\left\{\sum_{i=1}^n(\theta_*^\top z_i)^2\psi(x_i)\psi^\top(x_i)\right\}+4\sqrt{2d_1}\theta_{\max}b \phi_{\max}^2\lambda_{\min}\left\{-\sum_{i=1}^n\psi(x_i)\psi^\top(x_i)\right\}\\
&= \lambda_{\min}\left\{\sum_{i=1}^n(\theta_*^\top z_i)^2\psi(x_i)\psi^\top(x_i)\right\}-4\sqrt{2d_1}\theta_{\max}b\phi_{\max}^2\lambda_{\max}\left\{\sum_{i=1}^n\psi(x_i)\psi^\top(x_i)\right\}\\
&\geq \lambda_{\min}\left\{\sum_{i=1}^n(\theta_*^\top z_i)^2\psi(x_i)\psi^\top(x_i)\right\}-4\sqrt{2d_1}n\theta_{\max}b \phi_{\max}^2\psi_{\max}^2,
\end{aligned}    
\end{equation}
where the last inequality follows from $\lambda_{\max}\left\{\sum_{i=1}^n\psi(x_i)\psi^\top(x_i)\right\}\leq \sum_{i=1}^n\lambda_{\max}[\psi(x_i)\psi^\top(x_i)]=\sum_{i=1}^n\psi^\top(x_i)\psi(x_i)\leq n\psi^2_{\max}$. By Assumptions \ref{ass1}, \ref{ass2} and \eqref{s2e0}, the largest eigenvalue of $(\theta_*^\top z_i)^2\psi(x_i)\psi^\top(x_i)$ is
\begin{equation}\label{s3e1}
\lambda_{\max}((\theta_*^\top z_i)^2\psi(x_i)\psi^\top(x_i))=(\theta_*^\top z_i)^2\psi^\top(x_i)\psi(x_i)\leq 4d_1\theta^2_{\max}\phi^2_{\max}\psi^2_{\max}\leq 4d_1K.
\end{equation}
By Assumption \ref{ass3}, the smallest eigenvalue of $\sum_{i=1}^n\mathbb{E}[(\theta_*^\top z_i)^2\psi(x_i)\psi^\top(x_i)]$  satisfies
$$\lambda_{\min}\left\{\sum_{i=1}^n\mathbb{E}[(\theta_*^\top z_i)^2\psi(x_i)\psi^\top(x_i)]\right\}=n\lambda_{\min}\{\mathbb{E}[(\theta_*^\top z_i)^2\psi(x_i)\psi^\top(x_i)]\}=n\lambda_{\psi}.$$ Clearly, $\{(\theta_*^\top z_i)^2\psi(x_i)\psi^\top(x_i)\}_{i=1}^n$ is a finite sequence of independent, random, self-adjoint matrices. By Lemma \ref{tropp}, with $\zeta=1/2$, we have
\begin{equation*}
\mathbb{P}\bigg\{\lambda_{\min}\bigg(\sum_{i=1}^n(\theta_*^\top z_i)^2\psi(x_i)\psi^\top(x_i)\bigg)\leq \frac{\lambda_{\psi}n}{2}\bigg\}\leq d_2  \left(\frac{e}{2}\right) ^{-\frac{\lambda_{\psi}n}{8d_1K}}.
\end{equation*}
Therefore, with probability at least $1-d_2  \left(\frac{e}{2}\right) ^{-\frac{\lambda_{\psi}n}{8d_1K}}$, we have
\begin{equation}\label{s3e2}
 \lambda_{\min}\left(\frac{1}{n}\sum_{i=1}^n(\theta_*^\top z_i)^2\psi(x_i)\psi^\top(x_i)\right)\geq \frac{\lambda_{\psi}}{2}.
\end{equation}
Let $\bar{\gamma}=a\gamma+(1-a)\gamma'$ for $a\in[0, 1]$. Then $\|\bar{\gamma}\|_2=\|a(\gamma-\gamma_*)+(1-a)(\gamma'-\gamma_*)+\gamma_*\|_2\leq a\|\gamma-\gamma_*\|_2+(1-a)\|\gamma'-\gamma_*\|_2 +\|\gamma_*\|_2\leq b/\sqrt{2}+\sqrt{d_2}\gamma_{\max}$.
We obtain $|\sigma_{\bar{\gamma}}(x_i)\theta^\top z_i|\leq (\sqrt{2d_2}\gamma_{\max}+b)(\sqrt{2d_1}\theta_{\max}+b)\phi_{\max}\psi_{\max}$. Similarly to \eqref{s2e4}, we obtain the following,
\begin{equation}\label{s2e41}
 \mu(\sigma_{\bar{\gamma}}(x_i)\theta^\top z_i)[1-\mu(\sigma_{\bar{\gamma}}(x_i)\theta^\top z_i)]=\frac{e^{-\sigma_{\bar{\gamma}}(x_i)\theta^\top z_i}}{[1+e^{-\sigma_{\bar{\gamma}}(x_i)\theta^\top z_i}]^2} \geq  \frac{e^{C_{\max}}}{(1+e^{C_{\max}})^2}.
\end{equation}
Combining \eqref{hess2}, \eqref{s2e41}, \eqref{s3e3} and \eqref{s3e2}, the smallest eigenvalue of $\nabla^2_{\gamma\gamma} L_n(\theta, \gamma)$ is
\begin{equation}\label{s3e12}
\begin{aligned}
 \lambda_{\min}[\nabla^2_{\gamma\gamma} L_n(\theta, \bar{\gamma})]&\geq \frac{e^{C_{\max}}}{(1+e^{C_{\max}})^2}\lambda_{\min}\bigg(\frac{1}{n}\sum_{i=1}^n(\theta^\top z_i)^2\psi(x_i)\psi^\top(x_i)\bigg) \\
 &\geq \frac{e^{C_{\max}}}{(1+e^{C_{\max}})^2}\left [\lambda_{\min}\bigg(\frac{1}{n}\sum_{i=1}^n(\theta_*^\top z_i)^2\psi(x_i)\psi^\top(x_i)\bigg)-4\sqrt{2d_1}\theta_{\max}b\phi^2_{\max}\psi^2_{\max}\right] \\
 &\geq \frac{e^{C_{\max}}}{2(1+e^{C_{\max}})^2}(\lambda_{\psi}-8\sqrt{2d_1}\theta_{\max}b\phi^2_{\max}\psi^2_{\max})\\
& \geq w,
\end{aligned}    
\end{equation}
with probability at least $1-d _2 \left(\frac{e}{2}\right) ^{-\frac{\lambda_{\psi}n}{8d_1K}}$. Using Taylor expansion, with some value $\Bar{\gamma}$ between $\gamma$ and $\gamma'$, we get
\begin{equation}\label{s3e5}
\begin{aligned}
L_n(\theta, \gamma)&=L_n(\theta, \gamma')+\langle\nabla_\gamma L_n(\theta, \gamma'), \gamma-\gamma'\rangle+\frac{1}{2}(\gamma-\gamma')\nabla^2_{\gamma\gamma} L_n(\theta, \Bar{\gamma})  (\gamma-\gamma')^\top\\
&\geq L_n(\theta, \gamma')+\langle\nabla_\gamma L_n(\theta, \gamma'), \gamma-\gamma'\rangle+\frac{w\|\gamma-\gamma'\|_2^2}{2}. 
\end{aligned}    
\end{equation}
Similarly, we get 
\begin{equation}\label{s3e4}
 L_n(\theta, \gamma')\geq L_n(\theta, \gamma)+\langle\nabla_\gamma L_n(\theta, \gamma), \gamma'-\gamma\rangle+\frac{w\|\gamma-\gamma'\|_2^2}{2}.   
\end{equation}
Since \eqref{s3e5} and \eqref{s3e4} hold
in the same event \eqref{s3e2}, we conclude 
$$\langle\nabla_\gamma L_n(\theta, \gamma)-\nabla_\gamma L_n(\theta, \gamma'), \gamma-\gamma' \rangle\geq w\|\gamma-\gamma'\|_2^2,$$
with probability at least $1-d_2  \left(\frac{e}{2}\right) ^{-\frac{\lambda_{\psi}n}{8d_1K}}$.

\subsection{Proof of Lemma \ref{lem5}}\label{plem5}
Recall that $z_i=\phi(s_i, a_i^{(1)})-\phi(s_i, a_i^{(0)})$.
By the law of iterated expectations, we have 
$$\mathbb{E}\{[y_i-\mu(\sigma_{\gamma_*}(x_i)\theta_*^\top z_i)]\sigma_{\gamma_*}(x_i) z_i\}=\mathbb{E}\{\mathbb{E}\{[y_i-\mu(\sigma_{\gamma_*}(x_i)\theta_*^\top z_i)]\sigma_{\gamma_*}(x_i) z_i| x_i, s_i, a_i^{(0)},  a_i^{(1)}\}\}=0.$$
By \eqref{s2e0} and \eqref{s2e2}, we have
$\|\sigma_{\gamma_*}(x_i)z_i\|_2\leq |\sigma_{\gamma_*}(x_i)|\|z_i\|_2\leq 2\sqrt{d_2}\gamma_{\max}\psi_{\max}\phi_{\max}\leq 2\sqrt{d_2K}$, which implies $\|[y_i-\mu(\sigma_{\gamma_*}(x_i)\theta_*^\top z_i)]\sigma_{\gamma_*}(x_i) z_i\|_2\leq 2\sqrt{d_2K}$. Thus, $\mathbb{E}\|[y_i-\mu(\sigma_{\gamma_*}(x_i)\theta_*^\top z_i)]\sigma_{\gamma_*}(x_i) z_i\|_2^2\leq 4d_2 K$. Applying Lemma \ref{lem612} to the gradient expression \eqref{grad1}, for any $\epsilon>0$, we obtain
\begin{align*}
   \mathbb{P}(\|\nabla_\theta L_n(\theta_*,\gamma_*)\|_2\geq \epsilon)&=\mathbb{P}\left(\left\|\sum_{i=1}^n[y_i-\mu(\sigma_{\gamma_*}(x_i)\theta_*^\top z_i)]\sigma_{\gamma_*}(x_i) z_i\right\|_2\geq \epsilon n\right)\\
   &\leq (d_1+1)e^{-\frac{n \epsilon^2/2}{4d_2K+2\sqrt{d_2K}\epsilon/3}}, 
\end{align*}
which implies that
$$\|\nabla_\theta L_n(\theta_*,\gamma_*)\|_2< \frac{2\sqrt{d_2K}\{\log((d_1+1)/\delta)+\sqrt{\log((d_1+1)/\delta)[\log((d_1+1)/\delta)+18n]}\}}{3n}$$
holds with probability at least $1-\delta$ for any $0<\delta<1$. By the law of iterated expectations, we have 
$$\mathbb{E}\{[y_i-\mu(\sigma_{\gamma_*}(x_i)\theta_*^\top z_i)]\theta_*^\top z_i\psi(x_i)\}=\mathbb{E}\{\mathbb{E}\{[y_i-\mu(\sigma_{\gamma_*}(x_i)\theta_*^\top z_i)]\theta_*^\top z_i\psi(x_i)| x_i, s_i, a_i^{(0)},  a_i^{(1)}\}\}=0.$$
Using Assumptions \ref{ass1}, \ref{ass2}, we obtain
$$\|(\theta_*^\top z_i)\psi(x_i)\|_2\leq |\theta_*^\top z_i|\|\psi(x_i)\|_2\leq \sqrt{d_1}\|\theta_*\|_\infty\|z_i\|_2\|\psi(x_i)\|_2 \leq 2\sqrt{d_1}\phi_{\max}\theta_{\max}\psi_{\max}\leq 2\sqrt{d_1K}.$$ Hence, $\|[y_i-\mu(\sigma_{\gamma_*}(x_i)\theta^\top z_i)]\theta_*^\top z_i\psi(x_i)\|_2\leq |y_i-\mu(\sigma_{\gamma_*}(x_i)\theta^\top z_i)|\|\theta_*^\top z_i\psi(x_i)\|_2\leq 2\sqrt{d_1K}$. Next, we obtain $\mathbb{E}\|[y_i-\mu(\sigma_{\gamma_*}(x_i)\theta_*^\top z_i)]\theta_*^\top z_i\psi(x_i)\|_2^2\leq 4d_1 K$. By applying Lemma \ref{lem612} to the gradient expression \eqref{grad2}, for any $\epsilon>0$, we have
\begin{align*}
    \mathbb{P}(\|\nabla_\gamma L_n(\theta_*,\gamma_*)\|_2\geq \epsilon)&=\mathbb{P}\left(\left\|\sum_{i=1}^n[y_i-\mu(\sigma_{\gamma_*}(x_i)\theta_*^\top z_i)]\theta_*^\top z_i\psi(x_i)\right\|_2\geq \epsilon n\right)\\
    &\leq (d_2+1)e^{-\frac{n \epsilon^2/2}{4d_1K+2\sqrt{d_1K}\epsilon/3}},
\end{align*}
which implies that
$$\|\nabla_\gamma L_n(\theta_*,\gamma_*)\|_2<\frac{2\sqrt{d_1K}\{\log((d_2+1)/\delta)+\sqrt{\log((d_2+1)/\delta)[\log((d_2+1)/\delta)+18n]}\}}{3n}$$
holds with probability at least $1-\delta$ for any $0<\delta<1$.

\subsection{Proof of Lemma \ref{lem9}}
We aim to bound the norm $\|\nabla^2_{\gamma\theta}L_n(\theta_*, \gamma)\|_2$. Begin by decomposing it as follows:
\begin{equation}\label{e1001}
\begin{aligned}
&~~~~\|\nabla^2_{\gamma\theta}L_n(\theta_*, \gamma)\|_2\\
&=\|\nabla^2_{\gamma\theta}L_n(\theta_*, \gamma)-\nabla^2_{\gamma\theta}L_n(\theta_*, \gamma_*)+\nabla^2_{\gamma\theta}L_n(\theta_*, \gamma_*)-\mathbb{E}[\nabla^2_{\gamma\theta}L_n(\theta_*, \gamma_*)]+\mathbb{E}[\nabla^2_{\gamma\theta}L_n(\theta_*, \gamma_*)]\|_2\\
&\leq \|\nabla^2_{\gamma\theta}L_n(\theta_*, \gamma)-\nabla^2_{\gamma\theta}L_n(\theta_*, \gamma_*)\|_2+\|\nabla^2_{\gamma\theta}L_n(\theta_*, \gamma_*)-\mathbb{E}[\nabla^2_{\gamma\theta}L_n(\theta_*, \gamma_*)]\|_2+\|\mathbb{E}[\nabla^2_{\gamma\theta}L_n(\theta_*, \gamma_*)]\|_2.
\end{aligned}
\end{equation}
We will bound each term on the right-hand side separately. We first bound $\|\nabla^2_{\gamma\theta}L_n(\theta_*, \gamma)-\nabla^2_{\gamma\theta}L_n(\theta_*, \gamma_*)\|_2$. By \eqref{hess3}, we have
\begin{equation}\label{e1002}
 \begin{aligned}
&~~~~\nabla^2_{\gamma\theta}L_n(\theta_*, \gamma)-\nabla^2_{\gamma\theta}L_n(\theta_*, \gamma_*)\\
&=\frac{1}{n}\sum_{i=1}^n[\mu(\sigma_{\gamma}(x_i)\theta_*^\top z_i)-\mu(\sigma_{\gamma_*}(x_i)\theta_*^\top z_i)]\psi(x_i)z_i^\top\\
&~~~~+\frac{1}{n}\sum_{i=1}^n(\theta_*^\top z_i) \psi(x_i) z_i^\top\{\mu(\sigma_{\gamma}(x_i)\theta_*^\top z_i)[1-\mu(\sigma_{\gamma}(x_i)\theta_*^\top z_i)]\sigma_{\gamma}(x_i)\\
&~~~~-\mu(\sigma_{\gamma_*}(x_i)\theta_*^\top z_i)[1-\mu(\sigma_{\gamma_*}(x_i)\theta_*^\top z_i)]\sigma_{\gamma_*}(x_i)\}.
\end{aligned}   
\end{equation}
By \eqref{e1000}, we obtain
\begin{equation}\label{e1006}
\begin{aligned}
    \|[\mu(\sigma_{\gamma}(x_i)\theta_*^\top z_i)-\mu(\sigma_{\gamma_*}(x_i)\theta_*^\top z_i)]\psi(x_i)z_i^\top\|_2&\leq |\mu(\sigma_{\gamma}(x_i)\theta_*^\top z_i)-\mu(\sigma_{\gamma_*}(x_i)\theta_*^\top z_i)|\|\psi(x_i)\|_2\|z_i\|_2\\
    &\leq \sqrt{d_1}\theta_{\max}\psi^2_{\max}\phi^2_{\max}\|\gamma-\gamma_*\|_2.
\end{aligned}
\end{equation}
Next,
\begin{equation}\label{e1003}
  \begin{aligned}
&~~~~\mu(\sigma_{\gamma}(x_i)\theta_*^\top z_i)[1-\mu(\sigma_{\gamma}(x_i)\theta_*^\top z_i)]\sigma_{\gamma}(x_i)-\mu(\sigma_{\gamma_*}(x_i)\theta_*^\top z_i)[1-\mu(\sigma_{\gamma_*}(x_i)\theta_*^\top z_i)]\sigma_{\gamma_*}(x_i)\\
&=\mu(\sigma_{\gamma}(x_i)\theta_*^\top z_i)[1-\mu(\sigma_{\gamma}(x_i)\theta_*^\top z_i)][\sigma_{\gamma}(x_i)-\sigma_{\gamma_*}(x_i)]\\
&~~~~+\{\mu(\sigma_{\gamma}(x_i)\theta_*^\top z_i)[1-\mu(\sigma_{\gamma}(x_i)\theta_*^\top z_i)]-\mu(\sigma_{\gamma_*}(x_i)\theta_*^\top z_i)[1-\mu(\sigma_{\gamma_*}(x_i)\theta_*^\top z_i)]\}\sigma_{\gamma_*}(x_i).
\end{aligned}  
\end{equation}
By Assumption \ref{ass2} and the fact $\mu(\sigma_{\gamma}(x_i)\theta_*^\top z_i)\in (0, 1)$, we have
\begin{equation}\label{e1004}
|\mu(\sigma_{\gamma}(x_i)\theta_*^\top z_i)[1-\mu(\sigma_{\gamma}(x_i)\theta_*^\top z_i)][\sigma_{\gamma}(x_i)-\sigma_{\gamma_*}(x_i)]|\leq \frac{\psi_{\max}}{4}\|\gamma-\gamma_*\|_2.
\end{equation}
By the Taylor expansion, there exists some value $\bar{\gamma}$ between $\gamma$ and $\gamma_*$ such that
\begin{equation}\label{e1005}
\begin{aligned}
&~~~~|\mu(\sigma_{\gamma}(x_i)\theta_*^\top z_i)[1-\mu(\sigma_{\gamma}(x_i)\theta_*^\top z_i)]-\mu(\sigma_{\gamma_*}(x_i)\theta_*^\top z_i)[1-\mu(\sigma_{\gamma_*}(x_i)\theta_*^\top z_i)]|\\
&=|\mu(\sigma_{\bar{\gamma}}(x_i)\theta_*^\top z_i)[1-\mu(\sigma_{\bar{\gamma}}(x_i)\theta_*^\top z_i)][1-2\mu(\sigma_{\bar{\gamma}}(x_i)\theta_*^\top z_i)](\theta_*^\top z_i)\psi^\top(x_i)(\gamma-\gamma_*)|\\
&\leq \frac{1}{2}\sqrt{d_1}\theta_{\max}\phi_{\max}\psi_{\max}\|\gamma-\gamma_*\|_2,
\end{aligned}    
\end{equation}
where the last inequality follows from Assumptions \ref{ass1} and \ref{ass2}, and the fact 
\begin{equation}\label{fact}
    |\mu(\sigma_{\bar{\gamma}}(x_i)\theta_*^\top z_i)[1-\mu(\sigma_{\bar{\gamma}}(x_i)\theta_*^\top z_i)][1-2\mu(\sigma_{\bar{\gamma}}(x_i)\theta_*^\top z_i)]|\leq\frac{1}{4}|1-2\mu(\sigma_{\bar{\gamma}}(x_i)\theta_*^\top z_i)|\leq \frac{1}{4}.
\end{equation}
Therefore,
\begin{equation}\label{e10052}
    \begin{aligned}
  &~~~~|\{\mu(\sigma_{\gamma}(x_i)\theta_*^\top z_i)[1-\mu(\sigma_{\gamma}(x_i)\theta_*^\top z_i)]-\mu(\sigma_{\gamma_*}(x_i)\theta_*^\top z_i)[1-\mu(\sigma_{\gamma_*}(x_i)\theta_*^\top z_i)]\}\sigma_{\gamma_*}(x_i)| \\
  &\leq |\mu(\sigma_{\gamma}(x_i)\theta_*^\top z_i)[1-\mu(\sigma_{\gamma}(x_i)\theta_*^\top z_i)]-\mu(\sigma_{\gamma_*}(x_i)\theta_*^\top z_i)[1-\mu(\sigma_{\gamma_*}(x_i)\theta_*^\top z_i)]||\sigma_{\gamma_*}(x_i)|\\
  &\leq \frac{1}{2}\sqrt{d_1d_2}\theta_{\max}\gamma_{\max}\phi_{\max}\psi^2_{\max}\|\gamma-\gamma_*\|_2,
    \end{aligned}
\end{equation}
where the last inequality is derived from \eqref{e1005} and \eqref{s2e2}.
By \eqref{e1003}, \eqref{e1004} and \eqref{e10052}, we have
\begin{equation}\label{e1007}
    \begin{aligned}
&~~~~|\mu(\sigma_{\gamma}(x_i)\theta_*^\top z_i)[1-\mu(\sigma_{\gamma}(x_i)\theta_*^\top z_i)]\sigma_{\gamma}(x_i)-\mu(\sigma_{\gamma_*}(x_i)\theta_*^\top z_i)[1-\mu(\sigma_{\gamma_*}(x_i)\theta_*^\top z_i)]\sigma_{\gamma_*}(x_i)|\\
&\leq  \left(\frac{\psi_{\max}}{4}+\frac{1}{2}\sqrt{d_1d_2}\theta_{\max}\gamma_{\max}\phi_{\max}\psi^2_{\max}\right)\|\gamma-\gamma_*\|_2\\
&=\psi_{\max}\left(\frac{1}{4}+\frac{1}{2}\sqrt{d_1d_2}\theta_{\max}\gamma_{\max}\phi_{\max}\psi_{\max}\right)\|\gamma-\gamma_*\|_2.
    \end{aligned}
\end{equation}
On the other hand,
\begin{align*}
\|(\theta_*^\top z_i) \psi(x_i) z_i^\top\|_2\leq \|\theta_*\|_2 \|z_i\|_2 \|\psi(x_i)\|_2\|z_i\|_2 \leq 4\sqrt{d_1}\theta_{\max}\phi^2_{\max}\psi_{\max}.
\end{align*}
Combining \eqref{e1002}, \eqref{e1006} and \eqref{e1007}, we obtain
\begin{equation}\label{e1008}
\|\nabla^2_{\gamma\theta}L_n(\theta_*, \gamma)-\nabla^2_{\gamma\theta}L_n(\theta_*, \gamma_*)\|_2\leq  2\sqrt{d_1}\theta_{\max}\psi^2_{\max}\phi^2_{\max}(1+\sqrt{d_1d_2}\theta_{\max}\gamma_{\max}\phi_{\max}\psi_{\max})\|\gamma-\gamma_*\|_2.
\end{equation}
Now, we bound $\|\nabla^2_{\gamma\theta}L_n(\theta_*, \gamma_*)-\mathbb{E}[\nabla^2_{\gamma\theta}L_n(\theta_*, \gamma_*)]\|_2$.
Denote $o_i=\{y_i-\mu(\sigma_{\gamma_*}(x_i)\theta^\top z_i)-\mu(\sigma_\gamma(x_i)\theta_*^\top z_i)[1-\mu(\sigma_{\gamma_*}(x_i)\theta_*^\top z_i)]\sigma_{\gamma_*}(x_i)\theta_*^\top z_i\}\psi(x_i)z_i^\top$. Clearly, $\nabla^2_{\gamma\theta}L_n(\theta_*, \gamma_*)=\frac{1}{n}\sum_{i=1}^n o_i$. By Assumptions \ref{ass1} and \ref{ass2}, we have
\begin{equation}
\begin{aligned}
\|o_i\|_2\leq 2\psi_{\max}\phi_{\max}+\sqrt{d_1d_2}\gamma_{\max}\theta_{\max} \phi_{\max}^2\psi_{\max}^2
\end{aligned}
\end{equation}
and
\begin{equation}\label{e10010}
\begin{aligned}
\|\mathbb{E}o_i\|_2&=\|\mathbb{E}[\mathbb{E}(o_i|x_i,z_i)]\|_2\\
&=\|\mathbb{E}\{\mu(\sigma_{\gamma_*}(x_i)\theta_*^\top z_i)[1-\mu(\sigma_{\gamma_*}(x_i)\theta_*^\top z_i)]\sigma_{\gamma_*}(x_i)(\theta_*^\top z_i)\psi(x_i)z_i^\top\}\|_2\\
&= M.
\end{aligned}
\end{equation}
Therefore,
\begin{equation}
 \|o_i-\mathbb{E}o_i\|_2 \leq \|o_i\|_2+ \|\mathbb{E}o_i\|_2\leq 2\psi_{\max}\phi_{\max}+\sqrt{d_1d_2}\gamma_{\max}\theta_{\max} \phi_{\max}^2\psi_{\max}^2+M:=K_1.
\end{equation}
The matrix variance statistic is 
\begin{equation}
\left\|\sum_{i=1}^n\mathbb{E}[(o_i-\mathbb{E}o_i)(o_i-\mathbb{E}o_i)^\top\right\|_2\leq \sum_{i=1}^n\|\mathbb{E}[(o_i-\mathbb{E}o_i)(o_i-\mathbb{E}o_i)^\top\|_2\leq nK_1^2.
\end{equation}
We denote $d=d_1+d_2$. By Lemma \ref{lem612}, for any $\epsilon\geq 0$,  we have
\begin{equation}
    \begin{aligned}
 \mathbb{P}\left(\left\|\frac{1}{n}\sum_{i=1}^n o_i-\mathbb{E}\frac{1}{n}\sum_{i=1}^n o_i\right\|_2\geq \epsilon \right)&= \mathbb{P}\left(\left\|\sum_{i=1}^n o_i-\mathbb{E}\sum_{i=1}^n o_i\right\|_2\geq \epsilon n\right)\\
 &\leq (d_1+d_2)e^{\frac{-\epsilon^2n^2/2}{nK_1^2+nK_1\epsilon/3}}\\
 &=de^{\frac{-\epsilon^2n/2}{K_1^2+K_1\epsilon/3}}.
    \end{aligned}
\end{equation}
Equivalently, for any $\delta\in (0, 1)$, 
\begin{equation}
  \mathbb{P}\left(\left\|\frac{1}{n}\sum_{i=1}^n o_i-\mathbb{E}\frac{1}{n}\sum_{i=1}^n o_i\right\|_2\geq \frac{K_1\{\log(d/\delta)+\sqrt{\log(d/\delta)[\log(d/\delta)+18n]}\}}{3n}\right)\leq \delta.   
\end{equation}
With at least probability $1-\delta$, we have
\begin{equation}\label{e1009}
\begin{aligned}
   \|\nabla^2_{\gamma\theta}L_n(\theta_*, \gamma_*)-\mathbb{E}[\nabla^2_{\gamma\theta}L_n(\theta_*, \gamma_*)]\|_2&=  \|\nabla^2_{\theta\gamma}L(\theta_*,\gamma_*)-\mathbb{E}o_i\|_2\\
    &\leq \frac{K_1\{\log(d/\delta)+\sqrt{\log(d/\delta)[\log(d/\delta)+18n]}\}}{3n} . 
\end{aligned}
\end{equation}
By \eqref{e1001}, \eqref{e1008}, \eqref{e10010} and \eqref{e1009}, with at least probability $1-\delta$, we have
\begin{align*}
 &~~~~\|\nabla^2_{\gamma\theta}L_n(\theta_*, \gamma)\|_2\\
 &\leq 2\sqrt{d_1}\theta_{\max}\psi^2_{\max}\phi^2_{\max}(1+\sqrt{d_1d_2}\theta_{\max}\gamma_{\max}\phi_{\max}\psi_{\max})\|\gamma-\gamma_*\|_2+M\\
 &~~~~+   \frac{K_1\{\log(d/\delta)+\sqrt{\log(d/\delta)[\log(d/\delta)+18n]}\}}{3n}\\
 &\leq \sqrt{2d_1}b\theta_{\max}\psi^2_{\max}\phi^2_{\max}(1+\sqrt{d_1d_2}\theta_{\max}\gamma_{\max}\phi_{\max}\psi_{\max})+M\\
 &~~~~+   \frac{K_1\{\log(d/\delta)+\sqrt{\log(d/\delta)[\log(d/\delta)+18n]}\}}{3n}.
\end{align*}
We turn to bound the norm $\|\nabla^2_{\gamma\theta}L_n(\theta, \gamma_*)\|_2$. Begin by decomposing it as follows:
\begin{equation}\label{e10011}
\begin{aligned}
&~~~~\|\nabla^2_{\gamma\theta}L_n(\theta, \gamma_*)\|_2\\
&=\|\nabla^2_{\gamma\theta}L_n(\theta, \gamma_*)-\nabla^2_{\gamma\theta}L_n(\theta_*, \gamma_*)+\nabla^2_{\gamma\theta}L_n(\theta_*, \gamma_*)-\mathbb{E}[\nabla^2_{\gamma\theta}L_n(\theta_*, \gamma_*)]+\mathbb{E}[\nabla^2_{\gamma\theta}L_n(\theta_*, \gamma_*)]\|_2\\
&\leq \|\nabla^2_{\gamma\theta}L_n(\theta, \gamma_*)-\nabla^2_{\gamma\theta}L_n(\theta_*, \gamma_*)\|_2+\|\nabla^2_{\gamma\theta}L_n(\theta_*, \gamma_*)-\mathbb{E}[\nabla^2_{\gamma\theta}L_n(\theta_*, \gamma_*)]\|_2+\|\mathbb{E}[\nabla^2_{\gamma\theta}L_n(\theta_*, \gamma_*)]\|_2.
\end{aligned}
\end{equation}
We bound $\|\nabla^2_{\gamma\theta}L_n(\theta, \gamma_*)-\nabla^2_{\gamma\theta}L_n(\theta_*, \gamma_*)\|_2$. By \eqref{hess3}, we have
\begin{equation}\label{e10021}
 \begin{aligned}
&~~~~\nabla^2_{\gamma\theta}L_n(\theta, \gamma_*)-\nabla^2_{\gamma\theta}L_n(\theta_*, \gamma_*)\\
&=\frac{1}{n}\sum_{i=1}^n[\mu(\sigma_{\gamma_*}(x_i)\theta^\top z_i)-\mu(\sigma_{\gamma_*}(x_i)\theta_*^\top z_i)]\psi(x_i)z_i^\top\\
&~~~~+\frac{1}{n}\sum_{i=1}^n\sigma_{\gamma_*}(x_i) \psi(x_i) z_i^\top\{\mu(\sigma_{\gamma_*}(x_i)\theta^\top z_i)[1-\mu(\sigma_{\gamma_*}(x_i)\theta^\top z_i)]\theta^\top z_i\\
&~~~~-\mu(\sigma_{\gamma_*}(x_i)\theta_*^\top z_i)[1-\mu(\sigma_{\gamma_*}(x_i)\theta_*^\top z_i)]\theta_*^\top z_i\}.
\end{aligned}   
\end{equation}
By \eqref{e10001}, we obtain
\begin{equation}\label{e10061}
\|[\mu(\sigma_{\gamma_*}(x_i)\theta^\top z_i)-\mu(\sigma_{\gamma_*}(x_i)\theta_*^\top z_i)]\psi(x_i)z_i^\top\|_2=\sqrt{d_2}\gamma_{\max}\psi^2_{\max}\phi^2_{\max}\|\theta-\theta_*\|_2.
\end{equation}
Next,
\begin{equation}\label{e10031}
  \begin{aligned}
&~~~~\mu(\sigma_{\gamma_*}(x_i)\theta^\top z_i)[1-\mu(\sigma_{\gamma_*}(x_i)\theta^\top z_i)]\theta^\top z_i-\mu(\sigma_{\gamma_*}(x_i)\theta_*^\top z_i)[1-\mu(\sigma_{\gamma_*}(x_i)\theta_*^\top z_i)]\theta_*^\top z_i\\
&=\mu(\sigma_{\gamma_*}(x_i)\theta^\top z_i)[1-\mu(\sigma_{\gamma_*}(x_i)\theta^\top z_i)](\theta^\top z_i-\theta_*^\top z_i)\\
&~~~~+\{\mu(\sigma_{\gamma_*}(x_i)\theta^\top z_i)[1-\mu(\sigma_{\gamma_*}(x_i)\theta^\top z_i)]-\mu(\sigma_{\gamma_*}(x_i)\theta_*^\top z_i)[1-\mu(\sigma_{\gamma_*}(x_i)\theta_*^\top z_i)]\}\theta_*^\top z_i.
\end{aligned}  
\end{equation}
By \eqref{s2e0} and the fact $\mu(\sigma_{\gamma}(x_i)\theta_*^\top z_i)\in (0, 1)$, we have
\begin{equation}\label{e10041}
|\mu(\sigma_{\gamma}(x_i)\theta_*^\top z_i)[1-\mu(\sigma_{\gamma}(x_i)\theta_*^\top z_i)](\theta^\top z_i-\theta_*^\top z_i)|\leq \frac{\phi_{\max}}{2}\|\theta-\theta_*\|_2.
\end{equation}
By the Taylor expansion, there exists some value $\bar{\theta}$ between $\theta$ and $\theta_*$ such that
\begin{equation}\label{e10051}
\begin{aligned}
&~~~~|\mu(\sigma_{\gamma_*}(x_i)\theta^\top z_i)[1-\mu(\sigma_{\gamma_*}(x_i)\theta^\top z_i)]-\mu(\sigma_{\gamma_*}(x_i)\theta_*^\top z_i)[1-\mu(\sigma_{\gamma_*}(x_i)\theta_*^\top z_i)]|\\
&=|\mu(\sigma_{\gamma_*}(x_i)\bar{\theta}^\top z_i)[1-\mu(\sigma_{\gamma_*}(x_i)\bar{\theta}^\top z_i)][1-2\mu(\sigma_{\gamma_*}(x_i)\bar{\theta}^\top z_i)]\sigma_{\gamma_*}(x_i) z^\top_i(\theta-\theta_*)|\\
&\leq \frac{\sqrt{d_2}}{2}\gamma_{\max}\phi_{\max}\psi_{\max}\|\theta-\theta_*\|_2,
\end{aligned}    
\end{equation}
where the last inequality follows from Assumptions \ref{ass1} and \ref{ass2}, and the similar fact of \eqref{fact}. Therefore,
\begin{equation}\label{e100512}
\begin{aligned}
 &~~~~|\{\mu(\sigma_{\gamma_*}(x_i)\theta^\top z_i)[1-\mu(\sigma_{\gamma_*}(x_i)\theta^\top z_i)]-\mu(\sigma_{\gamma_*}(x_i)\theta_*^\top z_i)[1-\mu(\sigma_{\gamma_*}(x_i)\theta_*^\top z_i)]\}\theta_*^\top z_i|\\
 &\leq |\mu(\sigma_{\gamma_*}(x_i)\theta^\top z_i)[1-\mu(\sigma_{\gamma_*}(x_i)\theta^\top z_i)]-\mu(\sigma_{\gamma_*}(x_i)\theta_*^\top z_i)[1-\mu(\sigma_{\gamma_*}(x_i)\theta_*^\top z_i)]||\theta_*^\top z_i|\\
 &\leq \sqrt{d_1d_2}\gamma_{\max}\theta_{\max}\phi^2_{\max}\psi_{\max}\|\theta-\theta_*\|_2,  
\end{aligned}
\end{equation}
where the last inequality is derived from \eqref{e10051}. 
By \eqref{e10031}, \eqref{e10041} and \eqref{e100512}, we have
\begin{equation}\label{e10071}
    \begin{aligned}
&~~~~|\mu(\sigma_{\gamma_*}(x_i)\theta^\top z_i)[1-\mu(\sigma_{\gamma_*}(x_i)\theta^\top z_i)]\theta^\top z_i-\mu(\sigma_{\gamma_*}(x_i)\theta_*^\top z_i)[1-\mu(\sigma_{\gamma_*}(x_i)\theta_*^\top z_i)]\theta_*^\top z_i|\\
&\leq \phi_{\max}\left(\frac{1}{2}+\sqrt{d_1d_2}\gamma_{\max}\theta_{\max}\phi_{\max}\psi_{\max}\right)\|\theta-\theta_*\|_2.
    \end{aligned}
\end{equation}
On the other hand,
$$\|\sigma_{\gamma_*}(x_i) \psi(x_i) z_i^\top\|_2\leq\|\gamma_*\|_2\|\psi(x_i)\|_2\|z_i\|_2\|\psi(x_i)\|_2\leq 2\sqrt{d_2}\gamma_{\max}\psi_{\max}^2\phi_{\max}.$$
Combining \eqref{e10021}, \eqref{e10061} and \eqref{e10071}, we obtain
\begin{equation}\label{e10081}
\|\nabla^2_{\gamma\theta}L_n(\theta, \gamma_*)-\nabla^2_{\gamma\theta}L_n(\theta_*, \gamma_*)\|_2\leq  \sqrt{d_2}\gamma_{\max}\psi^2_{\max}\phi^2_{\max}(2+\sqrt{d_1d_2}\gamma_{\max}\theta_{\max}\phi_{\max}\psi_{\max})\|\theta-\theta_*\|_2.
\end{equation}
By \eqref{e10011}, \eqref{e10081}, \eqref{e10010} and \eqref{e1009}, with at least probability $1-\delta$, we have
\begin{align*}
 &~~~~\|\nabla^2_{\gamma\theta}L_n(\theta, \gamma_*)\|_2\\
 &\leq 2\sqrt{d_2}\gamma_{\max}\psi^2_{\max}\phi^2_{\max}(1+\sqrt{d_1d_2}\gamma_{\max}\theta_{\max}\phi_{\max}\psi_{\max})\|\theta-\theta_*\|_2+M\\
 &~~~~+   \frac{K_1\{\log(d/\delta)+\sqrt{\log(d/\delta)[\log(d/\delta)+18n]}\}}{3n}\\
 &\leq \sqrt{2d_2}b\gamma_{\max}\psi^2_{\max}\phi^2_{\max}(1+\sqrt{d_1d_2}\gamma_{\max}\theta_{\max}\phi_{\max}\psi_{\max})+M\\
 &~~~~+   \frac{K_1\{\log(d/\delta)+\sqrt{\log(d/\delta)[\log(d/\delta)+18n]}\}}{3n}.
\end{align*}
Noting that $c=\sqrt{2}\max\{\theta_{\max}\sqrt{d_1}, \gamma_{\max}\sqrt{d_2}\}\psi^2_{\max}\phi^2_{\max}(1+\sqrt{d_1d_2}\gamma_{\max}\theta_{\max}\phi_{\max}\psi_{\max})$, the proof is complete.

\subsection{Proof of Theorem \ref{thm1}}
By Assumption \ref{ass4}, we know $\min\{\lambda_\phi,\lambda_{\psi}\}-6c_0M>0$. 
Set $$b_0=\min\left\{\frac{\lambda_\phi-6c_0M}{8\sqrt{2d_2}\gamma_{\max}\phi^2_{\max}\psi_{\max}^2+6c_0c}, \frac{\lambda_\psi-6c_0M}{8\sqrt{2d_1}\theta_{\max}\phi^2_{\max}\psi_{\max}^2+6c_0c}\right\}$$ with $c$ defined in Lemma \ref{lem9}.
We begin by bounding $\|\theta_{t+1}-\theta_*\|_2^2$. According to the update in Algorithm \ref{alg1}, we have
\begin{equation}\label{t1e0}
\begin{aligned}
 \|\theta_{t+1}-\theta_*\|_2^2 &=\|\theta_t-\eta_1\nabla_\theta L_n(\theta_{t},\gamma_{t})-\theta_*\|_2^2\\
&=\|\theta_t-\theta_*\|_2^2+\eta_1^2\|\nabla_\theta L_n(\theta_{t},\gamma_{t})\|_2^2-2\eta_1\langle\nabla_\theta L_n(\theta_{t},\gamma_{t}), \theta_t-\theta_*\rangle.
\end{aligned}    
\end{equation}
By Lemmas \ref{lem2} and \ref{lem4}, we obtain
\begin{equation}\label{t1e4}
 \begin{aligned}
&~~~~\|\nabla_\theta L_n(\theta_{t},\gamma_{t})\|_2^2\\
&=\|\nabla_\theta L_n(\theta_{t},\gamma_{t})-\nabla_\theta L_n(\theta_{t},\gamma_*)+\nabla_\theta L_n(\theta_{t},\gamma_*)-\nabla_\theta L_n(\theta_*, \gamma_*)+\nabla_\theta L_n(\theta_*, \gamma_*)\|_2^2\\
&\leq 3[\|\nabla_\theta L_n(\theta_{t},\gamma_{t})-\nabla_\theta L_n(\theta_{t},\gamma_*)\|_2^2+\|\nabla_\theta L_n(\theta_{t},\gamma_*)-\nabla_\theta L_n(\theta_*, \gamma_*)\|_2^2+\|\nabla_\theta L_n(\theta_*, \gamma_*)\|_2^2]\\
&\leq 3\tilde{M}^2\|\gamma_t-\gamma_*\|_2^2+3d_2^2K^2\|\theta_t-\theta_*\|_2^2+3\|\nabla_\theta L_n(\theta_*, \gamma_*)\|_2^2.
\end{aligned}   
\end{equation}
Next, we have
\begin{equation}\label{t1e1}
\begin{aligned}
\langle\nabla_\theta L_n(\theta_{t},\gamma_{t}), \theta_t-\theta_*\rangle
&=\langle\nabla_\theta L_n(\theta_{t},\gamma_{t})-\nabla_\theta L_n(\theta_*,\gamma_{t}), \theta_t-\theta_*\rangle+\langle\nabla_\theta L_n(\theta_*,\gamma_{*}), \theta_t-\theta_*\rangle\\
&~~~~+\langle\nabla_\theta L_n(\theta_*,\gamma_{t})-\nabla_\theta L_n(\theta_*,\gamma_*), \theta_t-\theta_*\rangle\\
&\geq \langle\nabla_\theta L_n(\theta_{t},\gamma_{t})-\nabla_\theta L_n(\theta_*,\gamma_{t}), \theta_t-\theta_*\rangle-\frac{1}{2w}\|\nabla_\theta L_n(\theta_*,\gamma_{*})\|_2^2\\
&~~~~-\frac{w}{2}\|\theta_t-\theta_*\|_2^2+\langle\nabla_\theta L_n(\theta_*,\gamma_{t})-\nabla_\theta L_n(\theta_*,\gamma_*), \theta_t-\theta_*\rangle.
\end{aligned}
\end{equation}
By Lemma \ref{lem02}, with probability at least $1-d_1  \left(\frac{e}{2}\right) ^{-\frac{\lambda_{\phi}n}{8d_1K}}$,  we have for all $t \in {1, \cdots, T}$
\begin{equation}\label{t1e2}
\langle \nabla_\theta L_n(\theta_t, \gamma_t)-\nabla_\theta L_n(\theta_*, \gamma_t), \theta_t-\theta_*\rangle \geq  w\|\theta_t-\theta_*\|_2^2,
\end{equation}
From Lemma \ref{lem9}, with probability at least $1-\delta$, we have
\begin{align*}
|\langle\nabla_\theta L_n(\theta_*,\gamma_{t})-\nabla_\theta L_n(\theta_*, \gamma_*), \theta_t-\theta_*\rangle| &\leq \|\nabla_\theta L_n(\theta_*,\gamma_{t})-\nabla_\theta L_n(\theta_*, \gamma_*)\|_2\|\theta_t-\theta_*\|_2\\
&\leq\|\nabla^2_{\theta\gamma}L_n(\theta_*, \bar{\gamma}_t)\|_2\|\gamma_{t}-\gamma_*\|_2\|\theta_t-\theta_*\|_2 \\
&\leq (cb+M+f(d_1+d_2, n, \delta, K_1))\|\gamma_{t}-\gamma_*\|_2\|\theta_t-\theta_*\|_2\\
&\leq \frac{cb+M+f(d_1+d_2, n, \delta, K_1)}{2}(\|\gamma_{t}-\gamma_*\|_2^2+\|\theta_t-\theta_*\|_2^2).
\end{align*}
Therefore,
\begin{equation}\label{t1e3}
\begin{aligned}
\langle\nabla_\theta L_n(\theta_*,\gamma_{t})-\nabla_\theta L_n(\theta_*, \gamma_*), \theta_t-\theta_*\rangle\geq& -\frac{cb+M}{2}(\|\gamma_{t}-\gamma_*\|_2^2+\|\theta_t-\theta_*\|_2^2)\\
&-\frac{f(d_1+d_2, n, \delta, K_1)}{2}(\|\gamma_{t}-\gamma_*\|_2^2+\|\theta_t-\theta_*\|_2^2).
\end{aligned}
\end{equation}
Substituting \eqref{t1e2} and \eqref{t1e3} into \eqref{t1e1}, we obtain 
\begin{equation}\label{t1e5}
\begin{aligned}
\langle\nabla_\theta L_n(\theta_{t},\gamma_{t}), \theta_t-\theta_*\rangle\geq&\left(\frac{w-cb-M}{2}\right)\|\theta_t-\theta_*\|_2^2-\frac{cb+M}{2}\|\gamma_{t}-\gamma_*\|_2^2\\
&-\frac{f(d_1+d_2, n, \delta, K_1)}{2}(\|\gamma_{t}-\gamma_*\|_2^2+\|\theta_t-\theta_*\|_2^2)
-\frac{1}{2w}\|\nabla_\theta L_n(\theta_*,\gamma_{*})\|_2^2.
\end{aligned}
\end{equation}
Plugging \eqref{t1e4} and \eqref{t1e5} into \eqref{t1e0}, with probability at least $1-\delta-d_1  \left(\frac{e}{2}\right) ^{-\frac{\lambda_{\phi}n}{8d_2K}}$, we have
\begin{equation}\label{s1e6}
\begin{aligned}
  \|\theta_{t+1}-\theta_*\|_2^2&\leq [1+3d_2^2K^2\eta_1^2- (w-cb-M)\eta_1]\|\theta_t-\theta_*\|_2^2+(3\tilde{M}^2\eta_1+cb+M)\eta_1\|\gamma_{t}-\gamma_*\|_2^2\\
  &~~~~+\eta_1 f(d_1+d_2, n, \delta, K_1)(\|\gamma_{t}-\gamma_*\|_2^2+\|\theta_t-\theta_*\|_2^2)+(3\eta_1^2+\eta_1/w)\|\nabla_\theta L_n(\theta_*,\gamma_{*})\|_2^2.
\end{aligned}
\end{equation}
Now we turn to bound $\|\gamma_{t+1}-\gamma_*\|_2^2$. According to the update in Algorithm \ref{alg1}, we have
\begin{equation}\label{t1e00}
\begin{aligned}
 \|\gamma_{t+1}-\gamma_*\|_2^2 &=\|\gamma_t-\eta_2\nabla_\gamma L_n(\theta_{t+1},\gamma_{t})-\gamma_*\|_2^2\\
& = \|\gamma_t-\gamma_*\|_2^2+\eta_2^2\|\nabla_\gamma L_n(\theta_{t+1},\gamma_{t})\|_2^2-2\eta_2\langle\nabla_\gamma L_n(\theta_{t+1},\gamma_{t}), \gamma_t-\gamma_*\rangle.
\end{aligned}    
\end{equation}
By Lemmas \ref{lem2} and \ref{lem4}, we obtain
\begin{equation}\label{t1e04}
 \begin{aligned}
&~~~~~\|\nabla_\gamma L_n(\theta_{t+1},\gamma_{t})\|_2^2\\
&=\|\nabla_\gamma L_n(\theta_{t+1},\gamma_{t})-\nabla_\gamma L_n(\theta_{*},\gamma_t)+\nabla_\gamma L_n(\theta_{*},\gamma_t)-\nabla_\gamma L_n(\theta_{*},\gamma_*)+\nabla_\gamma L_n(\theta_{*},\gamma_*)\|_2^2\\
&\leq 3[\|\nabla_\gamma L_n(\theta_{t+1},\gamma_{t})-\nabla_\gamma L_n(\theta_{*},\gamma_t)\|_2+\|\nabla_\gamma L_n(\theta_{*},\gamma_t)-\nabla_\gamma L_n(\theta_{*},\gamma_*)\|_2+\|\nabla_\gamma L_n(\theta_{*},\gamma_*)\|_2^2]\\
&\leq 3\tilde{M}^2\|\theta_{t+1}-\theta_*\|_2^2+3d_1^2K^2\|\gamma_t-\gamma_*\|_2^2+3\|\nabla_\gamma L_n(\theta_{*},\gamma_*)\|_2^2.
\end{aligned}   
\end{equation}
Next,
\begin{equation}\label{t1e01}
\begin{aligned}
\langle\nabla_\gamma L_n(\theta_{t+1},\gamma_{t}), \gamma_t-\gamma_*\rangle &=\langle\nabla_\gamma L_n(\theta_{t+1},\gamma_{t})-\nabla_\gamma L_n(\theta_{t+1},\gamma_*), \gamma_t-\gamma_*\rangle+\langle\nabla_\gamma L_n(\theta_{*},\gamma_*), \gamma_t-\gamma_*\rangle\\
&~~~~+\langle\nabla_\gamma L_n(\theta_{t+1},\gamma_*)-\nabla_\gamma L_n(\theta_*,\gamma_*), \gamma_t-\gamma_*\rangle\\
&\geq \langle\nabla_\gamma L_n(\theta_{t+1},\gamma_{t})-\nabla_\gamma L_n(\theta_{t+1},\gamma_*), \gamma_t-\gamma_*\rangle-\frac{1}{2w}\|\nabla_\gamma L_n(\theta_{*},\gamma_*)\|_2^2\\
&~~~~-\frac{w}{2}\|\gamma_t-\gamma_*\|_2^2+\langle\nabla_\gamma L_n(\theta_{t+1},\gamma_*)-\nabla_\gamma L_n(\theta_*,\gamma_*), \gamma_t-\gamma_*\rangle.
\end{aligned}
\end{equation}
By Lemma \ref{lem02}, the following
\begin{equation}\label{t1e02}
\langle \nabla_\gamma L_n(\theta_{t+1}, \gamma_t)-\nabla_\gamma L_n(\theta_{t+1}, \gamma_*), \gamma_t-\gamma_*\rangle \geq  w\|\gamma_t-\gamma_*\|_2^2,
\end{equation}
holds with probability at least $1-d_2  \left(\frac{e}{2}\right) ^{-\frac{\lambda_{\psi}n}{8d_1K}}$, uniformly over $t$. By Lemma \ref{lem9}, with probability at least $1-\delta$, we have
\begin{align*}
&~~~~|\langle\nabla_\gamma L_n(\theta_{t+1}, \gamma_*)-\nabla_\gamma L_n(\theta_{*},\gamma_*), \gamma_t-\gamma_*\rangle|\\
&\leq \|\nabla_\gamma L_n(\theta_{t+1}, \gamma_*)-\nabla_\gamma L_n(\theta_{*},\gamma_*)\|_2\|\theta_{t+1}-\theta_*\|_2\|\gamma_t-\gamma_*\|_2\\
&\leq\|\nabla^2_{\theta\gamma}L_n(\bar{\theta}_t, \gamma_*)\|_2\|\theta_{t+1}-\theta_*\|_2\|\gamma_{t}-\gamma_*\|_2 \\
&\leq (cb+M+f(d_1+d_2, n, \delta, K_1))\|\theta_{t+1}-\theta_*\|_2\|\gamma_{t}-\gamma_*\|_2\\
&\leq \frac{cb+M+f(d_1+d_2, n, \delta, K_1)}{2}(\|\gamma_{t}-\gamma_*\|_2^2+\|\theta_{t+1}-\theta_*\|_2^2).
\end{align*}
Therefore,
\begin{equation}\label{t1e03}
\begin{aligned}
\langle\nabla_\gamma L_n(\theta_{t+1}, \gamma_*)-\nabla_\gamma L_n(\theta_{*},\gamma_*), \gamma_t-\gamma_*\rangle&\geq -\frac{f(d_1+d_2, n, \delta, K_1)}{2}(\|\gamma_{t}-\gamma_*\|_2^2+\|\theta_{t+1}-\theta_*\|_2^2)\\
&-\frac{cb+M}{2}(\|\gamma_{t}-\gamma_*\|_2^2+\|\theta_{t+1}-\theta_*\|_2^2).
\end{aligned}
\end{equation}
By \eqref{t1e01}, \eqref{t1e02} and \eqref{t1e03}, we conclude
\begin{equation}\label{t1e05}
\begin{aligned}
\langle\nabla_\gamma L_n(\theta_{t+1},\gamma_{t}), \gamma_t-\gamma_*\rangle\geq& \left(\frac{w-cb-M}{2}\right)\|\gamma_t-\gamma_*\|_2^2-\frac{cb+M}{2}\|\theta_{t+1}-\theta_*\|_2^2\\
&-\frac{f(d_1+d_2, n, \delta, K_1)}{2}(\|\gamma_{t}-\gamma_*\|_2^2+\|\theta_{t+1}-\theta_*\|_2^2)-\frac{1}{2w}\|\nabla_\gamma L_n(\theta_{*},\gamma_*)\|_2.    
\end{aligned}
\end{equation}
Combining \eqref{t1e00}, \eqref{t1e04} and \eqref{t1e05}, with probability at least $1-\delta-d_2 \left(\frac{e}{2}\right) ^{-\frac{\lambda_{\psi}n}{8d_1K}}$, we have
\begin{equation}\label{s106}
\begin{aligned}
\|\gamma_{t+1}-\gamma_*\|_2^2&\leq [1+3d_1^2K^2\eta_2^2- (w-cb-M)\eta_2]\|\gamma_t-\gamma_*\|_2^2+(3\tilde{M}^2\eta_2+cb+M)\eta_2\|\theta_{t+1}-\theta_*\|_2^2\\
&~~~~+\eta f(d_1+d_2, n, \delta, K_1)(\|\gamma_{t}-\gamma_*\|_2^2+\|\theta_{t+1}-\theta_*\|_2^2)+(3\eta_2^2+\eta_2/w)\|\nabla_\gamma L_n(\theta_{*},\gamma_*)\|_2^2.
\end{aligned}
\end{equation}
Under the condition $0\leq b<b_0$, we can verify $w-3(cb+M)>0$. Denote 
\begin{equation}\label{eta}
    \eta=\max\{\eta_1, \eta_2\}.
\end{equation}
We define 
$A_1=1+\max\{3d_2^2K^2\eta_1^2- (w-cb-M)\eta_1, 3d_1^2K^2\eta_2^2- (w-cb-M)\eta_2\}$ and $ A_2=3\tilde{M}^2\eta^2+(cb+M)\eta$.
If $(w-cb-M)^2-12d_2^2K^2\geq 0$, set $$\eta_1\in \left(0, \mathop{\min}\left\{\frac{w-cb-M-\sqrt{(w-cb-M)^2-12d_2^2K^2}}{6d_2^2K^2}, \frac{w-3(cb+M)}{3(d_2^2K^2+\tilde{M}^2)}\right\}\right).$$ Otherwise, set $\eta_1\in \left(0, \frac{w-3(cb+M)}{3d_2^2K^2+\tilde{M}^2)}\right)$. If $(w-cb-M)^2-12d_1^2K^2\geq 0$, set $$\eta_2\in \left(0, \mathop{\min}\left\{\frac{w-cb-M-\sqrt{(w-cb-M)^2-12d_1^2K^2}}{6d_1^2K^2}, \frac{w-3(cb+M)}{3(d_1^2K^2+\tilde{M}^2)}\right\}\right).$$ Otherwise, set $\eta_2\in \left(0, \frac{w-3(cb+M)}{3d_1^2K^2+\tilde{M}^2)}\right)$. Then, it is easy to verify
\begin{equation*}
 \begin{aligned}
 A_1>0, A_2>0\ \text{and}\ A_1+2A_2\in(0,1).
\end{aligned}   
\end{equation*}
We define 
\begin{equation}\label{a1a2}
    A_{1n}=A_1+\eta f(d_1+d_2, n, \delta, K_1)\ \text{and}\ A_{2n}=A_2+\eta f(d_1+d_2, n, \delta, K_1).
\end{equation}
 Since $A_1, A_2$, $\eta$ are constants and $f(d_1+d_2, n, \delta, K_1)$ decreases as $n$ increases, there exists some constant $n_1$ such that when $n>n_1$, we have
\begin{equation}\label{a1a22}
 \begin{aligned}
 A_{1n}>0, A_{2n}>0\ \text{and}\ A_{1n}+2A_{2n}\in(0,1).
\end{aligned}   
\end{equation}
It follows that
\begin{equation}\label{a1+a2}
A_{1n}\in(0,1),  A_{2n}\in (0, 1),  A_{1n}+A_{2n}\in(0,1)  
\end{equation}
and 
\begin{equation}\label{a1+a22}
 \begin{aligned}
A_{1n}+A_{2n}^2+A_{1n}A_{2n} &=A_{1n}+A_{2n}(A_{1n}+A_{2n})< A_{1n}+A_{2n}\in(0,1).
\end{aligned}   
\end{equation}
By \eqref{s1e6}, \eqref{s106} and Lemma \ref{lem5}, with probability at least $1-3\delta-d_1  \left(\frac{e}{2}\right) ^{-\frac{\lambda_{\phi}n}{8d_2K}}-d_2  \left(\frac{e}{2}\right) ^{-\frac{\lambda_{\psi}n}{8d_1K}}$ for  $0<\delta<1$, we have
\begin{equation}\label{s107}
\begin{aligned}
  \|\theta_{t+1}-\theta_*\|_2^2&\leq A_1\|\theta_t-\theta_*\|_2^2+A_2\|\gamma_{t}-\gamma_*\|_2^2+(3\eta^2+\eta/w)\|\nabla_\theta L_n(\theta_*,\gamma_{*})\|_2^2\\
  &~~~~+\eta f_3(n, \delta)(\|\gamma_{t}-\gamma_*\|_2^2+\|\theta_{t}-\theta_*\|_2^2)\\
  &\leq A_{1n}\|\theta_t-\theta_*\|_2^2+A_{2n}\|\gamma_{t}-\gamma_*\|_2^2+(3\eta^2+\eta/w)f^2(d_1+1, n,\delta, 2\sqrt{d_2K})
\end{aligned}
\end{equation}
and
\begin{equation}\label{s108}
\begin{aligned}
&~~~~\|\gamma_{t+1}-\gamma_*\|_2^2\\
&\leq A_1\|\gamma_t-\gamma_*\|_2^2+A_2\|\theta_{t+1}-\theta_*\|_2^2+(3\eta^2+\eta/w)\|\nabla_\gamma L_n(\theta_{*},\gamma_*)\|_2^2\\
&~~~~+\eta f_3(n, \delta)(\|\gamma_{t}-\gamma_*\|_2^2+\|\theta_{t+1}-\theta_*\|_2^2)\\
&=A_{1n}\|\gamma_t-\gamma_*\|_2^2+A_{2n}\|\theta_{t+1}-\theta_*\|_2^2+(3\eta^2+\eta/w)\|\nabla_\gamma L_n(\theta_{*},\gamma_*)\|_2^2\\
&\leq (A_{1n}+A^2_{2n})\|\gamma_t-\gamma_*\|_2^2+A_{1n}A_{2n}\|\theta_{t}-\theta_*\|_2^2+(A_{2n}+1)(3\eta^2+\eta/w)f^2(d_2+1, n,\delta, 2\sqrt{d_1K}).
\end{aligned}
\end{equation}
When calculating the above probability, we should note that the probability $\delta$ in \eqref{s1e6} and \eqref{s106} is derived from the same event.
Since $f(d_1+1, n ,\delta, 2\sqrt{d_2K})$ and $f(d_2+1, n ,\delta, 2\sqrt{d_1K})$ decrease in $n$, there exists $n_2$ such that when $n\geq n_2$ we have
$$\frac{(A_{1n}+A_{2n})b^2}{2}+(3\eta^2+\eta/w)f^2(d_1+1, n ,\delta, 2\sqrt{d_2K})\leq \frac{b^2}{2}$$
and
$$ \frac{(A_{1n}+A_{2n}^2+A_{1n}A_{2n})b^2}{2}+(A_{2n}+1)(3\eta^2+\eta/w)f^2(d_2+1, n ,\delta, 2\sqrt{d_1K})\leq \frac{b^2}{2}.$$
Note that we have $\|\theta_0-\theta_*\|_2^2\leq b^2/2$ and $\|\gamma_0-\gamma_*\|_2^2\leq b^2/2$ by initialization. Using \eqref{s107} and \eqref{s108}, we can inductively prove that $\|\theta_{t}-\theta_*\|_2^2\leq b^2/2$ and $\|\gamma_{t}-\gamma_*\|_2^2\leq b^2/2$ for all $t\geq 0$ when $n\geq n_2$.
By \eqref{s107} and \eqref{s108}, we obtain
\begin{align*}
\|\theta_{t+1}-\theta_*\|_2^2+\|\gamma_{t+1}-\gamma_*\|_2^2
&\leq (A_{1n}+A_{2n}+A_{2n}^2)\|\gamma_t-\gamma_*\|_2^2+(A_{1n}+A_{1n}A_{2n})\|\theta_{t}-\theta_*\|_2^2\\
&~~~~+(A_{2n}+2)(3\eta^2+\eta/w)f^2(\max\{d_1, d_2\}+1, n ,\delta, 2\sqrt{(d_1+d_2)K})\\
&\leq (A_{1n}+A_{2n}+A_{2n}^2)(\|\gamma_t-\gamma_*\|_2^2+\|\theta_{t}-\theta_*\|_2^2)\\
&~~~~(A_{2n}+2)(3\eta^2+\eta/w)f^2(\max\{d_1, d_2\}+1, n ,\delta, 2\sqrt{(d_1+d_2)K}),
\end{align*}
where the second inequality follows from
\begin{equation}\label{new1}
 A_{1n}+A_{2n}+A_{2n}^2-(A_{1n}+A_{1n}A_{2n})=A_{2n}(1+A_{2n}-A_{1n})> 0.   
\end{equation}
Let $n_0\geq \max\{n_1, n_2\}$ be chosen sufficiently large such that $d_2\left(\frac{e}{2}\right) ^{-\frac{\lambda_{\phi}n}{8d_1K}}<\frac{1}{2n}$ and $d_1\left(\frac{e}{2}\right) ^{-\frac{\lambda_{\phi}n}{8d_2K}}<\frac{1}{2n}$ hold for all $n\geq n_0$. We denote $\rho=A_{1n_0}+A_{2n_0}+A_{2n_0}^2$. By \eqref{a1a22}, we have
\begin{equation}\label{new2}
0<\rho<A_{1n_0}+2A_{2n_0}<1.
\end{equation}
By recursion, with probability at least $1-3\delta-\frac{1}{n}$, when $n>n_0$, we obtain
\begin{align*}
 \|\theta_{T}-\theta_*\|_2^2+\|\gamma_{T}-\gamma_*\|_2^2&\leq \rho^T b^2+\frac{(A_{2n}+2)(3w\eta^2+\eta)f^2(\max\{d_1, d_2\}+1, n ,\delta, 2\sqrt{(d_1+d_2)K})}{(1-\rho)w}   \\
 &<\rho^T b^2+\frac{c'_2f^2(\max\{d_1, d_2\}+1, n ,\delta, 2\sqrt{(d_1+d_2)K})}{(1-\rho)},
\end{align*}
where $c'_2=(A_{2n_0}+2)(3w\eta^2+\eta)/w$. By \eqref{f}, there exists some constant $c_1$ such that $c'_2f^2(\max\{d_1, d_2\}+1, n ,\delta, 2\sqrt{(d_1+d_2)K})\leq \frac{c_1\log (1/\delta)}{n}$ since both $d_1$ and $d_2$ are fixed. Since $\delta$ can be any value in (0, 1), the probability can be equivalently expressed as $1-\delta-\frac{1}{n}$. This completes the proof.

\section{Proofs for Uncertainty Quantification}\label{pfuq}
This section contains the proofs for the theoretical results in Section \ref{sec4.2} and Section \ref{sec05}.
\subsection{Proof of Lemma \ref{lem10}}
By Assumptions \ref{ass1} and \ref{ass2}, we have $|\sigma_{\gamma_*}(x)\theta_*^\top z|\leq 2 \theta_{\max}\gamma_{\max}\phi_{\max}\psi_{\max}$. Then
\begin{equation*}
    \mu(\sigma_{\gamma_*}(x)\theta_*^\top z)[1-\mu(\sigma_{\gamma_*}(x)\theta_*^\top z)]\geq \frac{e^{ 2 \theta_{\max}\gamma_{\max}\phi_{\max}\psi_{\max}}}{(1+e^{ 2 \theta_{\max}\gamma_{\max}\phi_{\max}\psi_{\max}})^2}:=\kappa>0.
\end{equation*}
By \eqref{hess2} and Assumption \ref{ass3}, we have
\begin{equation*}
    \mathcal{I}_{\gamma\gamma}(\theta_*, \gamma_*)=\mathbb{E}\{\mu(\sigma_{\gamma_*}(x)\theta_*^\top z)[1-\mu(\sigma_{\gamma_*}(x)\theta_*^\top z)](\theta_*^\top z)^2\psi(x)\psi^\top(x)\}\succeq \kappa \mathbb{E}[(\theta_*^\top z)^2\psi(x)\psi^\top(x)]\succeq\kappa \lambda_\psi I,
\end{equation*}
which is positive definite. Under Assumption \ref{ass5}, $\mathcal{I}(\tau_*)$ is positive definite by following Lemma \ref{schur}.

\subsection{Proof of Theorem \ref{thm6}}
Denote $\hat{\tau}_n=(\hat{\theta}_n^\top, \hat{\gamma}_n^\top)^\top$. The proof of Theorem \ref{thm6} proceeds by first showing that $\|\tau_T-\hat{\tau}_n\|_2 \to 0$ as $T \to \infty$. Once this consistency is established, the asymptotic normality of $\tau_T$ follows from that of the MLE $\hat{\tau}_n$, by appropriately controlling the growth rate of $T$ relative to $n$. We first state the asymptotic normality of the MLE $\hat{\tau}_n$.
\begin{theorem}\label{thm5}
Let the assumptions of Lemma \ref{lem10} hold. As $n\rightarrow\infty$, we have
  \begin{equation*}
\sqrt{n}(\hat{\tau}_n-\tau_*)\stackrel{D}{\longrightarrow} \mathcal{N}(0, \mathcal{I}^{-1}(\tau_*)). 
\end{equation*}  
\end{theorem}
Theorem \ref{thm5} provides the classical normal approximation for the MLE, with the covariance structure determined by the inverse of the Fisher information matrix. The proof follows the standard arguments for asymptotic normality of MLE (see, e.g., \cite{Vaart_1998, Casella}). For completeness, we provide the detailed proof.
\begin{proof}
We expand the first derivative of the negative log-likelihood around the true value $\tau_*$,
\begin{equation*}
\nabla L_n(\hat{\tau}_n)=\nabla L_n(\tau_*)+\nabla^2 L_n(\tau_*)(\hat{\tau}_n-\tau_*)+[\nabla^2 L_n(\bar{\tau})-\nabla^2 L_n(\tau_*)](\hat{\tau}_n-\tau_*)
\end{equation*}
where 
\begin{equation}\label{tau00}
    \bar{\tau}=c\hat{\tau}_n+(1-c)\tau_*
\end{equation}
for some $c\in[0, 1]$. Note that $\nabla L_n(\hat{\tau}_n)=\boldsymbol{0}$. Rearranging and multiplying by $\sqrt{n}$ gives us
\begin{equation}\label{t302}
\sqrt{n}\{\nabla^2 L_n(\tau_*)+[\nabla^2 L_n(\bar{\tau})-\nabla^2 L_n(\tau_*)]\}(\hat{\tau}_n-\tau_*)=-\sqrt{n}\nabla L_n(\tau_*).
\end{equation}
By \eqref{hess}, Assumptions \ref{ass1} and \ref{ass2}, there exist some positive constants $C_1$ and $C_1'$ such that
\begin{align*}
\|\nabla^2_{\theta\theta} L_n(\bar{\theta}, \bar{\gamma})-\nabla^2_{\theta\theta} L_n(\theta_*, \gamma_*)\|_2&\leq \frac{C_1}{n}\sum_{i=1}^n|\sigma^2_{\bar{\gamma}}(x_i)-\sigma^2_{\gamma_*}(x_i)|\\
&=\frac{C_1}{n}\sum_{i=1}^n|[\sigma_{\bar{\gamma}}(x_i)+\sigma_{\gamma_*}(x_i)][\sigma_{\bar{\gamma}}(x_i)-\sigma_{\gamma_*}(x_i)]|\\
&\leq C_1'\|\bar{\gamma}-\gamma_*\|_2.
\end{align*}
By \eqref{hess2}, Assumptions \ref{ass1} and \ref{ass2}, there exist some positive constants $C_2$ and $C_2'$ such that
\begin{align*}
\|\nabla^2_{\gamma\gamma} L_n(\bar{\theta}, \bar{\gamma})-\nabla^2_{\gamma\gamma} L_n(\theta_*, \gamma_*)\|_2&\leq \frac{C_2}{n}\sum_{i=1}^n|(\bar{\theta}^\top z_i)^2-(\theta_*^\top z_i)^2|\\
&=\frac{C_2}{n}\sum_{i=1}^n|[\bar{\theta}^\top z_i+\theta_*^\top z_i][\bar{\theta}^\top z_i-\theta_*^\top z_i]|\\
&\leq C_2'\|\bar{\theta}-\theta_*\|_2.
\end{align*}
By \eqref{hess3}, Assumptions \ref{ass1} and \ref{ass2}, there exist some positive constants $C_3, C_3'$ and $C_3''$ such that
\begin{align*}
&~~~~\|\nabla^2_{\gamma\theta} L_n(\bar{\theta}, \bar{\gamma})-\nabla^2_{\gamma\theta} L_n(\theta_*, \gamma_*)\|_2\\
&\leq \frac{C_3}{n}\sum_{i=1}^n[|\mu(\sigma_{\gamma_*}(x_i)\theta_*^\top z_i)-\mu(\sigma_{\bar\gamma}(x_i)\bar{\theta}^\top z_i)|+|\sigma_{\bar{\gamma}}(x_i)\bar{\theta}^\top z_i-\sigma_{\gamma_*}(x_i)\theta_*^\top z_i|]\\
&\leq \frac{C_3'}{n}\sum_{i=1}^n|\sigma_{\bar{\gamma}}(x_i)\bar{\theta}^\top z_i-\sigma_{\gamma_*}(x_i)\theta_*^\top z_i|\\
&=\frac{C_3'}{n}\sum_{i=1}^n|\sigma_{\bar{\gamma}}(x_i)\bar{\theta}^\top z_i-\sigma_{\bar{\gamma}}(x_i)\theta_*^\top z_i+\sigma_{\bar{\gamma}}(x_i)\theta_*^\top z_i-\sigma_{\gamma_*}(x_i)\theta_*^\top z_i|\\
&\leq C_3''(\|\bar{\theta}-\theta_*\|_2+\|\bar{\gamma}-\gamma_*\|_2),
\end{align*}
where the second inequality follows from the mean value theorem. Therefore, there exists some constant $C$ such that
\begin{equation*}
\|\nabla^2 L_n(\tau_*)-\nabla^2 L_n(\bar{\tau})\|_2\leq C(\|\bar{\theta}-\theta_*\|_2+\|\bar{\gamma}-\gamma_*\|_2)=cC(\|\hat{\theta}_n-\theta_*\|_2+\|\hat{\gamma}_n-\gamma_*\|_2),
\end{equation*}
where the last equality follows from \eqref{tau00}.
Therefore, by Lemma \ref{mlecon}, we obtain
\begin{equation}\label{t303}
\|\nabla^2 L_n(\tau_*)-\nabla^2 L_n(\bar{\tau})\|_2\stackrel{p}{\longrightarrow}0. 
\end{equation}
By the central limit theorem, we have
\begin{equation}\label{t304}
 -\sqrt{n}\nabla L_n(\tau_*)\stackrel{D}{\longrightarrow} \mathcal{N}(0, \mathcal{I}(\tau_*)).  
\end{equation}
Combining \eqref{t302}, \eqref{t303} and \eqref{t304}, by Slutsky's theorem, we have 
\begin{equation*}
\sqrt{n}(\hat{\tau}_n-\tau_*)\stackrel{D}{\longrightarrow} \mathcal{N}(0, \mathcal{I}^{-1}(\tau_*)). 
\end{equation*}
\end{proof}
Next, we propose the following Theorem that demonstrates that the estimator from Algorithm \ref{alg1} converges to the MLE when MLE is close to the true parameters.
\begin{theorem}\label{thm2}
Let the conditions in Theorem \ref{thm1} hold. Suppose $\|\hat{\theta}_n-\theta_*\|_2\leq \epsilon/\sqrt{2}$ and $\|\hat{\gamma}_n-\gamma_*\|_2\leq \epsilon/\sqrt{2}$ when $n\geq n_{\epsilon}$ for some constant $0\leq \epsilon<b$. When $n>\max\{n_0, n_{\epsilon}\}$,  with probability at least $1-\delta-d_1  \left(\frac{e}{2}\right) ^{-\frac{\lambda_{\phi}n}{8d_2K}}-d_2  \left(\frac{e}{2}\right) ^{-\frac{\lambda_{\psi}n}{8d_1K}}$, we have
$$\|\theta_T-\hat{\theta}_n\|^2_2+\|\gamma_T-\hat{\gamma}_n\|^2_2\leq \rho_1^T (b+\epsilon)^2,$$
where $0<\rho_1<1$.
\end{theorem}
 The condition $\|\hat{\theta}_n-\theta_*\|_2\leq \epsilon/\sqrt{2}$ and $\|\hat{\gamma}_n-\gamma_*\|_2\leq \epsilon/\sqrt{2}$ for large enough $n$ can be guaranteed by the consistency of MLE, see Lemma \ref{mlecon}. The proof of Theorem \ref{thm2} is postponed and presented after the proof of Theorem \ref{thm6}.

Now we prove Theorem \ref{thm6}. 
By Lemma \ref{mlecon}, for any $\epsilon>0$, there exists $n_{\epsilon}$ such that $\|\hat{\theta}_n-\theta_*\|_2\leq \epsilon/\sqrt{2}$ and $\|\hat{\gamma}_n-\gamma_*\|_2\leq \epsilon/\sqrt{2}$ for all $n>n_{\epsilon}$ with probability 1.
By Theorem \ref{thm2}, with probability at least $1-\delta-d_1  \left(\frac{e}{2}\right) ^{-\frac{\lambda_{\phi}n}{8d_2K}}-d_2  \left(\frac{e}{2}\right) ^{-\frac{\lambda_{\psi}n}{8d_1K}}$, we have
\begin{equation*}
\sqrt{n}\|\tau_T-\hat{\tau}_n\|_2=\sqrt{n}\sqrt{\|\theta_T-\hat{\theta}_n\|_2^2+\|\gamma_T-\hat{\gamma}_n\|_2^2}\leq \sqrt{n}\sqrt{\rho_1^T(b+\epsilon)^2}=(b+\epsilon)\sqrt{n\rho_1^T}.
\end{equation*}
Since $n\rho_1^T\rightarrow 0$ as $T\rightarrow\infty$ and $n\rightarrow\infty$, we have
\begin{equation*}
\sqrt{n}\|\tau_T-\hat{\tau}_n\|_2\stackrel{p}{\longrightarrow}0.
\end{equation*}
Therefore, by Slutsky's theorem and Theorem \ref{thm5}, we have
\begin{equation*}
    \sqrt{n}(\tau_T-\tau_*)=\sqrt{n}(\tau_T-\hat{\tau}_n)+\sqrt{n}(\hat{\tau}_n-\tau_*)\stackrel{D}{\longrightarrow} \mathcal{N}(0, \mathcal{I}^{-1}(\tau_*)).  
\end{equation*}

\subsection{Proof of Theorem \ref{thm2}}
We first bound $\|\theta_{t+1}-\hat{\theta}_n\|_2^2$. According to the update rule in Algorithm \ref{alg1}, we have 
\begin{equation}\label{t1e200}
\begin{aligned}
 \|\theta_{t+1}-\hat{\theta}_n\|_2^2 &=\|\theta_t-\eta_1\nabla_\theta L_n(\theta_{t},\gamma_{t})-\hat{\theta}_n\|_2^2\\
&=\|\theta_t-\hat{\theta}_n\|_2^2+\eta_1^2\|\nabla_\theta L_n(\theta_{t},\gamma_{t})\|_2^2-2\eta_1\langle\nabla_\theta L_n(\theta_{t},\gamma_{t}), \theta_t-\hat{\theta}_n\rangle.
\end{aligned}    
\end{equation}
Since $\nabla_\theta L_n(\hat{\theta}_n,\hat{\gamma}_n)=0$. Applying Lemmas \ref{lem2} and \ref{lem4} yields
\begin{equation}\label{t1e24}
 \begin{aligned}
\|\nabla_\theta L_n(\theta_{t},\gamma_{t})\|_2^2&=\|\nabla_\theta L_n(\theta_{t},\gamma_{t})-\nabla_\theta L_n(\theta_{t},\hat{\gamma}_n)+\nabla_\theta L_n(\theta_{t},\hat{\gamma}_n)-\nabla_\theta L_n(\hat{\theta}_n,\hat{\gamma}_n)\|_2^2\\
&\leq 2\|\nabla_\theta L_n(\theta_{t},\gamma_{t})-\nabla_\theta L_n(\theta_{t},\hat{\gamma}_n)\|_2^2+2\|\nabla_\theta L_n(\theta_{t},\hat{\gamma}_n)-\nabla_\theta L_n(\hat{\theta}_n,\hat{\gamma}_n)\|_2^2\\
&\leq 2\tilde{M}^2\|\gamma_t-\hat{\gamma}_n\|_2^2+2d_2^2K^2\|\theta_t-\hat{\theta}_n\|_2^2.
\end{aligned}   
\end{equation}
Next,
\begin{equation}\label{t1e21}
\langle\nabla_\theta L_n(\theta_{t},\gamma_{t}), \theta_t-\hat{\theta}_n\rangle=\langle\nabla_\theta L_n(\theta_{t},\gamma_{t})-\nabla_\theta L_n(\hat{\theta}_n,\gamma_{t}), \theta_t-\hat{\theta}_n\rangle+\langle\nabla_\theta L_n(\hat{\theta}_n,\gamma_{t})-\nabla_\theta L_n(\hat{\theta}_n,\hat{\gamma}_n), \theta_t-\hat{\theta}_n\rangle .
\end{equation}
By Lemma \ref{lem02}, with probability at least $1-d_1  \left(\frac{e}{2}\right) ^{-\frac{\lambda_{\phi}n}{8d_1K}}$, the following
\begin{equation}\label{t1e22}
\langle \nabla_\theta L_n(\theta_t, \gamma_t)-\nabla_\theta L_n(\hat{\theta}_n, \gamma_t), \theta_t-\hat{\theta}_n\rangle \geq  w\|\theta_t-\hat{\theta}_n\|_2^2,
\end{equation}
holds for all $t\in\{1,\cdots,T\}$. Denote $\bar{\gamma}_t=c\gamma_t+(1-c)\hat{\gamma}_n$ for $c\in[0, 1]$. We have 
\begin{align*}
    \|\bar{\gamma}_t-\gamma_*\|_2&=\|c\gamma_t+(1-c)\hat{\gamma}_n-\gamma_*\|_2\\
    &\leq c\|\gamma_t-\gamma_*\|_2+(1-c)\|\hat{\gamma}_n-\gamma_*\|_2\\
    &\leq cb/\sqrt{2}+(1-c)\epsilon/\sqrt{2}\\
    &\leq b/\sqrt{2},
\end{align*}
and
\begin{align*}
\|\nabla^2_{\theta\gamma}L_n(\hat{\theta}_n, \bar{\gamma}_t)\|_2&=\|\nabla^2_{\theta\gamma}L_n(\theta_*, \bar{\gamma}_t)+\nabla^2_{\theta\gamma}L_n(\hat{\theta}_n, \bar{\gamma}_t)-\nabla^2_{\theta\gamma}L_n(\theta_*, \bar{\gamma}_t)\|_2\\
&\leq \|\nabla^2_{\theta\gamma}L_n(\theta_*, \bar{\gamma}_t)\|_2+\|\nabla^2_{\theta\gamma}L_n(\hat{\theta}_n, \bar{\gamma}_t)-\nabla^2_{\theta\gamma}L_n(\theta_*, \bar{\gamma}_t)\|_2\\
&\leq \|\nabla^2_{\theta\gamma}L_n(\theta_*, \bar{\gamma}_t)\|_2+\sqrt{2}c_3\|\hat{\theta}_n-\theta_*\|_2\\
&\leq \|\nabla^2_{\theta\gamma}L_n(\theta_*, \bar{\gamma}_t)\|_2+c_3\epsilon
\end{align*}
for some positive constant $c_3$.
By Lemma \ref{lem9}, with probability at least $1-
\delta$,  we have
\begin{align*}
&~~~~|\langle\nabla_\theta L_n(\hat{\theta}_n,\gamma_{t})-\nabla_\theta L_n(\hat{\theta}_n,\hat{\gamma}_n), \theta_t-\hat{\theta}_n\rangle| \\
&\leq \|\nabla_\theta L_n(\hat{\theta}_n,\gamma_{t})-\nabla_\theta L_n(\hat{\theta}_n,\hat{\gamma}_n)\|_2\|\theta_t-\hat{\theta}_n\|_2\\
&\leq\|\nabla^2_{\theta\gamma}L_n(\hat{\theta}, \bar{\gamma}_t)\|_2\|\gamma_{t}-\hat{\gamma}_n\|_2\|\theta_t-\hat{\theta}_n\|_2\\
&\leq (\|\nabla^2_{\theta\gamma}L_n(\theta_*, \bar{\gamma}_t)\|_2+c_3\epsilon)\|\gamma_{t}-\hat{\gamma}_n\|_2\|\theta_t-\hat{\theta}_n\|_2\\
&\leq (cb+M+c_3\epsilon+f(d_1+d_2, n, \delta, K_1))\|\gamma_{t}-\hat{\gamma}_n\|_2\|\theta_t-\hat{\theta}_n\|_2\\
&\leq \frac{cb+M+c_3\epsilon+f(d_1+d_2, n, \delta, K_1)}{2}(\|\gamma_{t}-\hat{\gamma}_n\|_2^2+\|\theta_t-\hat{\theta}_n\|_2^2).
\end{align*}
Therefore,
\begin{equation}\label{t1e23}
\begin{aligned}
&~~~~ \langle\nabla_\theta L_n(\hat{\theta}_n,\gamma_{t})-\nabla_\theta L_n(\hat{\theta}_n,\hat{\gamma}_n), \theta_t-\hat{\theta}_n\rangle\\
&\geq -\frac{cb+M+c_3\epsilon+f(d_1+d_2, n, \delta, K_1)}{2}(\|\gamma_{t}-\hat{\gamma}_n\|_2^2+\|\theta_t-\hat{\theta}_n\|_2^2). 
\end{aligned}
\end{equation}
By \eqref{t1e21}, \eqref{t1e22} and \eqref{t1e23}, we obtain
\begin{equation}\label{t1e25}
\begin{aligned}
    \langle\nabla_\theta L_n(\theta_{t},\gamma_{t}), \theta_t-\hat{\theta}_n\rangle\geq& \left(w-\frac{cb+M+c_3\epsilon+f(d_1+d_2, n, \delta, K_1)}{2}\right)\|\theta_t-\hat{\theta}_n\|_2^2\\
    &-\frac{cb+M+c_3\epsilon+f(d_1+d_2, n, \delta, K_1)}{2}\|\gamma_{t}-\hat{\gamma}_n\|_2^2.
\end{aligned}    
\end{equation}
Combining \eqref{t1e200}, \eqref{t1e24} and \eqref{t1e25}, with probability at least $1-d_1  \left(\frac{e}{2}\right) ^{-\frac{\lambda_{\phi}n}{8d_2K}}$, we have
\begin{equation}\label{s1e36}
\begin{aligned}
  \|\theta_{t+1}-\hat{\theta}_n\|_2^2\leq& [1+2d_2^2K^2\eta_1^2- (2w-cb-M-c_3\epsilon-f(d_1+d_2, n, \delta, K_1))\eta_1]\|\theta_t-\hat{\theta}_n\|_2^2\\
  &+\{2\tilde{M}^2\eta_1^2+[cb+M+c_3\epsilon+f(d_1+d_2, n, \delta, K_1)]\eta_1\}\|\gamma_{t}-\hat{\gamma}_n\|_2^2.
  \end{aligned}
\end{equation}
Now we bound $\|\gamma_{t+1}-\hat{\gamma}_n\|_2^2$. According to the update in Algorithm \ref{alg1}, we have
\begin{equation}\label{t1e300}
\begin{aligned}
 \|\gamma_{t+1}-\hat{\gamma}_n\|_2^2 &=\|\gamma_t-\eta_2\nabla_\gamma L_n(\theta_{t+1},\gamma_{t})-\hat{\gamma}_n\|_2^2\\
& = \|\gamma_t-\hat{\gamma}_n\|_2^2+\eta_2^2\|\nabla_\gamma L_n(\theta_{t+1},\gamma_{t})\|_2^2-2\eta_2\langle\nabla_\gamma L_n(\theta_{t+1},\gamma_{t}), \gamma_t-\hat{\gamma}_n\rangle.
\end{aligned}    
\end{equation}
Note that $\nabla_\gamma L_n(\hat{\theta}_n,\hat{\gamma}_n)=0$. By Lemmas \ref{lem2} and \ref{lem4}, we obtain
\begin{equation}\label{t1e304}
 \begin{aligned}
\|\nabla_\gamma L_n(\theta_{t+1},\gamma_{t})\|_2^2&=\|\nabla_\gamma L_n(\theta_{t+1},\gamma_{t})-\nabla_\gamma L_n(\theta_{t+1},\hat{\gamma}_n)+\nabla_\gamma L_n(\theta_{t+1},\hat{\gamma}_n)-\nabla_\gamma L_n(\hat{\theta}_n,\hat{\gamma}_n)\|_2^2\\
&\leq 2\|\nabla_\gamma L_n(\theta_{t+1},\gamma_{t})-\nabla_\gamma L_n(\theta_{t+1},\hat{\gamma}_n)\|_2^2+2\|\nabla_\gamma L_n(\theta_{t+1},\hat{\gamma}_n)-\nabla_\gamma L_n(\hat{\theta}_n,\hat{\gamma}_n)\|_2^2\\
&\leq 2d_2^2K^2\|\gamma_t-\hat{\gamma}_n\|_2^2+2\tilde{M}^2\|\theta_{t+1}-\hat{\theta}_n\|_2^2.
\end{aligned}   
\end{equation}
Next,
\begin{equation}\label{t1e301}
\begin{aligned}
&~~~~\langle\nabla_\gamma L_n(\theta_{t+1},\gamma_{t}), \gamma_t-\hat{\gamma}_n\rangle\\
&=\langle\nabla_\gamma L_n(\theta_{t+1},\gamma_{t})-\nabla_\gamma L_n(\theta_{t+1},\hat{\gamma}_{n}), \gamma_t-\hat{\gamma}_n\rangle+\langle\nabla_\gamma L_n(\theta_{t+1},\hat{\gamma}_{n})-\nabla_\gamma L_n(\hat{\theta}_n,\hat{\gamma}_n), \gamma_t-\hat{\gamma}_n\rangle . 
\end{aligned}
\end{equation}
By Lemma \ref{lem02}, with probability at least $1-d_2  \left(\frac{e}{2}\right) ^{-\frac{\lambda_{\psi}n}{8d_1K}}$, the following
\begin{equation}\label{t1e302}
\langle \nabla_\gamma L_n(\theta_{t+1}, \gamma_t)-\nabla_\gamma L_n(\theta_{t+1}, \hat{\gamma}_n), \gamma_t-\hat{\gamma}_n\rangle \geq  w\|\gamma_t-\hat{\gamma}_n\|_2^2,
\end{equation}
holds for all $t\in\{1,\cdots,T\}$. Denote $\bar{\theta}_{t}=a\theta_t+(1-a)\hat{\theta}_n$ for $a\in[0, 1]$. We have 
\begin{align*}
    \|\bar{\theta}_t-\theta_*\|_2&=\|a\theta_t+(1-a)\hat{\theta}_n-\theta_*\|_2\\
    &\leq a\|\theta_t-\theta_*\|_2+(1-a)\|\hat{\theta}_n-\theta_*\|_2\\
    &\leq ab/\sqrt{2}+(1-a)\epsilon/\sqrt{2}\\
    &\leq b/\sqrt{2},
\end{align*}
and
\begin{align*}
\|\nabla^2_{\theta\gamma}L_n(\bar{\theta}_t, \hat{\gamma}_n)\|_2&=\|\nabla^2_{\theta\gamma}L_n(\bar{\theta}_t, \gamma_*)+\nabla^2_{\theta\gamma}L_n(\bar{\theta}_t, \hat{\gamma}_n)-\nabla^2_{\theta\gamma}L_n(\bar{\theta}_t, \gamma_*)\|_2\\
&\leq \|\nabla^2_{\theta\gamma}L_n(\bar{\theta}_t, \gamma_*)\|_2+\|\nabla^2_{\theta\gamma}L_n(\bar{\theta}_t, \hat{\gamma}_n)-\nabla^2_{\theta\gamma}L_n(\bar{\theta}_t, \gamma_*)\|_2\\
&\leq \|\nabla^2_{\theta\gamma}L_n(\bar{\theta}_t, \gamma_*)\|_2+\sqrt{2}c'_3\|\hat{\gamma}_n-\gamma_*\|_2\\
&\leq \|\nabla^2_{\theta\gamma}L_n(\bar{\theta}_t, \gamma_*)\|_2+c'_3\epsilon
\end{align*}
for some positive constant $c'_3$.
By Lemma \ref{lem9}, with probability at least $1-
\delta$, we have
\begin{align*}
&~~~~|\langle\nabla_\gamma L_n(\theta_{t+1}, \hat{\gamma}_{n})-\nabla_\gamma L_n(\hat{\theta}_{n},\hat{\gamma}_n), \gamma_t-\hat{\gamma}_n\rangle|\\
&\leq \|\nabla_\gamma L_n(\theta_{t+1}, \hat{\gamma}_{n})-\nabla_\gamma L_n(\hat{\theta}_n,\hat{\gamma}_n)\|_2\|\gamma_t-\hat{\gamma}_n\|_2\\
&\leq \nabla^2_{\theta\gamma}L_n(\bar{\theta}_{t+1}, \hat{\gamma}_n)\|\theta_{t+1}-\hat{\theta}_n\|_2\|\gamma_{t}-\hat{\gamma}_n\|_2\\
&\leq (\|\nabla^2_{\theta\gamma}L_n(\bar{\theta}_{t+1}, \gamma_*)\|_2+c'_3\epsilon)\|\theta_{t+1}-\hat{\theta}_n\|_2\|\gamma_{t}-\hat{\gamma}_n\|_2 \\
&\leq \frac{cb+c_3'\epsilon+M+f(d_1+d_2, n, \delta, K_1)}{2}(\|\gamma_{t}-\hat{\gamma}_n\|_2^2+\|\theta_{t+1}-\hat{\theta}_n\|_2^2).
\end{align*}
Therefore,
\begin{equation}\label{t1e303}
\begin{aligned}
&~~~~\langle\nabla_\gamma L_n(\theta_{t+1},\hat{\gamma}_{n})-\nabla_\gamma L_n(\hat{\theta}_n,\hat{\gamma}_n), \gamma_t-\hat{\gamma}_n\rangle\\
&\geq -\frac{cb+c_3'\epsilon+M+f(d_1+d_2, n, \delta, K_1)}{2}(\|\gamma_{t}-\hat{\gamma}_n\|_2^2+\|\theta_{t+1}-\hat{\theta}_n\|_2^2).
\end{aligned}
\end{equation}
By \eqref{t1e301}, \eqref{t1e302} and \eqref{t1e303}, we obtain
\begin{equation}\label{t1e305}
\begin{aligned}
\langle\nabla_\gamma L_n(\theta_{t+1},\gamma_{t}), \gamma_t-\hat{\gamma}_n\rangle\geq &\left(w-\frac{cb+c_3'\epsilon+M+f(d_1+d_2, n, \delta, K_1)}{2}\right)\|\gamma_t-\hat{\gamma}_n\|_2^2\\
&-\frac{cb+c_3'\epsilon+M+f(d_1+d_2, n, \delta, K_1)}{2}\|\theta_{t+1}-\hat{\theta}_n\|_2^2. 
\end{aligned}
\end{equation}
Combining \eqref{t1e300}, \eqref{t1e304} and \eqref{t1e305}, with probability at least $1-d_2 \left(\frac{e}{2}\right) ^{-\frac{\lambda_{\psi}n}{8d_1K}}$, we have
\begin{equation}\label{s1306}
\begin{aligned}
\|\gamma_{t+1}-\hat{\gamma}_n\|_2^2\leq& [1+2d_1^2K^2\eta_2^2- (2w-cb-c_3'\epsilon-M-f_3(n, \delta))\eta_2]\|\gamma_t-\hat{\gamma}_n\|_2^2\\
&+\{2\tilde{M}^2\eta_2^2+[cb+c_3'\epsilon+M+f(d_1+d_2, n, \delta, K_1)]\eta_2 \}\|\theta_{t+1}-\hat{\theta}_n\|_2^2.
\end{aligned}
\end{equation}
Let $A_{1n}$ and $A_{2n}$ be defined in \eqref{a1a2}, and $\eta$ be defined in \eqref{eta}. When $$\epsilon\leq \frac{\min\{d_2^2K^2\eta_1^2+w\eta_1, d_1^2K^2\eta_2^2+w\eta_2, \tilde{M}^2\eta\}}{\max\{c_3, c'_3\}},$$ we can verify 
$1+\max\{2d_2^2K^2\eta_1^2- (2w-cb-c_3\epsilon-M-f(d_1+d_2, n, \delta, K_1))\eta_2,2d_1^2K^2\eta_2^2- (2w-cb-c_3'\epsilon-M-f(d_1+d_2, n, \delta, K_1))\eta_2\}\leq A_{1n}$ and $2\tilde{M}^2\eta^2+[cb+\max\{c_3, c_3'\}\epsilon+M+f(d_1+d_2, n, \delta, K_1)]\eta\leq A_{2n}$.
By \eqref{s1e36} and \eqref{s1306}, with probability at least $1-\delta-d_1 \left(\frac{e}{2}\right) ^{-\frac{\lambda_{\phi}n}{8d_2K}}-d_2\left(\frac{e}{2}\right) ^{-\frac{\lambda_{\psi}n}{8d_1K}}$, we can obtain
\begin{equation}\label{new3}
  \|\theta_{t+1}-\hat{\theta}_n\|_2^2\leq A_{1n}\|\theta_{t}-\hat{\theta}_n\|_2^2+A_{2n}\|\gamma_{t}-\hat{\gamma}_n\|_2^2
\end{equation}
and 
\begin{equation}\label{new4}
 \begin{aligned}
\|\gamma_{t+1}-\hat{\gamma}_n\|_2^2&\leq A_{1n}\|\gamma_{t}-\hat{\gamma}_n\|_2^2+A_{2n}\|\theta_{t+1}-\hat{\theta}_n\|_2^2\\
&\leq (A_{1n}+A_{2n}^2)\|\gamma_{t}-\hat{\gamma}_n\|_2^2+A_{1n}A_{2n}\|\theta_{t}-\hat{\theta}_n\|_2^2.
\end{aligned}   
\end{equation}
Denote $n'_0=\max\{n_0, n_{\epsilon}\}$. By \eqref{new1},  \eqref{new3} and \eqref{new4}, we have
\begin{align*}
\|\theta_{t+1}-\hat{\theta}_n\|_2^2+\|\gamma_{t+1}-\hat{\gamma}_n\|_2^2
&\leq (A_{1n}+A_{2n}+A_{2n}^2)\|\gamma_t-\hat{\gamma}_n\|_2^2+(A_{1n}+A_{1n}A_{2n})\|\theta_{t}-\hat{\theta}_n\|_2^2
\\
&\leq (A_{1n}+A_{2n}+A_{2n}^2)(\|\gamma_t-\hat{\gamma}_n\|_2^2+\|\theta_{t}-\hat{\theta}_n\|_2^2)\\
&\leq \rho_1 (\|\gamma_t-\hat{\gamma}_n\|_2^2+\|\theta_{t}-\hat{\theta}_n\|_2^2),
\end{align*}
where $\rho_1=A_{1n'_0}+A_{2n'_0}+A_{2n'_0}^2<1$ according to \eqref{new2}. For the initialization, we have $\|\theta_0-\hat{\theta}_n\|_2\leq \|\theta_t-\theta_*\|_2+\|\theta_*-\hat{\theta}_n\|_2\leq (b+\epsilon)/\sqrt{2}$ and $\|\gamma_t-\hat{\gamma}_n\|_2\leq \|\gamma_t-\gamma_*\|_2+\|\gamma_*-\hat{\gamma}_n\|_2\leq (b+\epsilon)/\sqrt{2}$. Thus, $\|\theta_0-\hat{\theta}_n\|_2^2+\|\gamma_*-\hat{\gamma}_n\|_2^2\leq (b+\epsilon)^2$.
By recursion, with probability at least $1-\delta -d_1 \left(\frac{e}{2}\right) ^{-\frac{\lambda_{\phi}n}{8d_2K}}-d_2\left(\frac{e}{2}\right) ^{-\frac{\lambda_{\psi}n}{8d_1K}}$, when $n> n_0'$, we obtain
$$\|\theta_{T}-\hat{\theta}_n\|_2^2+\|\gamma_{T}-\hat{\gamma}_n\|_2^2\leq \rho_1^T(b+\epsilon)^2.$$

\subsection{Proof of Theorem \ref{thm3}}
We denote $v=(\phi^\top(s,a), 0)^\top\in\mathbb{R}^{d_1+d_2}$. Then, by Assumption \ref{ass1}, we have
$$v^\top (\tau_T-\tau_*)=(\theta_T-\theta_*)^\top \phi(s,a)=r_{\theta_T}(s, a)-r_{\theta_*}(s,a).$$ By Lemma \ref{block}, we have
\begin{align*}
v^\top \tilde{\mathcal{I}}^{-1}(\tau_*) v&=\phi^\top(s,a)[\mathcal{I}_{\theta\theta}(\theta_*, \gamma_*)-\mathcal{I}_{\theta\gamma} (\theta_*, \gamma_*)\mathcal{I}^{-1}_{\gamma\gamma} (\theta_*, \gamma_*)\mathcal{I}_{\theta\gamma} (\theta_*, \gamma_*)]^{-1}\phi(s,a)\\
&=\phi^\top(s,a)\tilde{\mathcal{I}}^{-1}(\theta_*, \gamma_*)\phi(s,a).
\end{align*}
Therefore, by Theorem \ref{thm6}, as $n\rightarrow\infty$ and $T\rightarrow\infty$, we have
\begin{equation*}
\sqrt{n}[r_{\theta_T}(s, a)-r_{\theta_*}(s,a)]\stackrel{D}{\longrightarrow} \mathcal{N}(0, \phi^\top(s,a)\tilde{\mathcal{I}}^{-1}(\theta_*, \gamma_*)\phi(s,a)). 
\end{equation*}

\subsection{Proof of Theorem \ref{thm4}}
Because the samples $\{(x_i,s_i, a^{(0)}_i, a^{(1)}_i)\}_{i=1}^n$ are i.i.d., the law of large numbers implies that, as $n\rightarrow\infty$, entry-wisely, we have
$$\hat{\mathcal{I}}_{\theta\theta}(\theta_*, \gamma_*)\stackrel{p}{\longrightarrow} \mathcal{I}_{\theta\theta}(\theta_*, \gamma_*), \ \hat{\mathcal{I}}_{\theta\gamma}(\theta_*, \gamma_*) \stackrel{p}{\longrightarrow} \mathcal{I}_{\theta\gamma}(\theta_*, \gamma_*), \ \hat{\mathcal{I}}_{\gamma\gamma}(\theta_*, \gamma_*) \stackrel{p}{\longrightarrow} \mathcal{I}_{\gamma\gamma}(\theta_*, \gamma_*).$$
Consequently, for sufficiently large $n$, we have
\begin{align*}
  \|I-\mathcal{I}^{-1}_{\theta\theta}(\theta_*, \gamma_*)\hat{\mathcal{I}}_{\theta\theta}(\theta_*, \gamma_*)\|_2&=\|\mathcal{I}^{-1}_{\theta\theta}(\theta_*, \gamma_*)[\mathcal{I}_{\theta\theta}(\theta_*, \gamma_*)-\hat{\mathcal{I}}_{\theta\theta}(\theta_*, \gamma_*)]\|_2\\
  &\leq \|\mathcal{I}^{-1}_{\theta\theta}(\theta_*, \gamma_*)\|_2\|\mathcal{I}_{\theta\theta}(\theta_*, \gamma_*)-\hat{\mathcal{I}}_{\theta\theta}(\theta_*, \gamma_*)\|_2<1  
\end{align*}
with probability approaching 1. By Lemma \ref{lem613}, this implies that $\mathcal{I}^{-1}_{\theta\theta}(\theta_*, \gamma_*)\hat{\mathcal{I}}_{\theta\theta}(\theta_*, \gamma_*)$ is invertible, and therefore $\hat{\mathcal{I}}_{\theta\theta}(\theta_*, \gamma_*)$ is invertible. An analogous argument applies to $\hat{\mathcal{I}}_{\gamma\gamma}(\theta_*, \gamma_*)$. By the continuous mapping theorem, the inverses of these matrices converge in probability to the inverses of the population matrices.
Since $S_\theta^2(\theta_*, \gamma_*)= [\hat{\mathcal{I}}_{\theta\theta}(\theta_*, \gamma_*)-\hat{\mathcal{I}}_{\theta\gamma}(\theta_*, \gamma_*)\hat{\mathcal{I}}^{-1}_{\gamma\gamma} (\theta_*, \gamma_*)\hat{\mathcal{I}}^{\top}_{\theta\gamma} (\theta_*, \gamma_*)]^{-1}$, as $n\rightarrow\infty$, we have 
\begin{equation}\label{thm41}
   \|S_\theta^2(\theta_*,\gamma_*)-\tilde{\mathcal{I}}^{-1}(\theta_*,\gamma_*)\|_2\stackrel{p}{\longrightarrow} 0. 
\end{equation}
Note that $\mu(\cdot)[1-\mu(\cdot)]\leq 1/4$. Then, by Assumptions \ref{ass1} and \ref{ass2}, there exist some constants $C_1$ and $C_2$, such that
\begin{align*}
\|\hat{\mathcal{I}}_{\theta\theta}(\theta_T, \gamma_T)-\hat{\mathcal{I}}_{\theta\theta}(\theta_*, \gamma_*)\|_2&\leq \frac{1}{4n}\left\|\sum_{i=1}^n\{[\gamma_T^\top\psi(x_i)]^2-[\gamma_*^\top\psi(x_i)]^2\}z_iz_i^\top\right\|_2\\
&= \frac{1}{4n}\left\|\sum_{i=1}^n\{[\gamma_T^\top\psi(x_i)-\gamma_*^\top\psi(x_i)][\gamma_T^\top\psi(x_i)+\gamma_*^\top\psi(x_i)]\}z_iz_i^\top\right\|_2\\
&\leq \frac{1}{4n}\sum_{i=1}^n\|\gamma_T-\gamma_*\|_2\|\psi(x_i)\|_2|\gamma_T^\top\psi(x_i)+\gamma_*^\top\psi(x_i)|\|_2z_iz_i^\top\|_2\\
&\leq C_1\|\gamma_T-\gamma_*\|_2
\end{align*}
and
\begin{align*}
  \|\hat{\mathcal{I}}_{\gamma\gamma}(\theta_T, \gamma_T)-\hat{\mathcal{I}}_{\gamma\gamma}(\theta_*, \gamma_*)\|_2&\leq \frac{1}{4n}\left\|\sum_{i=1}^n[(\theta_T^\top z)^2-(\theta^\top z)^2]\psi(x_i)\psi(x_i)^\top\right\|_2\\
  &=\frac{1}{4n}\left\|\sum_{i=1}^n[(\theta_T^\top z-\theta^\top z)(\theta_T^\top z+\theta^\top z)]\psi(x_i)\psi(x_i)^\top\right\|_2\\
  &\leq \frac{1}{4n}\sum_{i=1}^n\|\theta_T-\theta\|_2\|z\|_2|\theta_T^\top z+\theta^\top z|\|\psi(x_i)\psi(x_i)^\top\|_2\\
  &\leq C_2\|\theta_T-\theta_*\|_2.  
\end{align*}
By Assumptions \ref{ass1} and \ref{ass2}, there exists some constant $C_3$ such that
\begin{align*}
 &~~~~\|\hat{\mathcal{I}}_{\theta\gamma}(\theta_T, \gamma_T)-\hat{\mathcal{I}}_{\theta\gamma}(\theta_*, \gamma_*)\|_2\\
 &= \|\hat{\mathcal{I}}_{\theta\gamma}(\theta_T, \gamma_T)-\hat{\mathcal{I}}_{\theta\gamma}(\theta_*, \gamma_T)+\hat{\mathcal{I}}_{\theta\gamma}(\theta_*, \gamma_T)-\hat{\mathcal{I}}_{\theta\gamma}(\theta_*, \gamma_*)\|_2\\
 &= \frac{1}{4n}\left\|\sum_{i=1}^n\{[\sigma_{\gamma_T}(x_i)(\theta_T-\theta_*)^\top z_i]+(\gamma_T-\gamma_*)^\top\psi(x_i)(\theta_*^\top z_i)\}\psi(x_i)z_i^\top\right \|_2\\
 &\leq \frac{1}{4n}\sum_{i=1}^n[|\sigma_{\gamma_T}(x_i)|\|\theta_T-\theta_*\|_2 \|z_i\|_2+\|\gamma_T-\gamma_*\|_2\|\psi(x_i)\|_2|\theta_*^\top z_i|]\|\psi(x_i)z_i^\top\|_2\\
 &\leq C_3 (\|\theta_T-\theta_*\|_2+\|\gamma_T-\gamma_*\|_2).
\end{align*}
Since $$S_\theta^2(\theta_T, \gamma_T)= [\hat{\mathcal{I}}_{\theta\theta}(\theta_T, \gamma_T)-\hat{\mathcal{I}}_{\theta\gamma}(\theta_T, \gamma_T)\hat{\mathcal{I}}^{-1}_{\gamma\gamma} (\theta_T, \gamma_T)\hat{\mathcal{I}}^{\top}_{\theta\gamma} (\theta_T, \gamma_T)]^{-1},$$
by Theorem \ref{thm1}, we have $\|S_\theta^2(\theta_T,\gamma_T)-S_\theta^2(\theta_*,\gamma_*)\|_2\stackrel{p}{\longrightarrow} 0$ as $n\rightarrow\infty$ and $T\rightarrow\infty$. Combining \eqref{thm41}, we have
\begin{align*}
&~~~~\|S_\theta^2(\theta_T,\gamma_T)-\tilde{\mathcal{I}}^{-1}(\theta_*,\gamma_*)\|_2\\
&=\|S_\theta^2(\theta_T,\gamma_T)-S_\theta^2(\theta_*,\gamma_*)+S_\theta^2(\theta_*,\gamma_*)-\tilde{\mathcal{I}}^{-1}(\theta_*,\gamma_*)\|_2\\
&\leq \|S_\theta^2(\theta_T,\gamma_T)-S_\theta^2(\theta_*,\gamma_*)\|_2+\|S_\theta^2(\theta_*,\gamma_*)-\tilde{\mathcal{I}}^{-1}(\theta_*,\gamma_*)\|_2\\
&\stackrel{p}{\longrightarrow} 0.
\end{align*}
By Theorem \ref{thm3} and Slutsky's theorem, we have
$$\frac{\sqrt{n}[r_{\theta_T}(s, a)-r_{\theta_*}(s, a)]}{\sqrt{\phi^\top(s, a)S_\theta^2(\theta_T, \gamma_T)\phi(s, a)}}\stackrel{D}{\longrightarrow}\mathcal{N}(0, 1).$$

\subsection{Proof of Theorem \ref{thm4.1}}
The proof of Theorem \ref{thm4.1} follows the same strategy as used in the proofs for Theorems \ref{thm3} and \ref{thm4}, and hence is omitted.

\subsection{Proof of Corollary \ref{cor2}}
    When $T>\frac{\log [(1-\rho)b^2n]-\log \{c_2\log (1/\delta)\}}{-\log \rho}$ (equivalent to $T>c\log n$ for some constant $c$), we can verify $\rho^Tb^2< \frac{c_2\log (1/\delta)}{(1-\rho)n}$. By Theorem \ref{thm1}, with probability at least $1-\delta-\frac{1}{n}$, we have
    \begin{align*}
        \|\theta_T-\theta_*\|^2_2\leq \frac{2c_2\log (1/\delta)}{(1-\rho)n}.
    \end{align*}
    Therefore,
    \begin{align*}
\|\theta_T-\theta_*\|_{S^{-2}(\theta_T,\gamma_T)}\leq \|\theta_T-\theta_*\|_2\|S^{-2}(\theta_T,\gamma_T)\|_2\leq \frac{c_3}{\sqrt{n}},        
    \end{align*}
  where $c_3= \sqrt{\frac{2c_2\log (1/\delta)}{1-\rho}}\|S^{-2}(\theta_T,\gamma_T)\|_2$. Since $q_{1-\alpha/2}$ can take any positive value, we can choose $\alpha$ such that $ q_{1-\alpha/2}=c_3$. The proof is complete.

\subsection{Proof of Lemma \ref{lowpess}}
By Assumption \ref{ass1}, we have $r_\theta(s,a) = \theta^\top \phi(s,a) $.
Then, for any $\theta$, we can write
$r_\theta(s,a) = r_{\theta_T}(s,a) + \phi(s,a)^\top(\theta - \theta_T)$. Then minimizing $r_\theta(s,a)$ over $\tilde{\Theta}$ is equivalent to
\[
\min_{\theta \in \tilde{\Theta}} r_\theta(s,a)
= r_{\theta_T}(s,a) + \min_{\|\theta - \theta_T\|_{S_\theta^{-2}(\theta_T,\gamma_T)} \le \frac{q_{1-\alpha/2}}{\sqrt{n}}} (\theta - \theta_T)^\top\phi(s,a).
\]
 By the Cauchy-Schwarz inequality, we have
 \begin{equation}\label{lgl1}
  \begin{aligned}
    (\theta - \theta_T)^\top\phi(s,a)& =(\theta - \theta_T)^\top S_\theta^{-1}(\theta_T,\gamma_T)S_\theta(\theta_T,\gamma_T)\phi(s,a)\\
    &\geq -\|S_\theta^{-1}(\theta_T,\gamma_T)(\theta - \theta_T)\|_2\|S_\theta(\theta_T,\gamma_T)\phi(s,a)\|_2\\
&= -\sqrt{(\theta - \theta_T)^\top S_\theta^{-2}(\theta_T,\gamma_T)(\theta - \theta_T)}\sqrt{\phi^\top(s,a) S_\theta^{2}(\theta_T,\gamma_T)\,\phi(s,a)}\\
&\geq  -q_{1-\alpha/2}\sqrt{\frac{\phi^\top(s,a) S_\theta^{2}(\theta_T,\gamma_T)\,\phi(s,a)}{n}},
\end{aligned}   
 \end{equation}
where the last inequality is from the definition of $\tilde{\Theta}$. Pick $
\theta = \theta_T - \frac{q_{1-\alpha/2}}{\sqrt{n}}
\frac{S^{2}_\theta(\theta_T,\gamma_T)\phi(s,a)}
     {\sqrt{\phi(s,a)^\top S^{2}_\theta(\theta_T,\gamma_T)\phi(s,a)}},
$
which satisfies $\|\theta - \theta_T\|_{S^{-2}_\theta(\theta_T,\gamma_T)} \leq q_{1-\alpha/2}/\sqrt{n}$. We have 
\begin{equation}\label{lgl2}
    \begin{aligned}
  (\theta - \theta_T)^\top\phi(s,a)&=  -\frac{q_{1-\alpha/2}}{\sqrt{n}}
\frac{\phi^\top(s,a)S^{2}_\theta(\theta_T,\gamma_T)\phi(s,a)}
     {\sqrt{\phi(s,a)^\top S^{2}_\theta(\theta_T,\gamma_T)\phi(s,a)}}\\&=-q_{1-\alpha/2}\sqrt{\frac{\phi^\top(s,a) S_\theta^{2}(\theta_T,\gamma_T)\,\phi(s,a)}{n}}.
\end{aligned}
\end{equation}
By \eqref{lgl1} and \eqref{lgl2}, we have
$$\min_{\|\theta - \theta_T\|_{S_\theta^{-2}(\theta_T,\gamma_T)} \le \frac{q_{1-\alpha/2}}{\sqrt{n}}} (\theta - \theta_T)^\top\phi(s,a)=-q_{1-\alpha/2}\sqrt{\frac{\phi^\top(s,a) S_\theta^{2}(\theta_T,\gamma_T)\,\phi(s,a)}{n}}.$$
Therefore,
$$\min_{\theta \in \tilde{\Theta}} r_\theta(s,a)
= r_{\theta_T}(s,a)-q_{1-\alpha/2}\sqrt{\frac{\phi^\top(s,a) S_\theta^{2}(\theta_T,\gamma_T)\,\phi(s,a)}{n}}.$$
This completes the proof.

\subsection{Proof of Theorem \ref{thm7}}
 For notational simplicity, write $a_*=a_*(s)$ and $a_{pBoN}=a_{pBoN}(s)$. The reward difference between $a_*$ and $a_{pBoN}$ for $s$ is 
 \begin{equation}\label{dec1}
 \begin{aligned}
    &~~~~r_{\theta_*}(s, a_*)-r_{\theta_*}(s, a_{pBoN})\\
    &=[r_{\theta_*}(s, a_*)-\hat{r}(s, a_*)]+[\hat{r}(s, a_*)-\hat{r}(s, a_{pBoN})]+[\hat{r}(s, a_{pBoN})-r_{\theta_*}(s, a_{pBoN})].
\end{aligned}    
 \end{equation}
Since $a_{pBoN}$ is the optimal action under $\hat{r}(s, a)$, i.e., $a_{pBoN}(s)=\arg\max_{a\in \mathcal{A}_N(s)}\hat{r}(s, a)$, the second difference satisfies 
\begin{equation}\label{dec2}
    \hat{r}(s, a_*)-\hat{r}(s, a_{pBoN})\leq 0.
\end{equation}
For the third difference, by \eqref{pesf}, we have
\begin{align*}
\hat{r}(s, a_{pBoN})-r_{\theta_*}(s, a_{pBoN})=\min_{\theta\in\tilde{\Theta}}r_{\theta}(s, a_{pBoN})-r_{\theta_*}(s, a_{pBoN}),
\end{align*}
where $\tilde{\Theta}=\{\theta: \|\theta-\theta_T\|_{S_\theta^{-2}(\theta_T,\gamma_T)}\leq q_{1-\alpha/2}/\sqrt{n}\}$.
From Corollary \ref{cor2}, we have $\|\theta_*-\theta_T\|_{S_\theta^{-2}(\theta_T,\gamma_T)}\leq q_{1-\alpha/2}/\sqrt{n}$ with probability at least $1-\delta-\frac{1}{n}$, i.e., $\mathbb{P}(\theta_*\in\tilde{\Theta})\geq 1-\delta-\frac{1}{n}$.  Thus, with probability at least $1-\delta-\frac{1}{n}$, we have $\min_{\theta\in\tilde{\Theta}}r_{\theta}(s, a_{pBoN})\leq r_{\theta_*}(s, a_{pBoN})$, and then 
\begin{equation}\label{dec3}
    \hat{r}(s, a_{pBoN})-r_{\theta_*}(s, a_{pBoN})\leq 0.
\end{equation}
Now, we bound the first difference. By Assumption \ref{ass1}, we obtain 
\begin{align*}
    r_{\theta_*}(s, a_*)-\hat{r}(s, a_*)&=\theta_*^\top\phi(s, a_*)-\min_{\theta\in\tilde{\Theta}}\theta^\top \phi(s, a_*)\\
    &=\max_{\theta\in\tilde{\Theta}}(\theta_*-\theta)^\top\phi(s, a_*)\\
     &=\max_{\theta\in\tilde{\Theta}}(\theta_*-\theta_T+\theta_T-\theta)^\top\phi(s, a_*)\\
     &=(\theta_*-\theta_T)^\top\phi(s, a_*)+\max_{\theta\in\tilde{\Theta}}(\theta_T-\theta)^\top\phi(s, a_*)\\
      &\leq |(\theta_T-\theta_*)^\top\phi(s, a_*)|+\max_{\theta\in\tilde{\Theta}}|(\theta_T-\theta)^\top\phi(s, a_*)|.
\end{align*}
Since  $\theta_*\in\tilde{\Theta}$ with probability at least $1-\delta-\frac{1}{n}$ by Corollary \ref{cor2}, we have 
$|(\theta_T-\theta_*)^\top\phi(s, a_*)|\leq \max_{\theta\in\tilde{\Theta}}|(\theta_T-\theta)^\top\phi(s, a_*)|$ with probability at least $1-\delta-\frac{1}{n}$. Therefore, with probability at least $1-\delta-\frac{1}{n}$, we have 
\begin{equation}\label{dec4}
  \begin{aligned}
r_{\theta_*}(s, a_*)-\hat{r}(s, a_*)&\leq 2\max_{\theta\in\tilde{\Theta}}|(\theta_T-\theta)^\top\phi(s, a_*)|\\
     &\leq 2\max_{\theta\in\tilde{\Theta}}\|\theta_T-\theta\|_{S_\theta^{-2}(\theta_T,\gamma_T)}\|S_\theta(\theta_T,\gamma_T)\phi(s, a_*)\|_2\\
     &\leq 2q_{1-\alpha/2} \|S_\theta(\theta_T,\gamma_T)\phi(s, a_*)\|_2/\sqrt{n}\\
     &\leq 3q_{1-\alpha/2} \|\tilde{\mathcal{I}}^{-\frac{1}{2}}(\theta_*,\gamma_*)\phi(s, a_*)\|_2/\sqrt{n},
\end{aligned}  
\end{equation}
where the third inequality follows from the definition of $\tilde{\Theta}$, and the last inequality holds when $T$ and $n$ are large enough. Therefore, by \eqref{dec1}, \eqref{dec2}, \eqref{dec3}, \eqref{dec4}, with probability at least $1-\delta-\frac{1}{n}$, we have
$$\textsf{SubOpt}(a_{pBoN})=\mathbb{E}_s[r_{\theta_*}(s, a_*)-r_{\theta_*}(s, a_{pBoN})]\leq 2q_{1-\alpha/2} \|\tilde{\mathcal{I}}^{-\frac{1}{2}}(\theta_*,\gamma_*)\mathbb{E}_s\phi(s, a_*(s))\|_2/\sqrt{n}.$$

\section{Supporting Lemmas}\label{supprt}
\begin{lemma} (Corollary 5.2 \citep{Tropp2012}) \label{tropp}
Consider a finite sequence $\{\mathbf{X}_k\}$ of independent, random, self-adjoint matrices with dimension $d$ that satisfy
$$\mathbf{X}_k\succeq \boldsymbol{0}\ \textit{and}\ \lambda_{\max}(\mathbf{X}_k)\leq L\ \textit{almost surely.}$$
Compute the minimum eigenvalue of the sum of expectations,
$\mu_{\min}:=\lambda_{\min}\bigg(\sum_{k}\mathbb{E}\mathbf{X}_k\bigg)$. Then for $\zeta\in [0,1]$,
\begin{align*}
\mathbb{P}\bigg\{\lambda_{\min}\bigg(\sum_{k}\mathbf{X}_k\bigg)\leq (1-\zeta)\mu_{\min}\bigg\}\leq d\bigg[\frac{e^{-\zeta}}{(1-\zeta)^{1-\zeta}}\bigg]^{\mu_{\min}/L}.
\end{align*}
\end{lemma}

\begin{lemma}(Theorem 8.5.11, \citep{harville2011matrix})\label{block}
Let \( \boldsymbol{T} \) represent an \( m \times m \) matrix, $\boldsymbol{U}$ an $m\times n$ matrix, $\boldsymbol{V}$ an $n\times m$ matrix, and $\boldsymbol{W}$ an $n\times n$ matrix. Suppose that $\boldsymbol{T}$ is nonsingular. Then,
$\begin{pmatrix}
\boldsymbol{T} & \boldsymbol{U} \\
\boldsymbol{V} & \boldsymbol{W}
\end{pmatrix}\ \text{or equivalently} \
\begin{pmatrix}
\boldsymbol{W} & \boldsymbol{V} \\
\boldsymbol{U} & \boldsymbol{T}
\end{pmatrix}
$, is nonsingular if and only if the \( n \times n \) matrix
$\boldsymbol{Q} = \boldsymbol{W} - \boldsymbol{V}\boldsymbol{T}^{-1}\boldsymbol{U}$ is nonsingular, in which case
\begin{equation*}
\begin{pmatrix}
\boldsymbol{T} & \boldsymbol{U} \\
\boldsymbol{V} & \boldsymbol{W}
\end{pmatrix}^{-1} =\begin{pmatrix}
\boldsymbol{T}^{-1} + \boldsymbol{T}^{-1}\boldsymbol{U}\boldsymbol{Q}^{-1}\boldsymbol{V}\boldsymbol{T}^{-1} & -\boldsymbol{T}^{-1}\boldsymbol{U}\boldsymbol{Q}^{-1} \\
-\boldsymbol{Q}^{-1}\boldsymbol{V}\boldsymbol{T}^{-1} & \boldsymbol{Q}^{-1}
\end{pmatrix},
\end{equation*}
\begin{equation*}
\begin{pmatrix}
\boldsymbol{W} & \boldsymbol{V} \\
\boldsymbol{U} & \boldsymbol{T}
\end{pmatrix}^{-1} = 
\begin{pmatrix}
\boldsymbol{Q}^{-1} & -\boldsymbol{Q}^{-1}\boldsymbol{V}\boldsymbol{T}^{-1} \\
-\boldsymbol{T}^{-1}\boldsymbol{U}\boldsymbol{Q}^{-1} & \boldsymbol{T}^{-1} + \boldsymbol{T}^{-1}\boldsymbol{U}\boldsymbol{Q}^{-1}\boldsymbol{V}\boldsymbol{T}^{-1}
\end{pmatrix}.
\end{equation*}
\end{lemma}

\begin{lemma}(Theorem 14.8.5, \citep{harville2011matrix})\label{schur}
Let $\boldsymbol{A}=\begin{pmatrix}
    \boldsymbol{T}&\boldsymbol{U}\\
    \boldsymbol{U^\top}&\boldsymbol{W}
\end{pmatrix}$,
where $\boldsymbol{T}$ is of dimensions $m\times m$, $\boldsymbol{U}$ of dimensions $m\times n$, and $\boldsymbol{W}$ of dimensions $n\times n$. If $\boldsymbol{T}$ is symmetric positive definite and the Schur complement $\boldsymbol{W}-\boldsymbol{U}^\top \boldsymbol{T}^{-1}\boldsymbol{U}$
of $\boldsymbol{T}$ is positive definite, or if $\boldsymbol{W}$ is symmetric positive definite and the Schur complement $\boldsymbol{T}-\boldsymbol{U} \boldsymbol{W}^{-1}\boldsymbol{U}^\top$ of $\boldsymbol{W}$  is positive definite, then $\boldsymbol{A}$ is positive
definite.
\end{lemma}
\begin{lemma}(Theorem 17, \citep{ferguson1996})\label{mlecon}
Let \( X_1, X_2, \ldots \) be i.i.d. with density \( f(x \mid \theta) \), \( \theta \in \Theta \), and let \( \theta_0 \) denote the true value of \( \theta \). If
\begin{itemize}
    \item[(1)] \( \Theta \) is compact,
    \item[(2)] \( f(x \mid \theta) \) is upper semicontinuous in \( \theta \) for all \( x \),
    \item[(3)] there exists a function \( K(x) \) such that \( E_{\theta_0}[K(X)] < \infty \) and
    \[
    U(x, \theta) = \log f(x \mid \theta) - \log f(x \mid \theta_0) \leq K(x), \quad \text{for all } x \text{ and } \theta,
    \]
    \item[(4)] for all \( \theta \in \Theta \) and sufficiently small \( \rho > 0 \), \( \sup_{|\theta' - \theta| < \rho} f(x \mid \theta') \) is measurable in \( x \),
    \item[(5)] (identifiability) \( f(x \mid \theta) = f(x \mid \theta_0) \) (a.e. \( dv \)) \( \Rightarrow \theta = \theta_0 \),
\end{itemize}
then, for any sequence of maximum-likelihood estimates \( \hat{\theta}_n \) of \( \theta \),
\[
\hat{\theta}_n \xrightarrow{\text{a.s.}} \theta_0.
\]   
\end{lemma}
\begin{lemma}(Corollary 6.1.2, \citep{tropp2015introduction})\label{lem612}
 Consider a finite sequence $\{S_k\}$ of independent random matrices with common dimension $d_1 \times d_2$. Assume that each matrix has uniformly bounded deviation from its mean:
\[
\| S_k - \mathbb{E} S_k \| \leq L \quad \text{for each index } k.
\]

Introduce the sum
\[
Z = \sum_k S_k,
\]
and let $v(Z)$ denote the matrix variance statistic of the sum:
\begin{align*}
v(Z) &= \max \left\{ \left\| \mathbb{E} \left[ (Z - \mathbb{E}Z)(Z - \mathbb{E}Z)^* \right] \right\|, \,
\left\| \mathbb{E} \left[ (Z - \mathbb{E}Z)^* (Z - \mathbb{E}Z) \right] \right\| \right\} \\
&= \max \left\{ \left\| \sum_k \mathbb{E} \left[ (S_k - \mathbb{E}S_k)(S_k - \mathbb{E}S_k)^* \right] \right\|, \,
\left\| \sum_k \mathbb{E} \left[ (S_k - \mathbb{E}S_k)^* (S_k - \mathbb{E}S_k) \right] \right\| \right\}.
\end{align*}

Then
\[
\mathbb{E} \| Z - \mathbb{E}Z \| \leq \sqrt{2v(Z) \log(d_1 + d_2)} + \frac{1}{3} L \log(d_1 + d_2).
\]

Furthermore, for all $t \geq 0$,
\[
\mathbb{P} \left\{ \| Z - \mathbb{E}Z \| \geq t \right\} \leq (d_1 + d_2) \cdot \exp \left( \frac{-t^2/2}{v(Z) + Lt/3} \right).
\]
\end{lemma}
\begin{lemma}(Corollary 5.6.16, \citep{horn2012matrix})\label{lem613}
    A matrix $A$ is nonsingular if there is a matrix norm $\|\cdot\|$ such
that $\|I-A\|<1$.
\end{lemma}
\end{document}

%% file: def-smile.tex
\let\hat\widehat
\let\tilde\widetilde

\newtheorem{lemma}{{\bf Lemma}}

\newtheorem{corollary}{{\bf Corollary}}

\newtheorem{theorem}{{\bf Theorem}}

\newtheorem{assumption}{{\bf Assumption}}

